# Misspecified Nonconvex Statistical Optimization for Phase Retrieval

Zhuoran Yang[♮*] Lin F. Yang[♮*] Ethan X. Fang[†]
Tuo Zhao[‡] Zhaoran Wang[§] Matey Neykov[¶]

**Abstract**

Existing nonconvex statistical optimization theory and methods crucially rely on the correct specification of the underlying "true" statistical models. To address this issue, we take a first step towards taming model misspecification by studying the high-dimensional sparse phase retrieval problem with misspecified link functions. In particular, we propose a simple variant of the thresholded Wirtinger flow algorithm that, given a proper initialization, linearly converges to an estimator with optimal statistical accuracy for a broad family of unknown link functions. We further provide extensive numerical experiments to support our theoretical findings.

## 1 Introduction

We consider nonconvex optimization for high-dimensional phase retrieval with model misspecification. Phase retrieval finds applications in numerous scientific problems, for example, X-ray crystallography (Harrison, 1993), electron microscopy (Miao et al., 2008), and diffractive imaging (Bunk et al., 2007). See, for example, Candès et al. (2015b) and the references therein for a detailed survey. Most of existing work (Candès et al., 2015b; Cai et al., 2016) casts high-dimensional phase retrieval as a nonconvex optimization problem: Given $n$ data points $\{(y^{(i)}, \boldsymbol{x}^{(i)}) \in \mathbb{R} \times \mathbb{R}^p\}_{i \in [n]}$[1], one aims to solve

$$\underset{\boldsymbol{\beta} \in \mathbb{R}^p}{\text{minimize}} \frac{1}{n} \sum_{i=1}^{n} \big[y^{(i)} - (\boldsymbol{x}^{(i)\top} \boldsymbol{\beta})^2\big]^2, \quad \text{subject to } \|\boldsymbol{\beta}\|_0 \leq s, \tag{1.1}$$

where $\|\boldsymbol{\beta}\|_0$ denotes the number of nonzero entries in $\boldsymbol{\beta}$.

The nonconvex problem in (1.1) gives rise to two challenges in optimization and statistics. From the perspective of optimization, (1.1) is NP-hard in the worst case (Sahinoglou and Cabrera, 1991),

[♮]Equal contribution.
[*]Princeton University; e-mail: {zy6, lin.yang}@princeton.edu.
[†]Pennsylvania State University; e-mail: xxf13@psu.edu.
[‡]Georgia Institute of Technology; e-mail: tuo.zhao@isye.gatech.edu.
[§]Northwestern University; e-mail: zhaoranwang@gmail.com.
[¶]Carnegie Mellon University; e-mail: mneykov@stat.cmu.edu.
[1]Here we use the shorthand $[n] = \{1, 2, \ldots, n\}$.

that is, under computational hardness hypotheses, no algorithm can achieve the global minimum in polynomial time. Particularly, most existing general-purpose first-order or second-order optimization methods (Ghadimi and Lan, 2013; Bolte et al., 2014; Lu and Xiao, 2015; Hong et al., 2016; Ghadimi and Lan, 2016; Xu and Yin, 2017; Gonçalves et al., 2017) are only guaranteed to converge to certain stationary points. Meanwhile, since (1.1) can also be cast as a polynomial optimization problem, we can leverage various semidefinite programming approaches (Parrilo, 2003; Kim et al., 2016; Weisser et al., 2017; Ahmadi and Parrilo, 2017). However, in real applications the problem dimension of practical interest is often large, for example, $p$ can be of the order of millions. To the best of our knowledge, existing polynomial optimization approaches do not scale up to such large dimensions. The difficulty in optimization further leads to more challenges in statistics. From the perspective of statistics, researchers are interested in characterizing the statistical properties of $\widehat{\boldsymbol{\beta}}$ with respect to some underlying ground truth $\boldsymbol{\beta}^*$, for example, the estimation error $\|\widehat{\boldsymbol{\beta}} - \boldsymbol{\beta}^*\|_2$. Nevertheless, due to the lack of global optimality in nonconvex optimization, the statistical properties of the solutions obtained by existing algorithms remain rather difficult to analyze.

Recently, Cai et al. (2016) proposed a thresholded Wirtinger flow (TWF) algorithm to tackle the problem, which essentially employs proximal-type iterations. TWF starts from a carefully specified initial point, and iteratively performs gradient descent steps. In particular, at each iteration, TWF performs a thresholding step to preserve the sparsity of the solution. Cai et al. (2016) further prove that TWF achieves a linear rate of convergence to an approximate global minimum that has optimal statistical accuracy. Note that Cai et al. (2016) can establish such strong theoretical results because their algorithm and analysis exploit the underlying "true" data generating process of sparse phase retrieval, which enables bypassing the computational hardness barrier. In specific, they assume that the response $Y \in \mathbb{R}$ and the covariate $\boldsymbol{X} \in \mathbb{R}^p$ satisfy

$$\boldsymbol{X} \sim N(0, \mathbf{I}_p) \quad \text{and} \quad Y = (\boldsymbol{X}^\top \boldsymbol{\beta}^*)^2 + \epsilon, \tag{1.2}$$

in which $\epsilon \sim N(0, \sigma^2)$ and $\|\boldsymbol{\beta}^*\|_0 \leq s$. The data points $\{(y^{(i)}, \boldsymbol{x}^{(i)}) \in \mathbb{R} \times \mathbb{R}^p\}_{i \in [n]}$ are $n$ independent realizations of $(Y, \boldsymbol{X})$, and $n$ needs to be sufficiently large, for example, $n = \Omega(s^2 \log p)$. Intuitively speaking, these distributional and scaling assumptions essentially rule out many worst-case instances of (1.1), and hence ease the theoretical analysis. Such a statistical view of nonconvex optimization, however, heavily relies on the correct model specification that the data points are indeed generated by the model defined in (1.2). To the best of our knowledge, we are not aware of any existing results on nonconvex optimization under misspecification of the underlying statistical models.

In this paper, we propose a simple variant of the TWF algorithm to tackle the problem in (1.1) under statistical model misspecification. In specific, we prove that the proposed algorithm is robust to certain model misspecifications, and achieves a linear convergence rate to an approximate global minimum with optimal statistical accuracy. To be more precise, we focus our discussion on the single index model (SIM), which assumes that the response variable $Y \in \mathbb{R}$ and the covariate $\boldsymbol{X} \sim N(0, \mathbf{I}_p)$ are linked through

$$Y = h(\boldsymbol{X}^\top \boldsymbol{\beta}^*, \epsilon), \tag{1.3}$$

in which $\boldsymbol{\beta}^*$ is the parameter of interest, $\epsilon$ is the random noise, and $h\colon \mathbb{R}\times\mathbb{R}\to\mathbb{R}$ is a link function. Existing work on phase retrieval requires $h$ to be known and the entire model to be correctly specified. Interesting examples of $h$ include $h(u,v)=u^2+v$ as in (1.2), or $h(u,v)=|u|+v$ and $h(u,v)=|u+v|$, where, if the model is correctly specified, the data points are generated by

$$Y=|\boldsymbol{X}^\top\boldsymbol{\beta}^*|+\epsilon \quad\text{and}\quad Y=|\boldsymbol{X}^\top\boldsymbol{\beta}^*+\epsilon|, \tag{1.4}$$

respectively. In contrast with existing work, here we do not assume any specific form of $h$. The only assumption that we impose on $h$ is that $Y$ and $(\boldsymbol{X}^\top\boldsymbol{\beta}^*)^2$ have non-null covariance, i.e.,

$$\mathrm{Cov}\big[Y,(\boldsymbol{X}^\top\boldsymbol{\beta}^*)^2\big]\neq 0, \tag{1.5}$$

The condition in (1.5) is quite mild and encompasses all aforementioned examples. Perhaps surprisingly, for this highly misspecified statistical model, we prove that the proposed variant of TWF still achieves a linear rate of convergence to an approximate global minimum that has optimal statistical accuracy. Thus, our results lead to new understandings of the robustness of nonconvex optimization algorithms.

Note that the model in (1.3) is not identifiable, because the norm of $\boldsymbol{\beta}^*$ can be absorbed into the unknown function $h$ while preserving the same value of $Y$. We henceforth assume that $\|\boldsymbol{\beta}^*\|_2=1$ to make the model identifiable. Meanwhile, note that the moment condition in (1.5) remains the same if we replace $\boldsymbol{\beta}^*$ by $-\boldsymbol{\beta}^*$. Hence, even under the assumption that $\|\boldsymbol{\beta}^*\|_2=1$, the model in (1.2) is only identifiable up to the global sign of $\boldsymbol{\beta}^*$. Similar phenomenon also appears in the classical phase retrieval model in (1.2).

**Major Contributions.** Our contributions include:

- We propose a variant of the thresholded Wirtinger flow algorithm for misspecified phase retrieval, which reduces to the TWF algorithm in Cai et al. (2016) when the true model is correctly specified as in (1.2).

- We establish unified computational and statistical results for the proposed algorithm. In specific, initialized with a thresholded spectral estimator, the proposed algorithm converges linearly to an estimator with optimal statistical accuracy. Our theory illustrates the robustness of TWF, which sheds new light on the wide empirical success of nonconvex optimization algorithms on real problems with possible model misspecification (Li and Duan, 1989).

Recent advance on nonconvex statistical optimization largely depends on the correct specification of statistical models (Wang et al., 2014a,b; Gu et al., 2014; Zhao et al., 2017; Yi et al., 2015; Yang et al., 2015; Tan et al., 2016; Loh and Wainwright, 2015; Sun et al., 2015; Li et al., 2016, 2017). To the best of our knowledge, this paper establishes the first nonconvex optimization algorithm and theory that incorporate model misspecification.

**Related Work.** The problem of recovering a signal from the magnitude of its linear measurements has broad applications, including electron microscopy (Coene et al., 1992), optical imaging (Shechtman et al., 2015), and X-ray crystallography (Millane, 1990; Marchesini et al., 2003). There exists a large body of literature on the statistical and computational aspects of phase retrieval. In specific, Candès et al. (2013, 2015a); Waldspurger et al. (2015) establish efficient algorithms based upon semidefinite programs. For nonconvex methods, Netrapalli et al. (2013); Waldspurger (2016) propose alternating minimization algorithms and Sun et al. (2016) develop a trust-region algorithm. Also, Candès et al. (2015b) establish the Wirtinger flow algorithm, which performs gradient descent on the nonconvex loss function. Such an algorithm is then further extended by Chen and Candès (2015); Zhang et al. (2016) to even more general settings. Furthermore, for sparse phase retrieval in high dimensions, Cai et al. (2016) propose the truncated Wirtinger flow algorithm and prove that it achieves the optimal statistical rate of convergence under correct model specification. For a more detailed survey, we refer interested readers to Jaganathan et al. (2015) and the references therein.

Our work is also closely related to SIM, which is extensively studied in statistics. See, for example, Han (1987); McCullagh and Nelder (1989); Horowitz (2009) and the references therein. Most of this line of work estimates the parameter $\boldsymbol{\beta}^*$ and the unknown link function jointly through $M$-estimation. However, these $M$-estimators are often formulated as the global minima of nonconvex optimization problems, and hence are often computationally intractable to obtain. A more related line of research is sufficient dimension reduction, which assumes that $Y$ only depends on $\boldsymbol{X}$ through the projection of $\boldsymbol{X}$ onto a subspace $\mathcal{U}$. SIM falls into such a framework with $\mathcal{U}$ being the subspace spanned by $\boldsymbol{\beta}^*$. See Li and Duan (1989); Li (1991); Cook and Ni (2005) and the references therein for details. Most work in this direction proposes spectral methods based on the conditional covariance of $\boldsymbol{X}$ given $Y$, which is difficult to estimate consistently in high dimensions. In fact, the moment condition in (1.5) is inspired by Li (1992), which studies the case with $\mathbb{E}(Y\boldsymbol{X}^\top\boldsymbol{\beta}^*) = 0$. Note that this condition also holds for phase retrieval models.

Moreover, the recent success of high-dimensional regression (Bühlmann and van de Geer, 2011) also sparks the study of SIMs in high dimensions. Plan et al. (2014); Plan and Vershynin (2016); Thrampoulidis et al. (2015); Goldstein et al. (2016); Genzel (2017) propose to estimate the direction of $\boldsymbol{\beta}^*$ using least squares regression with $\ell_1$-regularization. They show that the $\ell_1$-regularized estimator enjoys optimal statistical rate of convergence. However, their method depends on the pivotal condition that $\mathrm{Cov}(Y, \boldsymbol{X}^\top\beta^*) \neq 0$, which is not satisfied by the phase retrieval model. A more related work is Neykov et al. (2016), which propose a two-step estimator based on convex optimization. Specifically, their estimator is constructed via refining the solution of a semidefinite program by $\ell_1$-regularized regression. Compared with our method, their estimator incurs higher computational cost. In addition, Yang et al. (2017a,b) propose efficient estimators for high-dimensional SIMs with non-Gaussian covariates based on convex optimization. Furthermore, Jiang and Liu (2014); Lin et al. (2015); Zhu et al. (2006) extend sliced inverse regression (Li, 1991) to the high-dimensional setting. However, they mainly focus on support recovery and consistency properties. Moreover, there is a line of work focussing on estimating both the parametric and the nonparametric component (Kalai and Sastry, 2009; Kakade et al., 2011; Alquier and Biau, 2013; Radchenko, 2015). These methods are either not applicable to our model or have suboptimal rates.

In summary, although there is a large body of related literature, the success of existing nonconvex optimization algorithms for high-dimensional phase retrieval largely relies on the correct specification of the underlying generative models. By establishing connections with SIM, we propose and analyze a simple variant of TWF, which is provable robust to misspecified models.

**Notation.** For an integer $m$, we use $[m]$ to denote the set $\{1, 2, \ldots, m\}$. Let $S \subseteq [m]$ be a set, we define $S^c = [m]\backslash S$ as its complement. Let $\boldsymbol{x} \in \mathbb{R}^p$ be a vector, we denote by $\boldsymbol{x}_i$ the $i$-th coordinate of $\boldsymbol{x}$. We also take $\boldsymbol{x}_S$ as the projection of $\boldsymbol{x}$ onto the index set $S$, i.e., $[\boldsymbol{x}_S]_i = \boldsymbol{x}_i \cdot \mathbb{1}(i \in S)$ for all $i \in [p]$. Here $[\boldsymbol{x}_S]_i$ is the $i$-th entry of $\boldsymbol{x}_S$. Furthermore, we denote by $\|\boldsymbol{x}\|_0$ the number of non-zero entries of $\boldsymbol{x}$, that is, $\|\boldsymbol{x}\|_0 = \sum_{j \in [p]} \mathbb{1}\{\boldsymbol{x}_j \neq 0\}$. We denote the support of $\boldsymbol{x}$ to be $\mathrm{supp}(\boldsymbol{x})$, which is defined as $\{j : \boldsymbol{x}_j \neq 0\}$. Hence, we have that $\|\boldsymbol{x}\|_0 = |\mathrm{supp}(\boldsymbol{x})|$. We denote by $\|\cdot\|$ the $\ell_2$-norm of a vector, by $\|\cdot\|_2$ the spectral (operator) norm of a matrix, and by $\|\cdot\|_{\max}$ the elementwise $\ell_\infty$-norm of a matrix. We also employ $\|\cdot\|_\alpha$ to denote the $\ell_\alpha$-norm of a vector for any $\alpha \geq 1$. Furthermore, let $\alpha, \beta \in [1, \infty)$, we use $\|\cdot\|_{\alpha \to \beta}$ to denote the induced operator norm of a matrix, i.e., let $\boldsymbol{A} \in \mathbb{R}^{m \times n}$, then $\|\boldsymbol{A}\|_{\alpha \to \beta} = \sup_{\boldsymbol{x} \neq \boldsymbol{0}} \{\|\boldsymbol{A}\boldsymbol{x}\|_\beta / \|\boldsymbol{x}\|_\alpha\}$. For two random variables $X$ and $Y$, we write $X \perp\!\!\!\perp Y$ when $X$ is independent of $Y$. We use $\mathrm{Var}(X)$ to denote the variance of $X$, and use $\mathrm{Cov}(X, Y)$ to denote the covariance between random variables $X$ and $Y$. Also, let $\{a_n\}_{n \geq 0}$ and $\{b_n\}_{n \geq 0}$ be two positive sequences. If there exists some constant $C$ such that $\limsup_{n \to \infty} a_n / b_n \leq C$, we write $a_n = \mathcal{O}(b_n)$, or equivalently, $b_n = \Omega(a_n)$. We write $a_n = o(b_n)$ if $\lim_{n \to \infty} a_n / b_n = 0$.

**Definition 1.1** (Sub-exponential variable and $\psi_1$-norm)**.** *A random variable $X \in \mathbb{R}$ is called sub-exponential if its $\psi_1$-norm is bounded. The $\psi_1$-norm of $X$ is defined as $\|X\|_{\psi_1} = \sup_{p \geq 1} p^{-1} (\mathbb{E}|X|^p)^{1/p}$.*

# 2 Nonconvex Optimization for Misspecified Phase Retrieval

We consider the phase retrieval problem under model misspecification. Specifically, given the covariate $\boldsymbol{X} \in \mathbb{R}^p$ and the signal parameter $\boldsymbol{\beta}^* \in \mathbb{R}^p$, we assume that the response variable $Y$ is given by the single index model in (1.3) for some unknown link function $h : \mathbb{R} \times \mathbb{R} \to \mathbb{R}$, where $\epsilon$ is the random noise which is assumed to be independent of $\boldsymbol{X}$. Throughout our discussion, for simplicity, we assume each covariate is sampled from $N(0, \mathbf{I}_p)$. Since the norm of $\boldsymbol{\beta}^*$ can be incorporated into $h$, we impose an additional constraint $\|\boldsymbol{\beta}^*\| = 1$, in order to make the model partially identifiable. Our goal is to estimate $\boldsymbol{\beta}^*$ using $n$ i.i.d. realizations of $(\boldsymbol{X}, Y)$, where we denote the set of $n$ samples by $\{(\boldsymbol{x}^{(i)}, y^{(i)})\}_{i \in [n]}$, and we do not have the knowledge of $h$ in advance. Throughout this paper, we consider a high-dimensional and sparse regime where $\boldsymbol{\beta}^*$ has at most $s$ non-zero entries for some $s < n$, and the dimensionality $p$ can be much larger than the sample size $n$.

## 2.1 Motivation

Our modified TWF algorithm is inspired by the following optimization problem

$$\overline{\boldsymbol{\beta}} = \operatorname*{argmin}_{\boldsymbol{\beta} \in \mathbb{R}^p} \mathrm{Var}\big[Y - (\boldsymbol{X}^\top \boldsymbol{\beta})^2\big], \tag{2.1}$$

which aims at estimating $\boldsymbol{\beta}^*$ by minimizing the variability of $Y - (\boldsymbol{X}^\top \boldsymbol{\beta})^2$. Intuitively, the variability would be minimized by a vector $\boldsymbol{\beta}$ which is parallel to $\boldsymbol{\beta}^*$. To see this, first note that $Y$ depends on $\boldsymbol{X}$ only through $\boldsymbol{X}^\top \boldsymbol{\beta}^*$. If $\boldsymbol{\beta}$ is not parallel to $\boldsymbol{\beta}^*$, $\boldsymbol{X}^\top \boldsymbol{\beta}$ would contain a component which is independent of $\boldsymbol{X}^\top \boldsymbol{\beta}^*$ and $Y$, which increases the variability of $Y - (\boldsymbol{X}^\top \boldsymbol{\beta})^2$. The following proposition justifies this intuition by showing that (2.1) has two global minima which are both parallel to $\boldsymbol{\beta}^*$. This proposition forms the basis for our modified version of the TWF algorithm.

**Proposition 2.1.** *Suppose* $\mathrm{Cov}[Y, (\boldsymbol{X}^\top \boldsymbol{\beta}^*)^2] = \rho > 0$. *Let* $\overline{\boldsymbol{\beta}}$ *be defined in* (2.1). *Then we have* $\overline{\boldsymbol{\beta}} = \boldsymbol{\beta}^* \cdot \sqrt{\rho/2}$ *or* $\overline{\boldsymbol{\beta}} = -\boldsymbol{\beta}^* \cdot \sqrt{\rho/2}$.

*Sketch of the Proof.* We first decompose $\boldsymbol{\beta}$ into $\boldsymbol{\beta} = \zeta \boldsymbol{\beta}^* + \boldsymbol{\beta}^\perp$, where $\zeta = \boldsymbol{\beta}^\top \boldsymbol{\beta}^*$ and $\boldsymbol{\beta}^\perp$ is perpendicular to $\boldsymbol{\beta}^*$. Since $\boldsymbol{X} \sim N(0, \mathbf{I}_p)$, we have $\boldsymbol{X}^\top \boldsymbol{\beta}^* \perp\!\!\!\perp \boldsymbol{X}^\top \boldsymbol{\beta}^\perp$ and $Y \perp\!\!\!\perp \boldsymbol{X}^\top \boldsymbol{\beta}^\perp$. In addition, the norm of $\boldsymbol{\beta}^\perp$ is given by $\|\boldsymbol{\beta}^\perp\|_2^2 = \|\boldsymbol{\beta}\|_2^2 - \zeta^2$. By expanding the variance term in (2.1) and some simple manipulations, we obtain

$$\mathrm{Var}\big[Y - (\boldsymbol{X}^\top \boldsymbol{\beta})^2\big] = \mathrm{Var}(Y) - 2\zeta^2 \rho + 2\|\boldsymbol{\beta}\|^4. \tag{2.2}$$

To find the minimizer of the right-hand side in (2.2), first we note that $\boldsymbol{\beta} = \mathbf{0}$ cannot be a minimizer. To see this, by direct computation, if we let $\widetilde{\boldsymbol{\beta}} = c\boldsymbol{\beta}^*$ with $0 < c < \sqrt{\rho}$, it can be shown that

$$\mathrm{Var}\big[Y - (\boldsymbol{X}^\top \widetilde{\boldsymbol{\beta}})^2\big] < \mathrm{Var}\big[Y - (\boldsymbol{X}^\top \mathbf{0})^2\big].$$

Now we fix $\|\boldsymbol{\beta}\| > 0$ as a constant. Then $|\zeta| \le \|\boldsymbol{\beta}\|$ by definition. By (2.2), $\mathrm{Var}[Y - (\boldsymbol{X}^\top \boldsymbol{\beta})^2]$ can be viewed as a function of $\zeta$. The minimum is achieved only when $\zeta^2 = \|\boldsymbol{\beta}\|^2$, which implies $\boldsymbol{\beta}^\perp = \mathbf{0}$. In other words, any minimizer $\overline{\boldsymbol{\beta}}$ given by (2.2) is parallel to $\boldsymbol{\beta}^*$.

Now we set $|\zeta| = \|\boldsymbol{\beta}\|$ in the right-hand side of (2.2), which now becomes a function of $\|\boldsymbol{\beta}\|$. We minimize it with respect to $\|\boldsymbol{\beta}\|$ and the minimizer is given by $\|\boldsymbol{\beta}\| = \sqrt{\rho/2}$. Therefore, we prove that the minima of $\mathrm{Var}[Y - (\boldsymbol{X}^\top \boldsymbol{\beta})^2]$ are $\boldsymbol{\beta}^* \sqrt{\rho/2}$ and $-\boldsymbol{\beta}^* \sqrt{\rho/2}$, which concludes the proof. See §B.1 for a detailed proof. $\square$

Proposition 2.1 suggests that estimating $\boldsymbol{\beta}^*$ up to a proportionality constant is feasible by minimizing (2.1) even without the knowledge of $h$. Specifically, it suggests that $\mathrm{Var}[Y - (\boldsymbol{X}^\top \boldsymbol{\beta})^2]$ is a reasonable loss function on the "population level" (i.e., with infinite samples) for the class of SIMs satisfying the moment condition in (1.5). In addition, we note that the assumption that $\rho > 0$ is not essential. The reason is, if $\rho < 0$, one could apply the same logic to $-Y$. That is, estimate $\boldsymbol{\beta}^*$ by minimizing $\mathrm{Var}[Y + (\boldsymbol{X}^\top \boldsymbol{\beta})^2]$. In the following, we explain the reason why it is preferable to use $\mathrm{Var}[Y - (\boldsymbol{X}^\top \boldsymbol{\beta})^2]$ over $\mathbb{E}[Y - (\boldsymbol{X}^\top \boldsymbol{\beta})^2]^2$ as a population loss function.

As we discussed in the introduction, the original motivation of the TWF algorithm is based on the likelihood perspective, which leads to the least squares loss function, i.e., $\mathbb{E}[Y - (\boldsymbol{X}^\top \boldsymbol{\beta})^2]^2$. When model (1.2) is correctly specified, and the random noise $\epsilon$ has mean zero, the least squares loss function coincides with the variance loss in (2.1). However, in cases where the expectation of $\epsilon$ is negative, the minimizer of the least squares loss function could be $\boldsymbol{\beta} = 0$. In this case, minimizing the least squares loss function will lead to uninformative results. In addition, it is not hard to see that the original TWF algorithm of Cai et al. (2016) is not well-defined in cases where $\mathbb{E}(Y) \leq 0$. Cai et al. (2016) do not exhibit this problem since they assume that model (1.2) is correctly specified, and that $\epsilon$ is centered at 0. These results implies that $\mathbb{E}(Y)$ is positive. In summary, the variance loss in (2.1) is more appropriate for robust estimation over the misspecified phase retrieval models.

In addition to relating $\boldsymbol{\beta}^*$ to the minimizer of the variance loss function, in the following proposition, we show that the moment condition in (1.5) implies spectral method can be used to estimate $\boldsymbol{\beta}^*$ and $\rho$ simultaneously.

**Proposition 2.2.** *Assuming* $\mathrm{Cov}[Y, (\boldsymbol{X}^\top \boldsymbol{\beta}^*)^2] = \rho$, *we have*

$$\mathbb{E}[(Y - \mu)\boldsymbol{X}\boldsymbol{X}^\top] = \mathbb{E}[Y(\boldsymbol{X}\boldsymbol{X}^\top - \mathbf{I}_p)] = \rho \cdot \boldsymbol{\beta}^* \boldsymbol{\beta}^{*\top}, \tag{2.3}$$

*where* $\mu = \mathbb{E}(Y)$. *Thus* $\pm\boldsymbol{\beta}^*$ *are the eigenvector of* $\mathbb{E}[(Y - \mu)\boldsymbol{X}\boldsymbol{X}^\top]$ *corresponding to eigenvalue* $\rho$, *and* $\rho$ *is the largest eigenvalue in magnitude.*

*Proof of Proposition 2.2.* The first identity follows from direct calculation. For any $\mathbf{v} \in \mathbb{R}^p$, we decompose $\mathbf{v}$ into $\mathbf{v} = \zeta\boldsymbol{\beta}^* + \boldsymbol{\beta}^\perp$, where $\zeta = \mathbf{v}^\top \boldsymbol{\beta}^*$ and $\boldsymbol{\beta}^\perp$ is perpendicular to $\boldsymbol{\beta}^*$. Since $\boldsymbol{X} \sim N(0, \mathbf{I}_p)$, $\boldsymbol{X}^\top \boldsymbol{\beta}^\perp$ is independent of $\boldsymbol{X}^\top \boldsymbol{\beta}^*$ and $Y$. By direct computation, we have

$$\mathbb{E}[(Y - \mu) \cdot \mathbf{v}^\top \boldsymbol{X}\boldsymbol{X}^\top \mathbf{v}] = \zeta^2 \rho = \rho \mathbf{v}^\top \boldsymbol{\beta}^* \boldsymbol{\beta}^{*\top} \mathbf{v}, \tag{2.4}$$

which establishes (2.3). In addition, (2.3) implies that $\rho$ is the eigenvalue of $\mathbb{E}[(Y - \mu)\boldsymbol{X}\boldsymbol{X}^\top]$ with the largest magnitude, and $\pm\boldsymbol{\beta}^*$ are the eigenvector corresponding to eigenvalue $\rho$. This concludes the proof of this proposition.

□

As shown in Proposition 2.2, an alternative way of estimating $\rho$ and $\boldsymbol{\beta}^*$ is to estimate the largest eigenvalue of $\mathbb{E}[(Y - \mu)\boldsymbol{X}\boldsymbol{X}^\top]$ in magnitude and the corresponding eigenvector. In the next section, we establish the sample versions of the optimization problems outlined by Propositions 2.1 and 2.2. These two problems correspond to the two stages of our modified version of the TWF algorithm, respectively.

## 2.2 Thresholded Wirtinger Flow Revisited

In this section, we present the modified TWF algorithm for misspecified phase retrieval. Note that In §2.1, we propose to use $\mathrm{Var}[Y - (\boldsymbol{X}^\top \boldsymbol{\beta})^2]$ as a more robust population loss function in comparison with the least squares loss function $\mathbb{E}[Y - (\boldsymbol{X}^\top \boldsymbol{\beta})^2]^2$ used by the classical Wirtinger Flow algorithm in (Candès et al., 2015b). In the following, we first establish the sample version of the minimization problem in (2.1).

Recall that we assume that $\boldsymbol{\beta}^*$ has at most $s$ non-zero entries. For the time being, suppose that $\rho > 0$, which will be immediately relaxed in §2.3. Given $\boldsymbol{X} \sim N(\mathbf{0}, \mathbf{I}_p)$ and $\mu = \mathbb{E}(Y)$, we have

$$\mathbb{E}[Y - (\boldsymbol{X}^\top \boldsymbol{\beta})^2] = \mu - \boldsymbol{\beta}^\top \mathbb{E}(\boldsymbol{X}\boldsymbol{X}^\top)\boldsymbol{\beta} = \mu - \|\boldsymbol{\beta}\|^2. \tag{2.5}$$

The sample version of $\mu - \|\boldsymbol{\beta}\|^2$ is defined by $\xi_n(\boldsymbol{\beta}) = \mu_n - \|\boldsymbol{\beta}\|^2$, where $\mu_n = n^{-1}\sum_{i=1}^n y^{(i)}$ is the sample mean of the response variables. We consider the following nonconvex optimization problem

$$\widehat{\boldsymbol{\beta}} = \operatorname*{argmin}_{\|\boldsymbol{\beta}\|_0 \le s} \ell_n(\boldsymbol{\beta}), \quad \text{where} \quad \ell_n(\boldsymbol{\beta}) = \frac{1}{n}\sum_{i=1}^n \big[y^{(i)} - (\boldsymbol{x}^{(i)\top}\boldsymbol{\beta})^2 - \xi_n(\boldsymbol{\beta})\big]^2. \tag{2.6}$$

Here the loss function $\ell_n(\boldsymbol{\beta})$ is the sample version of the variance loss function $\mathrm{Var}[Y - (\boldsymbol{X}^\top\boldsymbol{\beta})^2]$. In order to solve (2.6) efficiently, similar to TWF algorithm in Cai et al. (2016), we propose a gradient-based algorithm for the loss function $\ell_n(\boldsymbol{\beta})$. In addition, due to the cardinality constraint in (2.6), in each step, we apply a thresholding operator to the estimate after a gradient descent step. Note that the constraint in (2.6) involves $s = \|\boldsymbol{\beta}^*\|_0$, which is usually unknown. By selecting the threshold value adaptively, our proposed TWF algorithm actually does not require any prior knowledge of $s$, and can iteratively determine the sparsity of the estimate. Specifically, at the $k$-th iteration, we compute a threshold value

$$\tau(\boldsymbol{\beta}^{(k)}) = \kappa\bigg\{\frac{\log(np)}{n^2}\sum_{i=1}^n \Big[y^{(i)} - \big(\boldsymbol{x}^{(i)\top}\boldsymbol{\beta}^{(k)}\big)^2 - \mu_n + \|\boldsymbol{\beta}^{(k)}\|^2\Big]^2 \big(\boldsymbol{x}^{(i)\top}\boldsymbol{\beta}^{(k)}\big)^2\bigg\}^{1/2}, \tag{2.7}$$

where $\kappa$ is an appropriately chosen constant. We then take a thresholded gradient descent step

$$\boldsymbol{\beta}^{(k+1)} = \mathcal{T}_{\eta\cdot\tau(\boldsymbol{\beta}^{(k)})}\big[\boldsymbol{\beta}^{(k)} - \eta\nabla\ell_n(\boldsymbol{\beta}^{(k)})\big], \tag{2.8}$$

where $\eta > 0$ is the step size and $[\mathcal{T}_\tau(\boldsymbol{w})]_j = \boldsymbol{w}_j \cdot \mathbb{1}(|\boldsymbol{w}_j| \ge \tau)$ is the hard-thresholding operator with threshold value $\tau > 0$. The gradient $\nabla\ell_n(\cdot)$ in (2.8) is given by

$$\nabla\ell_n(\boldsymbol{\beta}) = \frac{4}{n}\sum_{i=1}^n \big[y^{(i)} - (\boldsymbol{x}^{(i)\top}\boldsymbol{\beta})^2 - \xi_n(\boldsymbol{\beta})\big]\big(\mathbf{I}_p - \boldsymbol{x}^{(i)}\boldsymbol{x}^{(i)\top}\big)\boldsymbol{\beta}. \tag{2.9}$$

After running the algorithm for $T$ iterations for some sufficiently large $T$, the last estimate $\boldsymbol{\beta}^{(T)}$ is standardized by $\boldsymbol{\beta}^{(T)}/\|\boldsymbol{\beta}^{(T)}\|$, and we take it as the final estimate of $\boldsymbol{\beta}^*$ (recall that we assume $\|\boldsymbol{\beta}^*\| = 1$). We note that the gradient step in (2.8) requires starting from an proper initialization $\boldsymbol{\beta}^{(0)}$. In the following section, we present a thresholded spectral algorithm to obtain a good initializer.

## 2.3 Initialization via Thresholded Spectral Method

Motivated by Proposition 2.2, we introduce the thresholded spectral method (TSM) to obtain an appropriate initialization for the TWF algorithm. Note that $\boldsymbol{\beta}^*$ is the leading eigenvector of $\mathbb{E}[(Y-\mu)\boldsymbol{X}\boldsymbol{X}^\top]$ when $\rho$ is positive. A natural idea is to apply spectral methods to the empirical counterpart of matrix $\mathbb{E}[(Y-\mu)\boldsymbol{X}\boldsymbol{X}^\top]$ based on the samples. However, since $\mathbb{E}[(Y-\mu)\boldsymbol{X}\boldsymbol{X}^\top]$ is a high-dimensional matrix, the leading eigenvector this sample matrix can have large estimation error. To resolve this issue, we apply spectral methods on a submatrix selected by the following screening step.

Specifically, we first conduct a thresholding step to select a subset of coordinates $\widehat{S}_0 \subseteq [p]$ by

$$\widehat{S}_0 = \left\{ j \in [p] : \left| \frac{1}{n} \sum_{i=1}^{n} y^{(i)} [(\boldsymbol{x}_j^{(i)})^2 - 1] \right| > \gamma \sqrt{\log(np)/n} \right\}, \tag{2.10}$$

where $\gamma$ is an appropriately chosen constant. The thresholding step in (2.10) is motivated by Proposition 2.2, which shows that the diagonal entries of the matrix $\mathbb{E}[Y(\boldsymbol{X}\boldsymbol{X}^\top - \mathbf{I})]$ are non-zero on the support of $\boldsymbol{\beta}^*$. When $n$ is sufficiently large, $n^{-1}\sum_{i=1}^n y^{(i)}[(\boldsymbol{x}_j^{(i)})^2 - 1]$ is close to its expectation for all $j \in [p]$. Thus one would expect that, with an appropriately chosen $\gamma > 0$, $|\mathbb{E}[Y \cdot (\boldsymbol{X}_j^2 - 1)]| > 0$ for all $j \in \widehat{S}_0$. This implies that the thresholding step constructs an $\widehat{S}_0$, which is a subset of the true support $\mathrm{supp}(\boldsymbol{\beta}^*)$. We remark that this step is closely related to the diagonal thresholding algorithm for sparse principle component analysis (Johnstone and Lu, 2009).

After obtaining $\widehat{S}_0$, to simplify the notation, we denote $\boldsymbol{x}_{\widehat{S}_0}^{(i)}$ by $\boldsymbol{w}^{(i)}$ for each $i \in [n]$. Recall that $\mu_n$ is the sample average of $y^{(1)}, \ldots, y^{(n)}$. We define $\boldsymbol{W} \in \mathbb{R}^{p \times p}$ by

$$\boldsymbol{W} = \frac{1}{n} \sum_{i=1}^{n} (y^{(i)} - \mu_n) \cdot \boldsymbol{w}^{(i)} (\boldsymbol{w}^{(i)})^\top. \tag{2.11}$$

Notice that the the $|\widehat{S}_0| \times |\widehat{S}_0|$ non-zero entries of $\boldsymbol{W}$ form the submatrix of $n^{-1}\sum_{i=1}^n (y^{(i)} - \mu_n) \cdot \boldsymbol{x}^{(i)}(\boldsymbol{x}^{(i)})^\top$ with both rows and columns in $\widehat{S}_0$. Let $\widehat{\boldsymbol{v}}$ be the the eigenvector of $\boldsymbol{W}$ corresponding to the largest eigenvalue in magnitude. Finally, we define the initial estimator $\boldsymbol{\beta}^{(0)}$ by

$$\boldsymbol{\beta}^{(0)} = \widehat{\boldsymbol{v}} \sqrt{|\rho_n|/2} \quad \text{where} \quad \rho_n = \frac{1}{n} \sum_{i=1}^{n} y^{(i)} \big(\boldsymbol{x}^{(i)\top} \widehat{\boldsymbol{v}}\big)^2 - \mu_n. \tag{2.12}$$

With this initial estimator, the algorithm proceeds with thresholded gradient descent defined in (2.8) and outputs the normalized estimator $\boldsymbol{\beta}^{(T)}/\|\boldsymbol{\beta}^{(T)}\|$. In the next section, we establish the statistical and computational rates of convergence for the TWF algorithm.

# 3 Statistical and Computational Guarantees

In this section, we show that the TWF algorithm is robust for misspecified phase retrieval models. In particular, we prove that the proposed algorithm linearly converges to an estimator, and the estimator achieves optimal statistical rate of convergence even under the unknown link function. Before going further, we first impose the following assumption on the distribution of $Y$ to facilitate our discussion.

**Assumption 3.1.** *The response variable $Y$ of the misspecified phase retrieval model in* (1.3) *follows a sub-exponential distribution as specified in Definition 1.1. Specifically, we assume that $\|Y\|_{\psi_1} \leq \Psi$ for some constant $\Psi > 0$. In addition, we assume that* $\mathrm{Cov}[Y, (\boldsymbol{X}^\top \boldsymbol{\beta}^*)^2] = \rho \neq 0$ *for some constant $\rho$.*

The assumption that $Y$ has sub-exponential tail is mild, and is satisfied if both the link function $h$ and the random noise $\epsilon$ are well-behaved. For example, if $\epsilon$ is a sub-exponential random variable, and $h(z,w)$ is Lipschitz continuous in both $z$ and $w$, this assumption is satisfied. Meanwhile, Assumption 3.1 also postulates that $Y$ has a non-null correlation with $(\boldsymbol{X}^\top\boldsymbol{\beta}^*)^2$. Note that in Proposition 2.1, we assume that $\rho > 0$. When $\rho$ is negative, we can still estimate $\boldsymbol{\beta}^*$ by applying (2.7) and (2.8) with $y^{(i)}$ replaced by $-y^{(i)}$ for all $i \in [n]$. As we show in the main result, the sign of $\rho$ is well estimated by the sign of $\rho_n$, which is a byproduct of our thresholded spectral method for initialization described in §2.3. Thus, a non-zero $\rho$ is sufficient for our TWF algorithm to recover the direction of $\boldsymbol{\beta}^*$ accurately.

To provide more insight on when $\rho \neq 0$ holds, we define the function $\varphi\colon \mathbb{R} \to \mathbb{R}$ by $\varphi(z) = \mathbb{E}_\epsilon[h(z,\epsilon)]$. Since $\boldsymbol{X}^\top\boldsymbol{\beta}^* \sim N(0,1)$, by the Stein's identity (Stein, 1981, Lemma 4), we have

$$\rho = \operatorname{Cov}\bigl[Y, (\boldsymbol{X}^\top\boldsymbol{\beta}^*)^2\bigr] = \mathbb{E}\bigl[\varphi(Z)\cdot(Z^2-1)\bigr] = \mathbb{E}\bigl[D^2\varphi(Z)\bigr],$$

where $Z \sim N(0,1)$, and $D^2\varphi\colon \mathbb{R} \to \mathbb{R}$ denotes the second-order distributional derivative[2] of $\varphi$. Thus, the non-zero assumption of $\rho$ essentially imposes certain smoothness condition on $\varphi$, and hence indirectly on $h$.

To further understand the TWF algorithm, we first provide more intuition on the initialization step. Recall that, as shown in Proposition 2.2, for all $j \in [p]$, we have

$$\mathbb{E}\bigl[Y(\boldsymbol{X}_j^2 - 1)\bigr] = \rho \cdot \boldsymbol{\beta}_j^{*2}. \tag{3.1}$$

Using standard concentration inequalities, we have that

$$\frac{1}{n}\cdot\sum_{i=1}^n y^{(i)}\cdot\bigl[(\boldsymbol{x}_j^{(i)})^2 - 1\bigr] = \frac{1}{n}\sum_{i=1}^n y^{(i)}(\boldsymbol{x}_j^{(i)})^2 - \mu_n$$

is close to its expected value $\rho\cdot\boldsymbol{\beta}_j^{*2}$ up to an additive error of order $\mathcal{O}(\sqrt{\log(np)/n})$, where $\mu_n$ is the sample average of $\{y^{(i)}\}_{i\in[n]}$. Thus, if for some $j \in [p]$, the magnitude of $\boldsymbol{\beta}_j^*$ is larger than $C\cdot[\log(np)/n]^{1/4}$ for some sufficiently large constant $C$, the signal term $\rho\cdot\boldsymbol{\beta}_j^{*2}$ dominates the additive error, which implies that $j$ is in $\widehat{S}_0$ defined in (2.10).

Thus, the thresholding step in (2.10) ensures that, with high probability, $\widehat{S}_0$ contains the effective support of $\boldsymbol{\beta}^*$, which is defined as the set of coordinates of $\boldsymbol{\beta}^*$ whose magnitudes are at least $\Omega([\log(np)/n]^{1/4})$. Furthermore, we show that the coordinates outside the effective support of $\boldsymbol{\beta}^*$ (i.e., the entries of $\boldsymbol{\beta}^*$ with smaller magnitudes) can be safely ignored in the initialization step, since they are indistinguishable from the error. In other words, the thresholding step can be viewed as applying the diagonal thresholding method in Johnstone and Lu (2009) to the empirical counterpart of $\mathbb{E}[Y\cdot(\boldsymbol{X}\boldsymbol{X}^\top - \mathbf{I}_p)]$.

More importantly, conducting the thresholding step in (2.10) before constructing $\boldsymbol{W}$ is indispensable. Since $|\widehat{S}_0| \le s$, after the thresholding step, the noise introduced to the spectral estimator is roughly proportional to $|\widehat{S}_0|$ linearly, which is significantly lower than the dimension $p$. This allows us to obtain a proper initial estimator with $\mathcal{O}[s^2\log(np)]$ samples. On the other hand, since the dimensionality $p$ is much larger than the sample size $n$, the spectral initial estimator without the thresholding step can incur a large error which makes the TWF algorithm diverge.

[2] See, for example, Foucart and Rauhut (2013) for the definition of the distributional derivative.

In what follows, we characterize the estimation error of the initial estimator $\boldsymbol{\beta}^{(0)}$ obtained by the thresholded spectral method. For ease of presentation, in the initialization step, we assume that $\widehat{S}_0$, $\boldsymbol{W}$, and $\rho_n$ are constructed using independent samples. Specifically, assuming we have $3n$ i.i.d. samples $\{(y^{(i)}, \boldsymbol{x}^{(i)})\}_{i\in[3n]}$, we use the first $n$ samples to construct $\widehat{S}_0$ in (2.10), and use the next $n$ samples $\{(y^{(i)}, \boldsymbol{x}^{(i)})\}_{n+1\leq i\leq 2n}$ to construct $\boldsymbol{W}$ by (2.11). In addition, we let $\mu_n$ be the sample average of $y^{(n+1)}, \ldots, y^{(2n)}$; we denote $\boldsymbol{x}^{(i)}_{\widehat{S}_0}$ by $\boldsymbol{w}^{(i)}$ for each $i \in \{n+1, \ldots, 2n\}$, and we let $\widehat{\boldsymbol{v}}$ be the the leading eigenvector of $\boldsymbol{W}$ corresponding to the largest eigenvalue in magnitude. Finally, given $\widehat{\boldsymbol{v}}$, we use the last $n$ observations to construct $\rho_n$ and the initial estimator $\boldsymbol{\beta}^{(0)}$ as in (2.12). Such construction frees us from possibly complicated dependent structures and allows us to present a simpler proof. In practice, the construction can be achieved by data splitting. Since we only raise the sample size by a constant factor of 3, the statistical and computational rates of our proposed method do not compromise. Note that the additional $2n$ independent samples can be avoided with more involved analysis using the Cauchy interlacing theorem (Golub and Van Loan, 2012). Since our scope focuses on nonconvex optimization for misspecified models, we adopt such construction with independence to ease the presentation and analysis. Moreover, since we only raise the sample size by a constant factor of 3, the statistical and computational rates of our proposed method do not sacrifice. In the second step of the TWF algorithm, we still assume the sample size to be $n$ so as to be consistent with the description of the algorithm.

In the following lemma, we provide an error bound for the initial estimator $\boldsymbol{\beta}^{(0)}$ obtained by the thresholded spectral method in (2.12). To measure the estimation error, we define

$$\operatorname{dist}(\mathbf{u}, \mathbf{v}) = \min\left(\|\mathbf{u} - \mathbf{v}\|, \|\mathbf{u} + \mathbf{v}\|\right)$$

for any $\mathbf{u}, \mathbf{v} \in \mathbb{R}^p$. As can be seen, this measure is invariant to the sign of $\boldsymbol{\beta}^*$, which is not identifiable. That is, for any $\mathbf{u}, \mathbf{v} \in \mathbb{R}^p$, we have $\operatorname{dist}(\mathbf{u}, \mathbf{v}) = \operatorname{dist}(\mathbf{u}, -\mathbf{v})$. Note that we provide an outline of the proof here and defer the detailed arguments to §A.1.

**Lemma 3.1.** *Suppose Assumption 3.1 holds. Let $\boldsymbol{\beta}^{(0)}$ be the initial solution obtained by the thresholded spectral method defined in (2.12). Recall that the constant $\Psi$ is specified in Assumption 3.1. For any constant $\alpha \in (0, 1]$, given large enough $n$ satisfying $n \geq Cs^2 \log(np)$, where $C$ is a generic constant depending only on $\Psi$ and $\alpha$, we have*

$$|\rho_n - \rho| \leq \alpha \cdot |\rho|, \quad \operatorname{supp}(\boldsymbol{\beta}^{(0)}) \subseteq \operatorname{supp}(\boldsymbol{\beta}^*), \quad \textit{and} \;\; \operatorname{dist}(\boldsymbol{\beta}^{(0)}, \overline{\boldsymbol{\beta}}) \leq \alpha \cdot \|\overline{\boldsymbol{\beta}}\|,$$

*where $\overline{\boldsymbol{\beta}}$ is the minimizer of $\operatorname{Var}[Y - (\boldsymbol{X}^\top \boldsymbol{\beta})^2]$ defined in (2.1), with probability at least $1 - \mathcal{O}(1/n)$.*

*Proof Sketch of Lemma 3.1.* The proof contains two steps. In the first step, we show that $\widehat{S}_0 \subseteq \operatorname{supp}(\boldsymbol{\beta}^*)$ with high probability. For each $j \in [p]$, we define

$$I_j = \frac{1}{n}\sum_{i=1}^{n} y^{(i)} \cdot (\boldsymbol{x}_j^{(i)})^2.$$

Then $\widehat{S}_0$ in (2.10) can be written as $\widehat{S}_0 = \{j \in [p] \colon |I_j| > \gamma\sqrt{\log(np)/n}\}$. Note that, for any $j \in [p]$, we have $\mathbb{E}[Y \cdot (X_j^2 - 1)] = \rho\boldsymbol{\beta}_j^{*2}$. By the law of large numbers, we expect $I_j$ to be close to zero for all $j \notin \operatorname{supp}(\boldsymbol{\beta}^*)$. Using concentration inequalities, we show that with probability at least $1 - \mathcal{O}(1/n)$, for any $j \in [p]$, we obtain

$$|I_j - \rho \cdot \boldsymbol{\beta}_j^{*2}| < \gamma\sqrt{\log(np)/n}.$$

Thus, for any $j \notin \operatorname{supp}(\boldsymbol{\beta}^*)$, we have $|I_j| < \gamma\sqrt{\log(np)/n}$ with probability at least $1-\mathcal{O}(1/n)$, which implies that $j \notin \widehat{S}_0$. Thus we conclude that $\widehat{S}_0 \subseteq \operatorname{supp}(\boldsymbol{\beta}^*)$ with high probability.

In addition, by similar arguments, it is not difficult to see that, if $\boldsymbol{\beta}_j^{*2} \geq C_1\sqrt{\log(np)/n}$ with $C_1 \geq 2\gamma/|\rho|$, we have, by the triangle inequality,

$$|I_j| \geq |\rho \cdot \boldsymbol{\beta}_j^{*2}| - |I_j - \rho \cdot \boldsymbol{\beta}_j^{*2}| \geq \gamma\sqrt{\log(np)/n}$$

with high probability. This implies that $j \in \widehat{S}_0$ high probability. Thus, $\widehat{S}_0$ captures all entries $j \in \operatorname{supp}(\boldsymbol{\beta}^*)$ for which $\boldsymbol{\beta}_j^{*2}$ is sufficiently large. Then we obtain that

$$\left\|\boldsymbol{\beta}^*_{\widehat{S}_0} - \boldsymbol{\beta}^*\right\|^2 = \mathcal{O}\Big[s\sqrt{\log(np)/n}\Big], \tag{3.2}$$

where $\boldsymbol{\beta}^*_{\widehat{S}_0}$ is the restriction of $\boldsymbol{\beta}^*$ on $\widehat{S}_0$.

In the second step, we show that the eigenvector $\widehat{\boldsymbol{v}}$ of $\boldsymbol{W}$ is a good approximation of $\boldsymbol{\beta}^*_{\widehat{S}_0}$. Recall that we denote $\boldsymbol{x}^{(i)}_{\widehat{S}_0}$ by $\boldsymbol{w}^{(i)}$ to simplify the notation. In addition, let $\boldsymbol{\beta}_0^* = \boldsymbol{\beta}^*_{\widehat{S}_0}/\|\boldsymbol{\beta}^*_{\widehat{S}_0}\|$. We can rewrite $\boldsymbol{W}$ as

$$\boldsymbol{W} = \frac{1}{n}\sum_{i=n+1}^{2n} (y^{(i)} - \mu) \cdot |\boldsymbol{\beta}_0^{*\top}\boldsymbol{w}^{(i)}|^2 \cdot \boldsymbol{\beta}_0^*\boldsymbol{\beta}_0^{*\top} + \boldsymbol{N},$$

where $\boldsymbol{N}$ can be viewed as an error matrix. We show that $\|\boldsymbol{N}\|_2 = \mathcal{O}(s\sqrt{\log(np)/n})$ with probability at least $1 - \mathcal{O}(1/n)$. Thus, for $n \geq Cs^2\log(np)$ with a sufficiently large constant $C$, $\widehat{\boldsymbol{v}}$ is close to $\boldsymbol{\beta}_0^*$ with high probability. Since $\|\boldsymbol{\beta}^*\| = 1$, (3.2) implies that $\boldsymbol{\beta}_0^*$ is close to $\boldsymbol{\beta}^*$. Thus, we show that $\widehat{\boldsymbol{v}}$ and $\boldsymbol{\beta}^*$ are close in the sense that

$$|\widehat{\boldsymbol{v}}^\top\boldsymbol{\beta}^*|^2 \geq 1 - C/|\rho| \cdot s\sqrt{\log(np)/n} \tag{3.3}$$

for some absolute constant $C$. Furthermore, since $\rho_n$ in (2.12) is calculated using samples independent of $\widehat{\boldsymbol{v}}$, by standard concentration inequalities, we have

$$\Big|\rho_n - \mathbb{E}\{Y \cdot [(\boldsymbol{X}^\top\widehat{\boldsymbol{v}})^2 - 1]\}\Big| = \Big|\rho_n - \rho \cdot |\widehat{\boldsymbol{v}}^\top\boldsymbol{\beta}^*|^2\Big| \leq \mathcal{O}\Big(\sqrt{\log n/n}\Big) \tag{3.4}$$

with probability at least $1 - \mathcal{O}(1/n)$. Hence, combining (3.3) and (3.4), we obtain that

$$|\rho_n - \rho| \leq \mathcal{O}\Big[s\sqrt{\log(np)/n}\Big] \leq \alpha|\rho| \tag{3.5}$$

for any fixed constant $\alpha \in (0,1)$. Here the last inequality holds when $n \geq Cs^2 \log(np)$ with constant $C$ sufficiently large. Finally, to bound $\mathrm{dist}(\boldsymbol{\beta}^{(0)}, \overline{\boldsymbol{\beta}})$, by direct calculation, we have

$$\mathrm{dist}(\boldsymbol{\beta}^{(0)}, \overline{\boldsymbol{\beta}}) \leq \mathrm{dist}(\widehat{\boldsymbol{v}}, \boldsymbol{\beta}^*) \cdot \sqrt{|\rho_n|/2} + \sqrt{|\rho_n - \rho|/2}. \tag{3.6}$$

Combining (3.3), (3.5), and (3.6), we obtain that

$$\mathrm{dist}(\boldsymbol{\beta}^{(0)}, \overline{\boldsymbol{\beta}}) \leq \mathcal{O}\big\{\big[s^2 \log(np)/n\big]^{1/4}\big\} \leq \alpha \|\overline{\boldsymbol{\beta}}\|,$$

which concludes the proof. □

Lemma 3.1 proves that the initial estimator is close to $\overline{\boldsymbol{\beta}}$ or $-\overline{\boldsymbol{\beta}}$ defined in (2.1) up to a constant order of error. We then proceed to characterize the computational and statistical properties of the thresholded Wirtinger flow algorithm in the next theorem.

**Theorem 3.1.** *Suppose Assumption 3.1 holds. Let the initial solution $\boldsymbol{\beta}^{(0)}$ be given by (2.12). Given $\alpha \leq 1/100$, $\kappa = \sqrt{80}$, $\eta \leq \eta_0/\rho$ for a constant $\eta_0$, and large enough $n$ satisfying*

$$n \geq C_1\big[s^2 \log(np) + s \log^5 n\big],$$

*for some constant $C_1$, there exists some constant $C_2$ such that for all $k = 0, 1, 2, \ldots, \mathrm{polylog}(pn)$, we have*

$$\mathrm{supp}(\boldsymbol{\beta}^{(k+1)}) \subseteq \mathrm{supp}(\boldsymbol{\beta}^*) \quad \textit{and} \quad \mathrm{dist}\big(\boldsymbol{\beta}^{(k+1)}, \overline{\boldsymbol{\beta}}\big) \leq \big(1 - \eta\rho\big)^k \cdot \mathrm{dist}\big(\boldsymbol{\beta}^{(0)}, \overline{\boldsymbol{\beta}}\big) + C_2\sqrt{s \log(np)/n},$$

*with probability at least $1 - \mathcal{O}(1/n)$.*

This theorem implies that given an appropriate initialization, the TWF algorithm maintains the solution sparsity throughout iterations, and attains a linear rate of convergence to $\overline{\boldsymbol{\beta}}$ up to some unavoidable finite-sample statistical error. Therefore, when the number of iterations $T$ satisfies

$$T \geq \left\lceil \frac{1}{\eta\rho} \cdot \log\left[\frac{\mathrm{dist}(\boldsymbol{\beta}^{(0)}, \overline{\boldsymbol{\beta}})}{\sqrt{s \log p/n}}\right]\right\rceil = \mathcal{O}[\log(n/s)] \tag{3.7}$$

and $T \leq \mathrm{poly}\log(np)$, with probability at least $1 - \mathcal{O}(1/n)$, there exists a constant $C_2' > 0$ such that

$$\mathrm{dist}\big(\boldsymbol{\beta}^{(T)}, \overline{\boldsymbol{\beta}}\big) \leq C_2'\sqrt{s \log p/n}.$$

Here we denote by $\lceil x \rceil$ the smallest integer no less than $x$.

Thus, Theorem 3.1 establishes a unified computational and statistical guarantees for our TWF algorithm. Specifically, this theorem implies that the estimation error of each $\boldsymbol{\beta}^{(t)}$ is the sum of the statistical error and the computation error. In specific, the statistical error is of order $\sqrt{s \log p/n}$, which cannot be further improved; the optimization error converges to zero in a linear rate as the algorithm proceeds. We point out that our analysis matches the optimal statistical rate of convergence for sparse phase retrieval as discussed in Cai et al. (2016). However, the theoretical analysis in Cai et al. (2016) only works for the correctly specified phase retrieval model, which is a special case of our model with link function $h(u, v) = u^2 + v$.

Moreover, note that the estimation error $\mathrm{dist}(\boldsymbol{\beta}^{(T)}, \overline{\boldsymbol{\beta}})$ is of order $\sqrt{s \log p / n}$ when $T$ is sufficiently large. While we require the sample complexity to be $n = \Omega(s^2 \log p)$. Similar phenomena have been observed by Cai et al. (2016), which is also conjectured to be the computational barrier of sparse phase retrieval models.

*Proof Sketch of Theorem 3.1.* Due to space limit, we provide an outline of the proof for the main theorem, and the detailed proof is provided in §A.2.

The proof of this theorem consists of two major steps. In the first step, let $S = \mathrm{supp}(\boldsymbol{\beta}^*)$ and let $\boldsymbol{z} \in \mathbb{R}^d$ be a point in the neighborhood of $\overline{\boldsymbol{\beta}}$ (or $-\overline{\boldsymbol{\beta}}$) such that $\mathrm{supp}(\boldsymbol{z}) \subseteq S$. We show that, starting from $\boldsymbol{z}$, the TWF update step restricted on $S$, moves toward the true solution $\overline{\boldsymbol{\beta}}$ (or $-\overline{\boldsymbol{\beta}}$) with high probability. In the second step, we show that the thresholding map forces the TWF updates to be supported on $S$ with high probability. The formal statement of the first step is presented in the following proposition.

**Proposition 3.2.** *Let $\kappa$ and $\eta$ be appropriately chosen positive constants as in (2.7) and (2.8). We denote $S = \mathrm{supp}(\boldsymbol{\beta}^*)$, and let $\mathcal{S} \subseteq \mathbb{R}^p$ be the set of all vectors in $\mathbb{R}^p$ that are supported on $S$. For any $\boldsymbol{z} \in \mathbb{R}^p$, we define a mapping $t \colon \mathbb{R}^p \to \mathbb{R}^p$ as*

$$t(\boldsymbol{z}) = \mathcal{T}_{\eta\tau(\boldsymbol{z})}\{\boldsymbol{z} - \eta[\nabla \ell_n(z)]_S\},$$

*where $\mathcal{T}$ is the hard-thresholding operator defined in (2.8), and the threshold value $\tau(\cdot)$ is defined in (2.7). For all $\boldsymbol{z} \in \mathcal{S}$ satisfying $\mathrm{dist}(\boldsymbol{z}, \overline{\boldsymbol{\beta}}) \le \alpha \|\overline{\boldsymbol{\beta}}\|$ with $\alpha \in (0, 1/100)$, given large enough $n$ satisfying $n \ge C[s^2 \log(np) + s \log^5 n]$, we have*

$$\mathrm{dist}\big[t(\boldsymbol{z}), \overline{\boldsymbol{\beta}}\big] \le (1 - \eta\rho) \cdot \mathrm{dist}(\boldsymbol{z}, \overline{\boldsymbol{\beta}}) + C' \sqrt{s \log(np)/n}$$

*for some constant $C'$, with probability at least $1 - 1/(5n)$.*

*Proof Sketch of Proposition 3.2.* Here we provide the outline of the proof, and the full proof of this proposition is presented in §A.3.

Without loss of generality, we assume $\mathrm{dist}(\boldsymbol{z}, \overline{\boldsymbol{\beta}}) = \|\boldsymbol{z} - \overline{\boldsymbol{\beta}}\|$, where the proof for the other case, $\mathrm{dist}(\boldsymbol{z}, \overline{\boldsymbol{\beta}}) = \|\boldsymbol{z} + \overline{\boldsymbol{\beta}}\|$, follows similarly. By the definition of the hard-thresholding operator $\mathcal{T}$, we first rewrite $t(\boldsymbol{z})$ as

$$t(\boldsymbol{z}) = \mathcal{T}_{\eta\tau(\boldsymbol{z})}\{\boldsymbol{z} - \eta[\nabla \ell_n(\boldsymbol{z})]_S\} = \boldsymbol{z} - \eta[\nabla \ell_n(\boldsymbol{z})]_S + \eta\tau(\boldsymbol{z})\boldsymbol{v},$$

where the last equality holds for some $\boldsymbol{v}$ satisfying $\boldsymbol{v} \in \mathcal{S}$ and $\|\boldsymbol{v}\|_\infty \le 1$.

Let $\boldsymbol{h} = \boldsymbol{z} - \overline{\boldsymbol{\beta}}$. We have $\boldsymbol{h} \in \mathcal{S}$ by definition. We then bound $\|t(\boldsymbol{z}) - \overline{\boldsymbol{\beta}}\|$ using $\|\boldsymbol{h}\|$. By the triangle inequality, we immediately have

$$\|t(\boldsymbol{z}) - \overline{\boldsymbol{\beta}}\| \le \underbrace{\|\boldsymbol{h} - \eta\nabla_S \ell_n(\boldsymbol{z})\|}_{R_1} + \underbrace{\eta\tau(\boldsymbol{z}) \cdot \|\boldsymbol{v}\|}_{R_2}.$$

In the following, we establish upper bounds for $R_1$ and $R_2$, respectively. We start with bounding $R_2$, which is essentially the error introduced by the hard-thresholding operator. Recall that by the definition of $\tau(\cdot)$ in (2.7), we have

$$\tau(\boldsymbol{z})^2 = \kappa^2 \cdot \frac{\log(np)}{n^2} \cdot \sum_{i=1}^{n} \big[y^{(i)} - (\boldsymbol{x}^{(i)\top}\boldsymbol{z})^2 - \mu_n + \|\boldsymbol{z}\|^2\big]^2 (\boldsymbol{x}^{(i)\top}\boldsymbol{z})^2.$$

Since $\boldsymbol{z} = \boldsymbol{h} + \overline{\boldsymbol{\beta}}$, we can rewrite $\tau(\boldsymbol{z})^2$ as a sum of terms of the form

$$\sum_{i=1}^{n} (y^{(i)})^a \big(\boldsymbol{x}^{(i)\top}\overline{\boldsymbol{\beta}}\big)^b \big(\boldsymbol{x}^{(i)\top}\boldsymbol{h}\big)^c,$$

where $a, b$, and $c$ are nonnegative integers. By carefully expanding these terms, we observe that the dominating term in $\tau(\boldsymbol{z})^2$ is

$$\kappa^2 \frac{\log(np)}{n^2} \sum_{i=1}^{n} \big(\boldsymbol{x}^{(i)\top}\overline{\boldsymbol{\beta}}\big)^2 \big(\boldsymbol{x}^{(i)\top}\boldsymbol{h}\big)^4.$$

Here, we derive an upper bound for this term as an example; see §A.3 for the details of bounding $\tau(\boldsymbol{z})^2$. By Hölder's inequality, we have

$$\sum_{i=1}^{n} \big(\boldsymbol{x}^{(i)\top}\overline{\boldsymbol{\beta}}\big)^2 \big(\boldsymbol{x}^{(i)\top}\boldsymbol{h}\big)^4 \le \bigg(\sum_{i=1}^{n} \big|\boldsymbol{x}^{(i)\top}\overline{\boldsymbol{\beta}}\big|^6\bigg)^{2/6} \bigg(\sum_{i=1}^{n} \big|\boldsymbol{x}^{(i)\top}\boldsymbol{h}\big|^6\bigg)^{4/6} \le \|\boldsymbol{A}\|_{2\to 6}^6 \|\overline{\boldsymbol{\beta}}\|^2 \cdot \|\boldsymbol{h}\|^4,$$

where $\|\boldsymbol{A}\|_{2\to 6}$ is the induced norm of $\boldsymbol{A}$. We show in §C that, when $n \ge C[s^2 \log(np) + s \log^5 n]$, we have $\|\boldsymbol{A}\|_{2\to 6}^6 = \mathcal{O}(n + s^3)$ with probability at least $1 - \mathcal{O}(1/n)$. Thus we have

$$\kappa^2 \frac{\log(np)}{n^2} \sum_{i=1}^{n} \big(\boldsymbol{x}^{(i)\top}\overline{\boldsymbol{\beta}}\big)^2 \big(\boldsymbol{x}^{(i)\top}\boldsymbol{h}\big)^4 \le C' \kappa^2 \frac{(n + s^3)\log(np)}{n^2} \|\overline{\boldsymbol{\beta}}\|^2 \cdot \|\boldsymbol{h}\|^4$$

for some constant $C'$. Using similar methods to bound other terms, we eventually obtain that

$$\tau(\boldsymbol{z})^2 \le C_\tau^2 \cdot (n + s^3) \cdot \log(np)/(n^2) \cdot \|\overline{\boldsymbol{\beta}}\|^2 \cdot \|\boldsymbol{h}\|^4 + \mathcal{O}\big[\log(np)/n\big],$$

where $C_\tau > 0$ is some constant. Since $\boldsymbol{v}$ is supported on $\mathrm{supp}(\boldsymbol{\beta}^*)$ with $\|\boldsymbol{v}\|_\infty \le 1$, we have $\|\boldsymbol{v}\| \le \sqrt{s}$. Hence, we obtain that

$$\begin{aligned}\tau(\boldsymbol{z})\|\boldsymbol{v}\| &= \mathcal{O}\Big[\kappa\sqrt{(ns + s^4)\log(np)/n^2} \cdot \|\overline{\boldsymbol{\beta}}\| \cdot \|\boldsymbol{h}\|^2\Big] + \mathcal{O}\Big[\sqrt{s\log(np)/n}\Big] \\ &= \mathcal{O}(\kappa\|\overline{\boldsymbol{\beta}}\|\|\boldsymbol{h}\|^2) + \mathcal{O}\Big[\sqrt{s\log(np)/n}\Big] = \mathcal{O}(\alpha\kappa\rho\|\boldsymbol{h}\|) + \mathcal{O}\Big[\sqrt{s\log(np)/n}\Big].\end{aligned}$$

Here in the second equality, we use $n \ge C[s^2 \log(np) + s \log^5 n]$, and the last equality follows from $\|\boldsymbol{h}\| \le \alpha\|\overline{\boldsymbol{\beta}}\|$ and $\|\overline{\boldsymbol{\beta}}\|^2 = \rho/2$ as shown in Proposition 2.1. Since $\alpha$ can be a arbitrarily small constant, we finally have

$$R_2 \le 3/4 \cdot \eta\rho\|\boldsymbol{h}\| + \mathcal{O}\Big[\sqrt{s\log(np)/n}\Big].$$

In addition, deriving the bound for $R_1$ follows from similar arguments. In the proof, we heavily rely on the fact that for any $\boldsymbol{g} \in \mathbb{R}$, $[\boldsymbol{g} - (\boldsymbol{g}^\top \boldsymbol{\beta}^*)\boldsymbol{\beta}^*]^\top \boldsymbol{X}$ is independent with $Y$. This is because $[\boldsymbol{g} - (\boldsymbol{g}^\top \boldsymbol{\beta}^*)\boldsymbol{\beta}^*]^\top \boldsymbol{\beta}^* = 0$, and $Y$ depends on $\boldsymbol{X}$ only through $\boldsymbol{X}^\top \boldsymbol{\beta}^*$. For the details of the proof, we refer the readers to §A.3. In particular, we show that, with high probability,

$$R_1 = (1 - 7/4\eta\rho) \cdot \|\boldsymbol{h}\| + \mathcal{O}\Big[\sqrt{s \log(np)/n}\Big].$$

Combining the bounds for $R_1$ and $R_2$, we conclude that for all $\boldsymbol{z} \in \mathcal{S}$,

$$\operatorname{dist}\big[t(z), \overline{\boldsymbol{\beta}}\big] = \|t(\boldsymbol{z}) - \overline{\boldsymbol{\beta}}\| = (1 - \eta\rho) \cdot \|\boldsymbol{h}\| + \mathcal{O}\Big[\sqrt{s \log(np)/n}\Big],$$

for sufficiently small $\alpha$ and $\eta$, which completes the proof of Proposition 3.2. □

To complete the proof of Theorem 3.1, in the the second step, we show that with high probability, the hard-thresholding operator keeps the thresholded gradient updates of the TWF algorithm supported on $\operatorname{supp}(\boldsymbol{\beta}^*)$. We establish this result by considering another sequence $\{\widetilde{\boldsymbol{\beta}}^{(k)}\}_{k \geq 0}$ which is specified by $\widetilde{\boldsymbol{\beta}}^{(0)} = \boldsymbol{\beta}^{(0)}$ and

$$\widetilde{\boldsymbol{\beta}}^{(k+1)} = \mathcal{T}_{\eta\tau(\widetilde{\boldsymbol{\beta}}^{(k)})}\Big\{\widetilde{\boldsymbol{\beta}}^{(k)} - \eta[\nabla \ell_n(\widetilde{\boldsymbol{\beta}}^{(k)})]_S\Big\} \tag{3.8}$$

for each $k \geq 0$, where we let $S = \operatorname{supp}(\boldsymbol{\beta}^*)$. That is, these two sequences $\{\widetilde{\boldsymbol{\beta}}^{(k)}\}_{k \geq 0}$ and $\{\boldsymbol{\beta}^{(k)}\}_{k \geq 0}$ have the same starting point $\boldsymbol{\beta}^{(0)}$, and the former is constructed by restricting the TWF updates to $\operatorname{supp}(\boldsymbol{\beta}^*)$.

In the sequel, we prove by induction that these two sequences coincide with high probability. Specifically, we show that

$$\boldsymbol{\beta}^{(k)} = \widetilde{\boldsymbol{\beta}}^{(k)}, \quad \operatorname{supp}(\widetilde{\boldsymbol{\beta}}^{(k)}) \subseteq \operatorname{supp}(\boldsymbol{\beta}^*), \quad \text{and} \quad \operatorname{dist}(\widetilde{\boldsymbol{\beta}}^{(k)}, \overline{\boldsymbol{\beta}}) \leq \alpha\|\overline{\boldsymbol{\beta}}\| \tag{3.9}$$

for all $k \geq 0$ with high probability. As shown in Lemma 3.1, (3.9) holds for $\boldsymbol{\beta}^{(0)}$ with probability at least $1 - \mathcal{O}(1/n)$.

Assuming (3.9) holds for some $k$, we consider $\widetilde{\boldsymbol{\beta}}^{(k+1)}$. We first note that, by the construction of $\widetilde{\boldsymbol{\beta}}^{(k+1)}$ in (3.8), we immediately have $\operatorname{supp}(\widetilde{\boldsymbol{\beta}}^{(k+1)}) \subseteq S$ provided $\operatorname{supp}(\widetilde{\boldsymbol{\beta}}^{(k)}) \subseteq S$. Moreover, by Proposition 3.2, when $n$ is sufficiently large, we have

$$\begin{aligned}\operatorname{dist}(\widetilde{\boldsymbol{\beta}}^{(k+1)}, \overline{\boldsymbol{\beta}}) &\leq (1 - \eta\rho) \cdot \operatorname{dist}(\widetilde{\boldsymbol{\beta}}^{(k)}, \overline{\boldsymbol{\beta}}) + C'\sqrt{s \log(np)/n} \\ &\leq (1 - \eta\rho) \cdot \operatorname{dist}(\widetilde{\boldsymbol{\beta}}^{(k)}, \overline{\boldsymbol{\beta}}) + \eta\rho\alpha\|\overline{\boldsymbol{\beta}}\| \leq \alpha\|\overline{\boldsymbol{\beta}}\|.\end{aligned}$$

Thus, it only remains to show that $\widetilde{\boldsymbol{\beta}}^{(k+1)} = \boldsymbol{\beta}^{(k+1)}$ under the inductive assumption $\widetilde{\boldsymbol{\beta}}^{(k)} = \boldsymbol{\beta}^{(k)}$. It suffices to show that the hard-thresholding operator sets any $j \notin S$ to zero. That is, we show that

$$\max_{j \notin S}\big|[\nabla \ell_n(\widetilde{\boldsymbol{\beta}}^{(k)})]_j\big| \leq \tau(\widetilde{\boldsymbol{\beta}}^{(k)})$$

holds with high probability. Since $\widetilde{\boldsymbol{\beta}}^{(k)}$ is supported on $S$, for any $j \notin S$, $\boldsymbol{x}_j^{(i)}$ is independent of

$$\varphi_i = \big[y^{(i)} - (\boldsymbol{x}^{(i)\top}\widetilde{\boldsymbol{\beta}}^{(k)})^2 - \xi_n(\widetilde{\boldsymbol{\beta}}^{(k)})\big] \cdot (\boldsymbol{x}^{(i)\top}\widetilde{\boldsymbol{\beta}}^{(k)}).$$

Moreover, by the definition of $\nabla \ell_n(\cdot)$ in (2.9), we have $[\nabla \ell_n(\widetilde{\boldsymbol{\beta}}^{(k)})]_j = -4n^{-1}\sum_{i=1}^n \varphi_i \cdot \boldsymbol{x}_j^{(i)}$. Therefore, when we condition on $\{\varphi_i\}_{i\in[n]}$, $[\nabla \ell_n(\widetilde{\boldsymbol{\beta}}^{(k)})]_j$ is a centered Gaussian random variable with variance $16/(n^2)\cdot\sum_{i=1}^n \varphi_i^2$. Thus, with probability at least $1-1/(n^2p)$, for all $j \notin S$, we have

$$\left|[\nabla \ell_n(\widetilde{\boldsymbol{\beta}}^{(k)})]_j\right| \le \left[\frac{80\log(np)}{n^2}\sum_{i=1}^n \varphi_j^2\right]^{1/2} \le \tau(\widetilde{\boldsymbol{\beta}}^k).$$

Therefore, we conclude that, with probability at least $1-1/(n^2p)$, we obtain

$$\widetilde{\boldsymbol{\beta}}^{(k+1)} = \mathcal{T}_{\eta\tau(\widetilde{\boldsymbol{\beta}}^k)}\left\{\widetilde{\boldsymbol{\beta}}^{(k)} - \eta[\nabla \ell_n(\widetilde{\boldsymbol{\beta}}^{(k)})]_S\right\} = \mathcal{T}_{\eta\tau(\widetilde{\boldsymbol{\beta}}^k)}[\widetilde{\boldsymbol{\beta}}^{(k)} - \eta\nabla \ell_n(\widetilde{\boldsymbol{\beta}}^{(k)})] = \boldsymbol{\beta}^{(k+1)},$$

which completes the induction step. By summing up the probabilities that the aforementioned events do not happen, we conclude that the theorem statement holds for all $k = 1, 2, \ldots \operatorname{poly}\log(np)$. □

# 4 Experimental Results

In this section, we evaluate the finite-sample performance of the proposed variant of TWF algorithm for misspecified phase retrieval models. We present results on both simulated and real data.

## 4.1 Simulated Data

For simulated data, we consider three link functions for the misspecified phase retrieval model

$$h_1(u,v) = |u| + v, \quad h_2(u,v) = |u+v|, \quad h_3(u,v) = 4u^2 + 3\sin(|u|) + v. \tag{4.1}$$

One can verify that the above link functions satisfy Assumption 3.1. Hence, the model in (1.3) with $h \in \{h_1, h_2, h_3\}$ is a misspecified phase retrieval model. Throughout the experiments, we fix $p = 1000$, $s \in \{5, 8, 10\}$, and let $n$ vary. We sample the random noise $\epsilon$ from the standard Gaussian distribution and let the covariate $\boldsymbol{X} \sim N(0, \mathbf{I}_p)$. For the signal parameter $\boldsymbol{\beta}^*$, we choose $\operatorname{supp}(\boldsymbol{\beta}^*)$ uniformly at random among all subsets of $[p]$ with cardinality $s$. Moreover, the nonzero entries of $\boldsymbol{\beta}^*$ is sampled uniformly from the unit sphere in $\mathbb{R}^s$. For tuning parameters of the initialization, we set $\gamma = 2$ in (2.10). We observe that our numerical results are not sensitive to the choice of $\gamma$, although in theory $\widehat{S}_0$ in (2.10) has to be close to $\operatorname{supp}(\boldsymbol{\beta}^*)$ so as to have a good initialization. For the other tuning parameters of the TWF algorithm, we set $\kappa$ in (2.7) and the step size $\eta$ to 15 and 0.005, respectively. Finally, we terminate the TWF algorithms, when $\|\boldsymbol{\beta}^{(t)} - \boldsymbol{\beta}^{(t-1)}\| \le 10^{-4}$.

We first illustrate the statistical rate of convergence, which is of the order $\sqrt{s\log p/n}$ as quantified in Theorem 3.1. In particular, we compare the estimation error of $\boldsymbol{\beta}^{(T)}$ with $\sqrt{s\log p/n}$, where $\boldsymbol{\beta}^{(T)}$ is the final estimator obtained by the TWF algorithm. Since $\boldsymbol{\beta}^*$ has unit norm and its sign is not identifiable, we use the cosine distance $1 - |\langle \widehat{\boldsymbol{\beta}}, \boldsymbol{\beta}^*\rangle|$ as a measure of the estimation error, where $\widehat{\boldsymbol{\beta}} = \boldsymbol{\beta}^{(T)}/\|\boldsymbol{\beta}^{(T)}\|$. This is equivalent to using $\operatorname{dist}(\cdot,\cdot)$, since

$$1 - |\langle \widehat{\boldsymbol{\beta}}, \boldsymbol{\beta}^*\rangle| = 1/2 \cdot \operatorname{dist}(\widehat{\boldsymbol{\beta}}, \boldsymbol{\beta}^*)^2.$$

Recall that Theorem 3.1 requires the sample size $n$ to be larger than $C \cdot s^2 \log p$ for some constant $C$. The term $s\sqrt{\log p/n}$ essentially quantifies the difficulty of recovering $\boldsymbol{\beta}^*$, which can be viewed as the inverse signal-to-noise (SNR) ratio in our problem. In the following, we plot $1 - |\langle \widehat{\boldsymbol{\beta}}, \boldsymbol{\beta}^* \rangle|$ against the inverse SNR $s\sqrt{\log p/n}$ in Figure 1 based on 100 independent trials for each $n$. Since we set $s$ to be small fixed constants in each case, as shown in these figures, the estimation error is bounded by a linear function of $\sqrt{s \log p/n}$, corroborating the statistical rate of convergence in Theorem 3.1.

In addition, we also the study the optimization error, which converges to zero at a linear rate, as shown in Theorem 3.1. We first define $\mathrm{Err}_t = 1 - |\langle \boldsymbol{\beta}^{(t)}, \boldsymbol{\beta}^* \rangle| / \|\boldsymbol{\beta}^{(t)}\|$ as the cosine error of $\boldsymbol{\beta}^{(t)}$ for all $t \geq 0$. Note that $\mathrm{Err}_t$ converges to a statistical error term that does not decrease with $t$. When $T$ is sufficiently large such that the algorithm converges, $\mathrm{Err}_t - \mathrm{Err}_T$ can be viewed as the optimization error of $\boldsymbol{\beta}^{(t)}$. Hence, viewed as a function of the number of iterations $t$, $\log(\mathrm{Err}_t - \mathrm{Err}_T)$ is close to a linear function of $t$ with a negative slope. To verify this result, in the following experiments, we set $p = 1000$, $s = 5$, and $n = 863$, which satisfy $n = 5s^2 \cdot \log p$ approximately. In addition, similar to the previous experiments, we study the three link functions in (4.1). For each link function in (4.1), we run the TWF algorithm for $T = 1000$ iterations with the tuning parameters set to $\gamma = 2$, $\kappa = 15$, and $\eta = 0.005$. In Figure 2, we plot $\log(\mathrm{Err}_t - \mathrm{Err}_T)$ against $t$ based on 50 independent trials. Here we report $t \in \{101, \ldots, 300\}$. For the 50 sequences of $\{\log(\mathrm{Err}_t - \mathrm{Err}_T)\}_{t \in [T]}$, we also compute the mean and standard error. In Figure 2, the blue curves are the mean values of these sequences and the red areas are confidence bands with one standard error. As shown in these figures, at the second stage of the algorithm, the optimization error decreases to zero at a linear rate of convergence, which corroborates our theory.

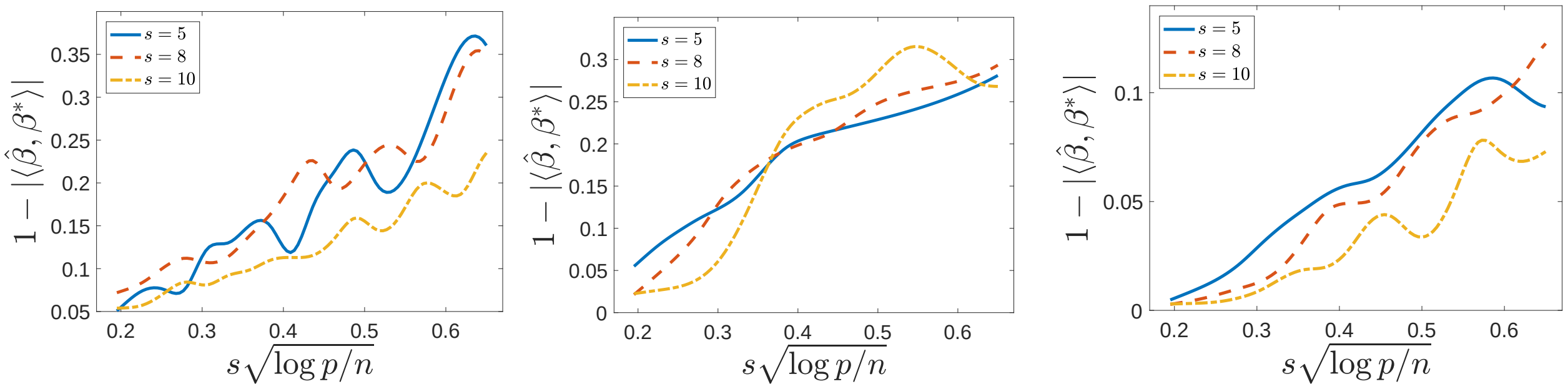

(a) $h_1(u,v) = |u| + v$ (b) $h_2(u,v) = |u+v|$, (c) $h_3(u,v) = 4u^2 + 3\sin(|u|) + v$

Figure 1: Plots of the cosine distance $1 - |\langle \widehat{\boldsymbol{\beta}}, \boldsymbol{\beta}^* \rangle|$ against the inverse signal-to-noise ratio $s\sqrt{\log p/n}$, in which $\widehat{\boldsymbol{\beta}}$ is obtained by normalizing the final output $\boldsymbol{\beta}^{(T)}$. The link function is one of $h_1$, $h_2$, and $h_3$ in (4.1). Besides, we set $p = 1000$, $s \in \{5, 8, 10\}$, and let $n$ vary. We generate each of the figures based on 100 independent trials for each $(n, s, p)$.

## 4.2 Real-World Data

We consider an example in image processing. Let $\mathbf{M} \in \mathbb{R}^{H \times W}$ be an image, where $H$ and $W$ denote the height and width of $\mathbf{M}$, correspondingly. Without loss of generality, we assume that

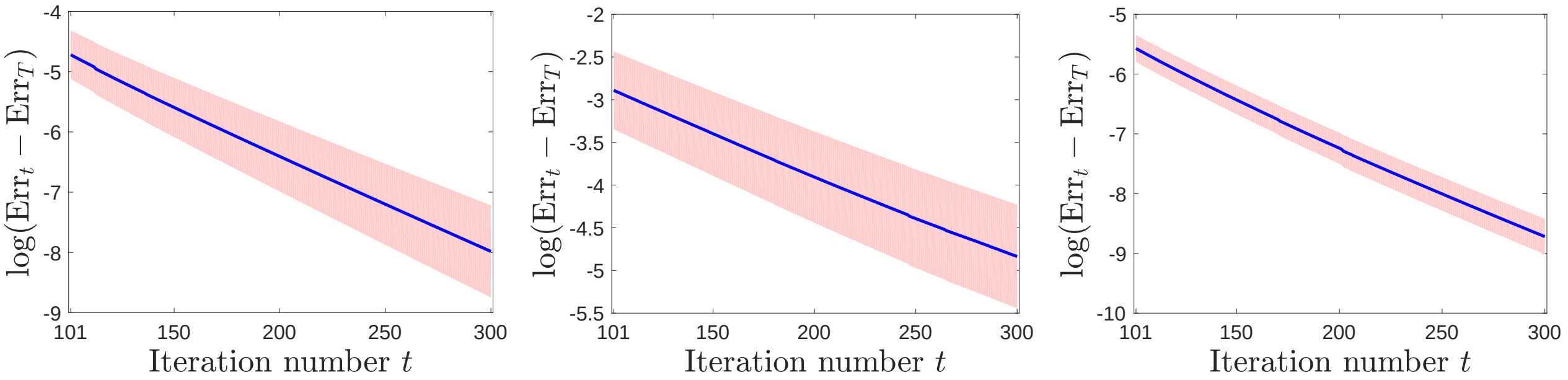

(a) $h_1(u, v) = |u| + v$ (b) $h_2(u, v) = |u + v|$, (c) $h_3(u, v) = 4u^2 + 3\sin(|u|) + v$

Figure 2: Plots of the log optimization error $\log(\mathrm{Err}_t - \mathrm{Err}_T)$ against the number of iterations $t$ for $t \in \{101, 102, \ldots, 300\}$, in which $\mathrm{Err}_t$ is the cosine error of the $t$-th iterate of the TWF algorithm and $T = 1000$ is the total number of iterations. Here the link function is one of $h_1$, $h_2$, and $h_3$. In addition, we set $p = 1000$, $s = 5$, and $n = 864 \approx 5s^2 \cdot \log p$. These figures are generated based on 50 independent trials for each link function.

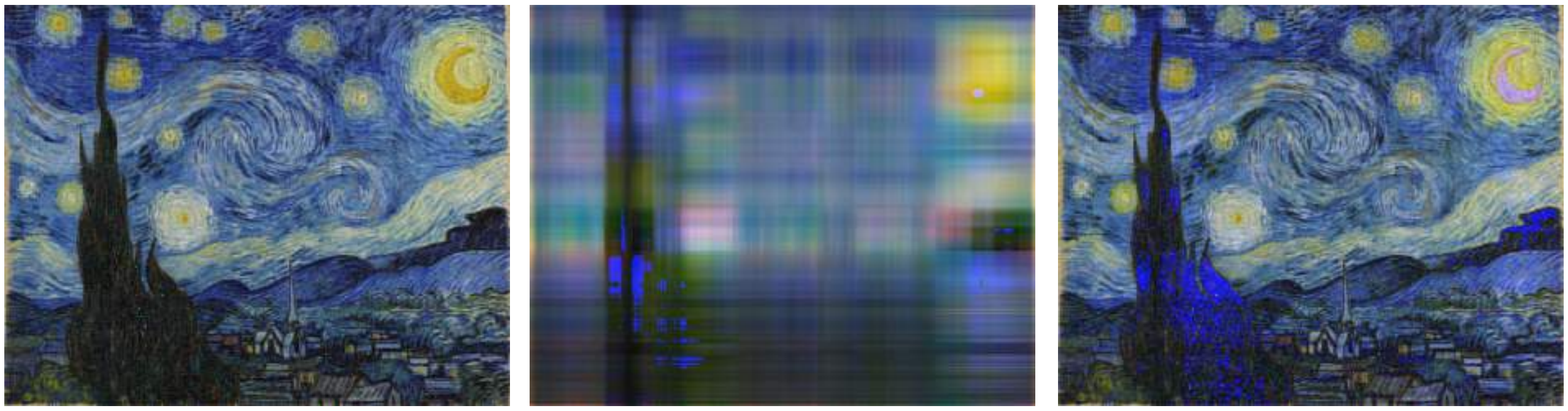

(a) Image corresponding to $\boldsymbol{\beta}^*$. (b) Image obtained by $\boldsymbol{\beta}^{(0)}$. (c) Image obtained by $\boldsymbol{\beta}^{(T)}$.

Figure 3: Plots of the image used for misspecified phase retrieval and the images reconstructed using the TWF algorithm. The rank-80 approximation of the original *Starry Night* is plotted in (a), which corresponds to the signal parameter $\boldsymbol{\beta}^*$. We plot the image obtained using the initialization $\boldsymbol{\beta}^{(0)}$ in (b), which is a poor reconstruction of the image in (a). In (c) we plot the reconstructed image using $\boldsymbol{\beta}^{(T)}$ with $T = 1000$, whose difference from the image in (a) is negligible.

$H \le W$. Let $\mathbf{M} = \sum_{i=1}^{H} \boldsymbol{\alpha}_i \mathbf{u}_i \mathbf{v}_i^\top$ be the singular value decomposition of $\mathbf{M}$, where $\{\boldsymbol{\alpha}_i\}_{i \in [H]}$ are the singular values of $\mathbf{M}$ in the descending order, and $\mathbf{u}_i$ and $\mathbf{v}_i$ are the corresponding singular vectors. The best rank-$s$ approximation of $\mathbf{M}$ is given by $\widetilde{\mathbf{M}} = \sum_{i=1}^{s} \boldsymbol{\alpha}_i \mathbf{u}_i \mathbf{v}_i^\top$. The success of image compression implies that real-world images can be well approximated by some low-rank matrix, that is, $\widetilde{\mathbf{M}}$ is close to $\mathbf{M}$ for some $s \ll H$. When the singular vectors $\{\mathbf{u}_i, \mathbf{v}_i\}_{i \in [H]}$ are all fixed, $\widetilde{\mathbf{M}}$ can be represented by an $s$-sparse vector of singular values $\boldsymbol{\alpha} = (\boldsymbol{\alpha}_1, \ldots, \boldsymbol{\alpha}_s, 0, \ldots, 0)^\top \in \mathbb{R}^H$. In this example, we treat $\boldsymbol{\alpha}$ as the signal parameter, and treat $\{\mathbf{u}_i, \mathbf{v}_i\}_{i \in [H]}$ as a given collection of vectors.

In Figure 3-(a) we plot the low-rank approximation of *Starry Night* by Vincent van Gogh. Here the height and width of the image is $H = 1014$ and $W = 1280$. For each channel of the image, we compute the best rank-$s$ approximation with $s = 80$. Let $\boldsymbol{\alpha}$ be the representation of this low-rank matrix and define $\boldsymbol{\beta}^* = \boldsymbol{\alpha}/\|\boldsymbol{\alpha}\|$. We use $\boldsymbol{\beta}^*$ as the signal parameter in the misspecified phase retrieval model and reconstruct $\boldsymbol{\beta}^*$ using the TWF algorithm. Let $\widehat{\boldsymbol{\beta}}$ be the estimator, we can then use the quality of the reconstructed image $\widehat{\mathbf{M}} = \|\boldsymbol{\alpha}\| \cdot \sum_{j=1}^{H} \widehat{\boldsymbol{\beta}}_j \mathbf{u}_j \mathbf{v}_j^\top$ to evaluate the performance of the TWF algorithm.

In specific, we set the link function as $h_2$ in (4.1). We sample $n = 10s^2 \log(p)$ observations of the model in (1.3) with $\epsilon \sim N(0, 1)$, where $s = 80$, $p = H = 1014$, and the covariate $\boldsymbol{X} \sim N(0, \mathbf{I}_p)$. For the tuning parameters of the TWF algorithm, similar to the simulated data, we set $\gamma = 2$, $\tau = 15$, and $\eta = 0.005$. Here we output the final estimator $\boldsymbol{\beta}^{(T)}$ and the corresponding image after $T = 1000$ gradient steps. For comparison, in Figure 3-(b) we plot the image constructed using $\boldsymbol{\beta}^{(0)}$, which is obtained by spectral initialization. It is evident that this image is blurred and does not capture the pattern of the target image in Figure 3-(a). In sharp contrast, as shown in Figure 3-(c), the image constructed using $\boldsymbol{\beta}^{(T)}$ is very close to the target image. Furthermore, the difference is nearly indiscernible, which illustrates the empirical success of the TWF algorithm for real-data.

# 5 Concluding Remarks

In this paper, we establish unified statistical and computational results for a simple variant of the TWF algorithm, which is robust over a large class of misspecified phase retrieval models. In specific, we prove that, with a proper initialization, the proposed algorithm linearly converges to an estimator with optimal statistical accuracy. Perhaps surprisingly, both our sample complexity $n = \Omega(s^2 \log p)$ and statistical rate of convergence $\sqrt{s \log p/n}$ match the best possible results for the case where the sparse phase retrieval model is correctly specified. To the best of our knowledge, our paper makes the first attempt to understand the robustness of nonconvex statistical optimization under model misspecification. We hope that our techniques can be further extended to more general problems in the future.

# A Proofs of Main Results

In this section, we prove Lemma 3.1 and Theorem 3.1 in §A.1 and §A.2, respectively. In what follows, we let $S = \{1, 2, \ldots s\}$ be the true support of $\boldsymbol{\beta}^*$ without loss of generality.

## A.1 Proof of Lemma 3.1

*Proof.* For ease of notation, we let $I_\ell = n^{-1} \sum_{i=1}^n y^{(i)} \cdot (\boldsymbol{x}_\ell^{(i)})^2$ for any $\ell \in [p]$. In addition, we let $\boldsymbol{y} = (y^{(1)}, \ldots, y^{(n)})^\top \in \mathbb{R}^n$ be the vector of responses, $\boldsymbol{A} = (\boldsymbol{x}^{(1)}, \boldsymbol{x}^{(2)}, \ldots, \boldsymbol{x}^{(3n)})^\top \in \mathbb{R}^{3n \times d}$ be the data matrix, and let $\mu_n = n^{-1} \sum_{i=1}^n y^{(i)}$. Then, $\widehat{S}_0$ defined in (2.10) can be written as

$$\widehat{S}_0 = \Big\{ \ell \in [p] : |I_\ell - \mu_n| > \gamma \sqrt{\log(np)/n} \Big\}.$$

Also, we define $\widetilde{S}_0$ as the intersection of $\widehat{S}_0$ and the support of $\boldsymbol{\beta}^*$, or equivalently,

$$\widetilde{S}_0 = \Big\{ \ell \in \mathrm{supp}(\boldsymbol{\beta}^*) : |I_\ell - \mu_n| > \gamma \sqrt{\log(np)/n} \Big\}. \tag{A.1}$$

Recall that we assume that $\mathrm{supp}(\boldsymbol{\beta}^*) = S = \{1, 2, \ldots, s\}$ without loss of generality. Since $\boldsymbol{y}$ only depends on $\boldsymbol{A}$ through $\boldsymbol{A}_S$, we have that $\{I_1, I_2, \ldots, I_s, \mu_n\}$ and $\boldsymbol{A}_{S^c}$ are independent. Thus, $\widetilde{S}_0$ defined in (A.1) is also independent of $\boldsymbol{A}_{S^c}$. In what follows, we denote $\boldsymbol{x}_{\widetilde{S}_0}^{(i)}$ by $\widetilde{\boldsymbol{w}}^{(i)}$ for $i \in \{n+1, \ldots, 2n\}$. Similarly to $\boldsymbol{W}$ in (2.11), we define $\widetilde{\boldsymbol{v}}$ as the eigenvector of

$$\widetilde{\boldsymbol{W}} = \frac{1}{n} \sum_{i=n+1}^{2n} (y^{(i)} - \mu_n) \cdot \widetilde{\boldsymbol{w}}^{(i)} (\widetilde{\boldsymbol{w}}^{(i)})^\top$$

corresponding to the largest eigenvalue in magnitude. Furthermore, we let

$$\widetilde{\boldsymbol{\beta}}^{(0)} = \widetilde{\boldsymbol{v}} \sqrt{|\widetilde{\rho}_n|/2}, \quad \text{where} \quad \widetilde{\rho}_n = \frac{1}{n} \sum_{i=2n+1}^{3n} y^{(i)} \cdot (\boldsymbol{x}^{(i)\top} \widetilde{\boldsymbol{v}})^2 - \mu_n. \tag{A.2}$$

Then, by definition, $\widetilde{\boldsymbol{\beta}}^{(0)}$ is independent of $\boldsymbol{A}_{S^c}$, and $\mathrm{supp}(\widetilde{\boldsymbol{\beta}}^{(0)}) \subseteq \mathrm{supp}(\boldsymbol{\beta}^*)$.

**Step 1.** We first show that $\widetilde{\boldsymbol{\beta}}^{(0)} = \boldsymbol{\beta}^{(0)}$ with high probability, which implies that the support of $\boldsymbol{\beta}^{(0)}$ is contained in $\mathrm{supp}(\boldsymbol{\beta}^*)$. By definition, it suffices to show that $\widetilde{S}_0 = \widehat{S}_0$. Note that

$$I_\ell - \mu_n = \frac{1}{n} \sum_{j=1}^n y^{(i)} \big[ (\boldsymbol{x}_\ell^{(i)})^2 - 1 \big], \tag{A.3}$$

which corresponds to the sample average of $\mathbb{E}[Y \cdot (\boldsymbol{X}_\ell^2 - 1)]$. Thus, for any $\ell \notin \mathrm{supp}(\boldsymbol{\beta}^*)$, since $\boldsymbol{\beta}_\ell^* = 0$, by (3.1) and (A.3) we have $\mathbb{E}(I_\ell - \mu_n) = 0$. In addition, since $\boldsymbol{X}_\ell$ and $Y$ are independent, we further obtain

$$\mathbb{E}(I_\ell - \mu_n \mid y^{(1)}, \ldots, y^{(n)}) = 0.$$

Since $\boldsymbol{X}_\ell^2$ is a $\chi^2$-random variable with one degree of freedom for each $\ell \in [p]$, using concentration inequalities for $\chi^2$-random variables, for example, Lemma 2.7 of Laurent and Massart (2000), we obtain that, for any $t > 0$ and $\ell \notin \text{supp}(\boldsymbol{\beta}^*)$,

$$\mathbb{P}(n \cdot |I_\ell - \mu_n| \geq 2\sqrt{t} \cdot \|\boldsymbol{y}\| + 2t \cdot \|\boldsymbol{y}\|_\infty \mid y^{(1)}, \ldots, y^{(n)}) \leq \exp(-t).$$

Taking the union bound for all $\ell \notin \text{supp}(\boldsymbol{\beta}^*)$, we obtain that

$$\mathbb{P}\Big(n \cdot \max_{\ell \notin S} |I_\ell - \mu_n| \geq 2\sqrt{t} \cdot \|\boldsymbol{y}\| + 2t \cdot \|\boldsymbol{y}\|_\infty \;\Big|\; y^{(1)}, \ldots, y^{(n)}\Big) \leq p \cdot \exp(-t), \tag{A.4}$$

In addition, under Assumption 3.1, $Y$ is a sub-exponential random variables with $\psi_1$-norm bounded by $\Psi$. Thus, for any $t > 0$, by union bound, we have

$$\mathbb{P}(\|\boldsymbol{y}\|_\infty > t) = \mathbb{P}(\max_i |y^{(i)}| \geq t) \leq n \cdot \mathbb{P}(|Y| \geq t) \leq 2n \cdot \exp(-t/\Psi). \tag{A.5}$$

Choosing a constant $C_1$ satisfying $C_1 \log n \geq \Psi\{2 \log n + \log(200)\}$, by (A.5) we obtain

$$\mathbb{P}(\|\boldsymbol{y}\|_\infty > C_1 \log n) \leq 1/(100n). \tag{A.6}$$

In addition, since $\|Y\|_{\psi_1} \leq \Psi$, we have $\text{Var}(Y^2) \leq C_2$ for some constant $C_2$ which only depends on $\Psi$. Then by Chebyshev's inequality we have

$$\begin{aligned}\mathbb{P}(\|\mathbf{y}\| > t) &= \mathbb{P}\Big[\sum_{j=1}^n (y^{(i)})^2 - n \cdot \mathbb{E}(Y^2) > t^2 - n \cdot \mathbb{E}(Y^2)\Big] \\ &\leq \big[t^2 - n \cdot \mathbb{E}(Y^2)\big]^{-2} \cdot \sum_{j=1}^n \text{Var}\big[(y^{(i)})^2\big] \leq C_2 n \cdot \big[t^2 - n \cdot \mathbb{E}(Y^2)\big]^{-2}.\end{aligned} \tag{A.7}$$

Setting $t = C_3\sqrt{n}$ in (A.7) with $C_3$ being a sufficiently large constant, we have

$$\mathbb{P}(\|\boldsymbol{y}\| > C_3\sqrt{n}) \leq 1/(100n). \tag{A.8}$$

We note that the constant $C_3$ only depends on $\Psi$, and can be computed using $C_2$. In the following, we condition on the event that $\|\boldsymbol{y}\|_\infty \leq C_1 \log n$ and $\|\boldsymbol{y}\| \leq C_3\sqrt{n}$, which holds with probability at least $1 - 1/(50n)$ by (A.6) and (A.8). Setting $t = \log(20np)$ in (A.4), we obtain that with probability at least $1 - 1/(10n)$, for all $\ell \notin \text{supp}(\boldsymbol{\beta}^*)$,

$$|I_\ell - \mu_n| \leq 1/n \cdot (2\sqrt{t} \cdot \|\boldsymbol{y}\| + 2t \cdot \|\boldsymbol{y}\|_\infty) \leq C\sqrt{\log(np)/n}, \tag{A.9}$$

where $C$ is a constant that can be computed using $C_1$ and $C_3$. Therefore, when $\gamma$ is a sufficiently large constant, (A.9) implies that $|I_\ell - \mu_n| \leq \gamma\sqrt{\log(np)/n}$ for all $\ell \notin \text{supp}(\boldsymbol{\beta}^*)$. Recall that $\widetilde{S}_0 = \widehat{S}_0 \cap \text{supp}(\boldsymbol{\beta}^*)$. By the definition of $\widehat{S}_0$ in (2.10), we have $\widehat{S}_0 \setminus \text{supp}(\boldsymbol{\beta}^*) = \emptyset$. Thus, it holds that $\widehat{S}_0 = \widetilde{S}_0$, which leads to $\widetilde{\boldsymbol{\beta}}^{(0)} = \boldsymbol{\beta}^{(0)}$, and we have $\text{supp}(\boldsymbol{\beta}^{(0)}) \subseteq \text{supp}(\boldsymbol{\beta}^*)$.

**Step 2.** In this step, we show that $\widehat{S}_0$ contains all the coordinates of $\boldsymbol{\beta}^*$ that are sufficiently large in magnitude. We prove this by showing the norm of $\boldsymbol{\beta}^* - \boldsymbol{\beta}^*_{\widehat{S}_0}$ is small.

First, we note that, for any $\ell \in \text{supp}(\boldsymbol{\beta}^*)$, we have $\mathbb{E}(I_\ell) = \mu + \rho \cdot \boldsymbol{\beta}_\ell^{*2}$. In addition, since both $Y$ and $(\boldsymbol{X}_\ell)^2$ are sub-exponential random variables with bounded $\psi_1$-norms, by Lemma C.4, we obtain that with probability at least $1 - 1/(10n)$,

$$\Big| I_\ell - \mu - \rho \cdot \boldsymbol{\beta}_\ell^{*2} \Big| \le C_4 \sqrt{\log n / n}, \quad \text{for all} \quad \ell \in \text{supp}(\boldsymbol{\beta}^*). \tag{A.10}$$

for some absolute constant $C_4$. Moreover, since $\|Y\|_{\psi_1} \le \Psi$, by a Bernstein-type inequality for sub-exponential random variables (Proposition 5.16 in Vershynin (2010)), we have

$$\mathbb{P}(|\mu_n - \mu| > t) \le 2 \exp\big[-C \cdot \min\{nt^2/(\Psi^2), nt/\Psi\}\big], \tag{A.11}$$

where $C$ is a constant. Setting $t = C_5 \Psi \cdot \sqrt{\log n / n}$ in (A.11), where $C_5$ is a constant, we conclude that with probability at least $1 - 1/(10n)$,

$$|\mu_n - \mu| \le C_5 \Psi \cdot \sqrt{\log n / n}. \tag{A.12}$$

Then we combine (A.10), (A.12), and apply the triangle inequality to obtain that

$$\big| I_\ell - \mu_n - \rho \cdot \boldsymbol{\beta}_\ell^{*2} \big| \le \Big| I_\ell - \mu_n - \rho \cdot \boldsymbol{\beta}_\ell^{*2} \Big| + |\mu_n - \mu| \le C_6 \cdot \sqrt{\log(np)/n}, \quad \forall \ell \in \text{supp}(\boldsymbol{\beta}^*) \tag{A.13}$$

with probability at least $1 - 1/(10n)$, where $C_6$ is a sufficiently large constant. Combining (A.9) and (A.13), we conclude that with probability at least $1 - 1/(5n)$, for any $\ell \in [p]$, there exists a constant $C_\Psi > 0$ which only depends on $\Psi$ such that

$$|I_\ell - \mu_n - \rho \cdot \boldsymbol{\beta}_\ell^{*2}| \le C_\Psi \cdot \sqrt{\log(np)/n}. \tag{A.14}$$

In what follows, we condition on the event that (A.14) holds. Consider the set

$$S_0 = \Big\{ \ell \in \text{supp}(\boldsymbol{\beta}^*) : |\rho| \cdot \boldsymbol{\beta}_\ell^{*2} \ge (\gamma + C_\Psi) \cdot \sqrt{\log(np)/n} \Big\}. \tag{A.15}$$

Combining (A.14), (A.15), and by the triangle inequality, for all $\ell \in S_0$, it holds that

$$|I_\ell - \mu_n| \ge \Big| \rho \boldsymbol{\beta}_\ell^{*2} \Big| - |I_\ell - \mu_n - \rho \cdot \boldsymbol{\beta}_\ell^{*2}| \ge \gamma \sqrt{\log(np)/n}.$$

This implies that $S_0 \subseteq \widehat{S}_0$ Thus, we have

$$\|\boldsymbol{\beta}^* - \boldsymbol{\beta}^*_{\widehat{S}_0}\|^2 \le \|\boldsymbol{\beta}^* - \boldsymbol{\beta}^*_{S_0}\|^2 = |\text{supp}(\boldsymbol{\beta}^*)| \cdot \max_{\ell \notin S_0} \boldsymbol{\beta}_\ell^{*2} \le (\gamma + C_\Psi)/|\rho| \cdot s \sqrt{\log(np)/n}. \tag{A.16}$$

which concludes this step.

**Step 3.** In this step, we show that the eigenvector $\widehat{\boldsymbol{v}}$ of $\boldsymbol{W}$ defined in (2.11) is close to $\boldsymbol{\beta}^*$. Recall that $\widehat{S}_0$ is independent of $\boldsymbol{W}$ by our sample splitting setup, and we denote $\boldsymbol{x}^{(i)}_{\widehat{S}_0}$ by $\boldsymbol{w}^{(i)}$. In addition, we define

$$\boldsymbol{\beta}_0^* = \boldsymbol{\beta}^*_{\widehat{S}_0} / \|\boldsymbol{\beta}^*_{\widehat{S}_0}\|_2. \tag{A.17}$$

First, we note that by (A.16), the denominator in (A.17) is nonzero, and thus $\boldsymbol{\beta}_0^*$ is well-defined. Hereafter, we let $\boldsymbol{w}_0^{(i)} = \boldsymbol{w}^{(i)} - \boldsymbol{\beta}_0^{*\top}\boldsymbol{w}^{(i)} \cdot \boldsymbol{\beta}_0^*$. Then, by definition, we have

$$\boldsymbol{w}^{(i)}(\boldsymbol{w}^{(i)})^\top = \boldsymbol{w}_0^{(i)}(\boldsymbol{w}_0^{(i)})^\top + \boldsymbol{\beta}_0^{*\top}\boldsymbol{w}^{(i)} \cdot [\boldsymbol{w}_0^{(i)}\boldsymbol{\beta}_0^{*\top} + \boldsymbol{\beta}_0^*(\boldsymbol{w}_0^{(i)})^\top] + (\boldsymbol{\beta}_0^{*\top}\boldsymbol{w}^{(i)})^2 \cdot \boldsymbol{\beta}_0^*\boldsymbol{\beta}_0^{*\top}. \tag{A.18}$$

Plugging (A.18) into the definition $\boldsymbol{W}$ in (2.11), we can rewrite $\boldsymbol{W}$ as

$$\boldsymbol{W} = \frac{1}{n}\sum_{i=n+1}^{2n}(y^{(i)} - \mu) \cdot |\boldsymbol{\beta}_0^{*\top}\boldsymbol{w}^{(i)}|^2 \cdot \boldsymbol{\beta}_0^*\boldsymbol{\beta}_0^{*\top} + \boldsymbol{N},$$

where $\boldsymbol{N} \in \mathbb{R}^{p\times p}$ can be viewed as an error matrix. We further decompose $\boldsymbol{N}$ into three terms as $\boldsymbol{N} = \boldsymbol{N}_1 + \boldsymbol{N}_2 + \boldsymbol{N}_3 \in \mathbb{R}^{p\times p}$, where

$$\boldsymbol{N}_1 = \frac{1}{n}\sum_{i=n+1}^{2n}(\mu_n - \mu) \cdot \boldsymbol{w}^{(i)}(\boldsymbol{w}^{(i)})^\top, \quad \boldsymbol{N}_2 = \frac{1}{n}\sum_{i=n+1}^{2n}(y^{(i)} - \mu) \cdot \boldsymbol{\beta}_0^{*\top}\boldsymbol{w}^{(i)} \cdot [\boldsymbol{w}_0^{(i)}\boldsymbol{\beta}_0^{*\top} + \boldsymbol{\beta}_0^*(\boldsymbol{w}_0^{(i)})^\top]$$

$$\boldsymbol{N}_3 = \frac{1}{n}\sum_{i=n+1}^{2n}(y^{(i)} - \mu) \cdot \boldsymbol{w}_0^{(i)}(\boldsymbol{w}_0^{(i)})^\top.$$

We then bound the spectral norm of $\boldsymbol{N}$ by the triangle inequality that $\|\boldsymbol{N}\|_2 \le \|\boldsymbol{N}_1\|_2 + \|\boldsymbol{N}_2\|_2 + \|\boldsymbol{N}_3\|_2$, and then further bound each term respectively.

We first show that $\boldsymbol{N}_1$, $\boldsymbol{N}_2$, and $\boldsymbol{N}_3$ all have mean zero. For $\boldsymbol{N}_1$, recall that $\mu_n$ is computed using the first $n$ samples, which are independent of $\boldsymbol{w}^{(n+1)}, \ldots, \boldsymbol{w}^{(2n)}$. Since $\mathbb{E}(\mu_n) = \mu$, we have $\mathbb{E}(\boldsymbol{N}_1) = \mathbf{0}$. In addition, for any $i \in \{n+1, \ldots, 2n\}$ and any $\boldsymbol{b} \in \mathbb{R}^p$, by direct calculation, we have

$$\begin{aligned}\mathbb{E}\big[(\boldsymbol{b}^\top\boldsymbol{w}_0^{(i)}) \cdot (\boldsymbol{\beta}^{*\top}\boldsymbol{x}^{(i)})\big] &= \mathbb{E}\big[(\boldsymbol{b}^\top\boldsymbol{w}^{(i)}) \cdot (\boldsymbol{\beta}_0^{*\top}\boldsymbol{x}^{(i)})\big] - \mathbb{E}\big[(\boldsymbol{\beta}_0^{*\top}\boldsymbol{w}^{(i)}) \cdot (\boldsymbol{\beta}^{*\top}\boldsymbol{x}^{(i)})\big] \cdot \boldsymbol{b}^\top\boldsymbol{\beta}_0^* \\ &= \mathbb{E}\big[(\boldsymbol{b}_{\widehat{S}_0}^\top\boldsymbol{x}^{(i)}) \cdot (\boldsymbol{\beta}^{*\top}\boldsymbol{x}^{(i)})\big] - \mathbb{E}\big[(\boldsymbol{\beta}_{\widehat{S}_0}^{*\top}\boldsymbol{x}^{(i)}) \cdot (\boldsymbol{\beta}^{*\top}\boldsymbol{x}^{(i)})\big] \cdot \boldsymbol{b}^\top\boldsymbol{\beta}_0^* = \boldsymbol{b}^\top\boldsymbol{\beta}_{\widehat{S}_0}^* - \|\boldsymbol{\beta}_{\widehat{S}_0}^*\|_2 \cdot \boldsymbol{b}^\top\boldsymbol{\beta}_0^* = 0.\end{aligned}$$

Thus, $\boldsymbol{b}^\top\boldsymbol{w}_0^{(i)}$ and $\boldsymbol{\beta}^{*\top}\boldsymbol{x}^{(i)}$ are uncorrelated Gaussian random variables, which are then independent. Hence $\boldsymbol{w}_0^{(i)}$ and $\boldsymbol{\beta}^{*\top}\boldsymbol{x}^{(i)}$ are independent, which further implies that $\boldsymbol{w}_0^{(i)}$ is independent of $y^{(i)}$. Thus, we have $\mathbb{E}(\boldsymbol{N}_3) = \mathbf{0}$. Similarly, for any $\boldsymbol{b} \in \mathbb{R}^p$, we have

$$\mathbb{E}\big[(\boldsymbol{b}^\top\boldsymbol{w}_0^{(i)}) \cdot (\boldsymbol{\beta}_0^{*\top}\boldsymbol{w}^{(i)})\big] = \mathbb{E}\big[\boldsymbol{b}^\top(\mathbf{I}_p - \boldsymbol{\beta}_0^*\boldsymbol{\beta}_0^{*\top})\boldsymbol{w}^{(i)}\boldsymbol{w}^{(i)}\boldsymbol{\beta}_0^*\big] = 0,$$

and $\boldsymbol{w}_0^{(i)}$ is independent of $(y^{(i)} - \mu) \cdot \boldsymbol{\beta}_0^{*\top}\boldsymbol{w}^{(i)}$, which implies that $\mathbb{E}(\boldsymbol{N}_2) = \mathbf{0}$.

Then, we derive an upper bound for $\|\boldsymbol{N}_3\|_{\max}$, which is used to upper bound $\|\boldsymbol{N}_3\|_2$. Upper bounds for $\|\boldsymbol{N}_1\|_2$ and $\|\boldsymbol{N}_2\|_2$ can be established by similar arguments. Note that for each $j_1, j_2 \in \mathrm{supp}(\boldsymbol{\beta}^*)$, both $(y^{(i)} - \mu)$ and $\boldsymbol{w}_{j_1}^{(i)}\boldsymbol{w}_{j_2}^{(i)}$ are sub-exponential random variables with constant sub-exponential norms. By Lemma C.4, we have

$$\big|(\boldsymbol{N}_3)_{j_1,j_2}\big| \le C_N\sqrt{\log(ns)/n},$$

with probability at least $1 - 1/(100ns^2)$, where $C_N$ is a constant that does not depends on $j_1$ and $j_2$. Taking a union bound over all $s^2$ non-zero entries of $\boldsymbol{N}_3$, we have

$$\|\boldsymbol{N}_3\|_{\max} \le C_N \cdot \sqrt{\log n/n}$$

with probability at least $1 - 1/(100n)$ for some constant $C_N$. We note that similar upper bounds for $\boldsymbol{N}_1$ and $\boldsymbol{N}_2$ can be derived using the same argument. By taking a union bound over all three matrices, we conclude that, with probability at least $1 - 1/(30n)$, we have

$$\|\boldsymbol{N}\|_{\max} \le C_N \sqrt{\log n / n}. \tag{A.19}$$

We condition on the event that (A.19) holds hereafter. Note that $\boldsymbol{N}$ is supported on $\{(j_1, j_2)\colon j_1, j_2 \in \operatorname{supp}(\boldsymbol{\beta}^*)\}$. By standard inequalities of different matrix norms, we have

$$\|\boldsymbol{N}\|_2 \le \|\boldsymbol{N}\|_F \le \sqrt{|\operatorname{supp}(\boldsymbol{N})|} \cdot \|\boldsymbol{N}\|_{\max} \le C_N \cdot s\sqrt{\log n/n},$$

where $\|\cdot\|_F$ is the matrix Frobenius norm.

Next, we introduce a lemma that bounds the spectral perturbation of spiked matrices.

**Lemma A.1** (Perturbation Lemma for Spiked Matrices). *Let $\mathbf{A} = \lambda \mathbf{v}\mathbf{v}^\top + \mathbf{N} \in \mathbb{R}^{d\times d}$ be a matrix with $\|\mathbf{v}\| = 1$. Suppose that $\|\mathbf{N}\|_2 \le \varphi$ for some $\varphi > 0$ satisfying $\varphi < |\lambda|/2$. Let $\widehat{\mathbf{v}}$ be the unit eigenvector of $\mathbf{A}$ corresponding to the largest eigenvalue in magnitude, then we have $|\mathbf{v}^\top \widehat{\mathbf{v}}|^2 \ge 1 - 2\varphi/|\lambda|$.*

*Proof.* See §B.2 for a detailed proof. □

To apply this lemma, we first note that, since $\widehat{S}_0$ is independent of $(y^{(i)}, \boldsymbol{w}^{(i)})$ for all $i \in \{n+1, \ldots, 2n\}$, we have

$$\mathbb{E}\big[(y^{(i)} - \mu) \cdot |\boldsymbol{\beta}_0^{*\top} \boldsymbol{w}^{(i)}|^2\big] = |\rho| \cdot |\boldsymbol{\beta}_0^{*\top}\boldsymbol{\beta}^*|^2 = |\rho| \cdot \big\|\boldsymbol{\beta}^*_{\widehat{S}_0}\big\|^2.$$

By the second part of (A.40) in the `Mean-Concentration` condition, we have

$$\left| \frac{1}{n} \sum_{i=n+1}^{2n} (y^{(i)} - \mu) \cdot |\boldsymbol{\beta}_0^{*\top} \boldsymbol{w}^{(i)}|^2 - |\rho| \cdot \big\|\boldsymbol{\beta}^*_{\widehat{S}_0}\big\|^2 \right| \le K \cdot \sqrt{\log(np)/n},$$

where $K > 0$ is a constant. By the triangle inequality, we have

$$\begin{aligned} \frac{1}{n} \sum_{i=n+1}^{2n} (y^{(i)} - \mu) \cdot |\boldsymbol{\beta}_0^{*\top} \boldsymbol{w}^{(i)}|^2 &\ge |\rho| \cdot \big\|\boldsymbol{\beta}^*_{\widehat{S}_0}\big\|^2 - K \cdot \sqrt{\log(np)/n} \\ &= |\rho| \cdot \big(\|\boldsymbol{\beta}^*\|^2 - \big\|\boldsymbol{\beta}^* - \boldsymbol{\beta}^*_{\widehat{S}_0}\big\|^2\big) - K \cdot \sqrt{\log(np)/n}. \end{aligned} \tag{A.20}$$

Combining (A.16) and (A.20), we obtain

$$\frac{1}{n} \sum_{i=n+1}^{2n} (y^{(i)} - \mu) \cdot |\boldsymbol{\beta}_0^{*\top} \boldsymbol{w}^{(i)}|^2 \ge |\rho| - (\gamma + C_\Psi + K) \cdot s\sqrt{\log(np)/n} \ge |\rho|/2,$$

where we use the fact that $n \ge Cs^2 \log(np)$ with constant $C$ sufficiently large. Applying Lemma A.1 with $\mathbf{v} = \boldsymbol{\beta}_0^*$, $\varphi = C_N \cdot s\sqrt{\log n/n}$ and $\lambda = |\rho|/2$ , we conclude that

$$|\langle \widehat{\boldsymbol{v}}, \boldsymbol{\beta}_0^* \rangle|^2 \ge 1 - 4C_N/|\rho| \cdot s\sqrt{\log(np)/n}. \tag{A.21}$$

In what follows, we bound the distance between $\widehat{\boldsymbol{v}}$ and $\boldsymbol{\beta}^*$. For notational simplicity, we define $\alpha_1 = \arccos(\langle \widehat{\boldsymbol{v}}, \boldsymbol{\beta}_0^* \rangle)$ and $\alpha_2 = \arccos(\langle \boldsymbol{\beta}_0^*, \boldsymbol{\beta}^* \rangle)$. Then we have

$$1 - |\langle \widehat{\boldsymbol{v}}, \boldsymbol{\beta}^* \rangle|^2 = \sin^2(\alpha_1 + \alpha_2) = [\sin(\alpha_1)\cos(\alpha_2) + \cos(\alpha_1)\sin(\alpha_2)]^2 \\ \leq 2\sin^2(\alpha_1) + 2\sin^2(\alpha_2). \tag{A.22}$$

Moreover, by (A.21), we have

$$\sin^2(\alpha_1) = 1 - |\langle \widehat{\boldsymbol{v}}, \boldsymbol{\beta}_0^* \rangle|^2 \leq 4C_N/|\rho| \cdot s\sqrt{\log(np)/n}. \tag{A.23}$$

In addition, by the definition of $\boldsymbol{\beta}_0^*$ in (A.17), we have

$$\sin^2(\alpha_2) = 1 - |\langle \boldsymbol{\beta}_0^*, \boldsymbol{\beta}^* \rangle|^2 = 1 - \|\boldsymbol{\beta}^*_{\widehat{S}_0}\|_2^2 = \|\boldsymbol{\beta}^* - \boldsymbol{\beta}^*_{\widehat{S}_0}\|^2 \leq (\gamma + C_\psi)/|\rho| \cdot s\sqrt{\log(np)/n}, \tag{A.24}$$

where the last inequality follows from (A.16). Finally, combining (A.22), (A.23), and (A.24), we conclude that, there exists a constant $C_{\text{init}}$ such that

$$|\langle \widehat{\boldsymbol{v}}, \boldsymbol{\beta}^* \rangle|^2 \geq 1 - C_{\text{init}}/|\rho| \cdot s\sqrt{\log(np)/n}. \tag{A.25}$$

Thus, we have that $\widehat{\boldsymbol{v}}$ is close to $\boldsymbol{\beta}^*$, which concludes this step.

**Step 4.** We conclude the proof for the lemma in this step. We first show that the approximation error of $\rho_n$ is small. Then, we bound $\text{dist}(\boldsymbol{\beta}^{(0)}, \overline{\boldsymbol{\beta}})$.

To begin with, we write $\widehat{\boldsymbol{v}} = \cos(\alpha_v) \cdot \boldsymbol{\beta}^* + \sin(\alpha_v) \cdot \boldsymbol{v}_1$, where $\boldsymbol{v}_1$ a unit vector perpendicular to $\boldsymbol{\beta}^*$ and $\cos(\alpha_v) = \widehat{\boldsymbol{v}}^\top \boldsymbol{\beta}^*$. Then, it hods that

$$\widehat{\boldsymbol{v}}\widehat{\boldsymbol{v}}^\top = \cos^2(\alpha_v) \cdot \boldsymbol{\beta}^* \boldsymbol{\beta}^{*\top} + \cos(\alpha_v)\sin(\alpha_v) \cdot (\boldsymbol{\beta}^* \boldsymbol{v}_1^\top + \boldsymbol{v}_1 \boldsymbol{\beta}^{*\top}) + \sin^2(\alpha_v) \cdot \boldsymbol{v}_1 \boldsymbol{v}_1^\top. \tag{A.26}$$

Combining the definition of $\rho_n$ in (2.12) and (A.26), we rewrite $\rho_n$ as

$$\rho_n = \frac{\cos^2(\alpha_v)}{n} \sum_{i=2n+1}^{3n} y^{(i)} \cdot [(\boldsymbol{x}^{(i)\top} \boldsymbol{\beta}^*)^2 - 1] + \frac{\sin^2(\alpha_v)}{n} \sum_{i=2n+1}^{3n} y^{(i)} \cdot [(\boldsymbol{x}^{(i)\top} \boldsymbol{v}_1)^2 - 1] \\ + \frac{2}{n}\cos(\alpha_v)\sin(\alpha_v) \cdot \sum_{i=2n+1}^{3n} y^{(i)} \cdot [(\boldsymbol{x}^{(i)\top} \boldsymbol{v}_1) \cdot (\boldsymbol{x}^{(i)\top} \boldsymbol{\beta}^*)]. \tag{A.27}$$

For the first term on the right-hand side of (A.27), recall that both $y^{(i)}$ and $[(\boldsymbol{x}^{(i)\top}\boldsymbol{\beta}^*)^2 - 1]$ are sub-exponential random variables. By Proposition 2.2 and Lemma C.4, we have

$$\left| \frac{1}{n} \sum_{i=2n+1}^{3n} y^{(i)} \cdot [(\boldsymbol{x}^{(i)\top} \boldsymbol{\beta}^*)^2 - 1] - \rho \right| \leq C\sqrt{\log n/n} \tag{A.28}$$

with probability at least $1 - 1/(30n)$, where $C$ is an absolute constant depending on $\Psi$. Similarly, for the second term, since $\boldsymbol{v}_1$ is perpendicular to $\boldsymbol{\beta}^*$, $\boldsymbol{X}^\top \boldsymbol{v}_1$ is independent of $Y$ and $\boldsymbol{X}^\top \boldsymbol{\beta}^*$. Thus, by Lemma C.4, with probability at least $1 - 1/(30n)$, we have

$$\left| \frac{1}{n} \sum_{i=2n+1}^{3n} y^{(i)} \cdot [(\boldsymbol{x}^{(i)\top} \boldsymbol{v}_1)^2 - 1] \right| \leq C\sqrt{\log n/n}. \tag{A.29}$$

Finally, for the third term on the right-hand side of (A.27), by Lemma C.4, with probability at least $1 - 1/(30n)$, we have

$$\left| \frac{1}{n} \sum_{i=2n+1}^{3n} y^{(i)} \cdot \left[ (\boldsymbol{x}^{(i)\top} \boldsymbol{v}_1) \cdot (\boldsymbol{x}^{(i)\top} \boldsymbol{\beta}^*) \right] \right| \leq C \sqrt{\log n / n}. \tag{A.30}$$

Combining (A.27), (A.28), (A.29), and (A.30), we obtain with probability at least $1 - 1/(10n)$ that

$$\begin{aligned} |\rho_n - \rho| &\leq \left[1 - \cos^2(\alpha_n)\right] \cdot |\rho| + \left[\cos^2(\alpha_v) + \sin^2(\alpha_v) + 2\cos(\alpha_v) \cdot \sin(\alpha_v)\right] \cdot C\sqrt{\log n / n} \\ &= (1 - |\widehat{\boldsymbol{v}}^\top \boldsymbol{\beta}^*|^2) \cdot |\rho| + 2C\sqrt{\log n / n}. \end{aligned} \tag{A.31}$$

Moreover, combining (A.31) and (A.25), we have

$$|\rho_n - \rho| \leq C_{\text{init}} \cdot s\sqrt{\log(np)/n} + 2C\sqrt{\log n / n} \leq K_1 \cdot s\sqrt{\log(np)/n}, \tag{A.32}$$

with probability at least $1 - 1/(5n)$, where $K_1$ is an absolute constant.

Finally, we bound the distance between $\boldsymbol{\beta}^{(0)}$ and $\overline{\boldsymbol{\beta}}$ defined in (2.1). Without loss of generality, we assume $\langle \widehat{\boldsymbol{v}}, \boldsymbol{\beta}^* \rangle \geq 0$ and let $\overline{\boldsymbol{\beta}} = \boldsymbol{\beta}^* \cdot \sqrt{|\rho|/2}$. By the triangle inequality, we have

$$\begin{aligned} \|\boldsymbol{\beta}^{(0)} - \overline{\boldsymbol{\beta}}\| &\leq \left\| \widehat{\boldsymbol{v}} \cdot \sqrt{|\rho_n|/2} - \boldsymbol{\beta}^* \cdot \sqrt{|\rho_n|/2} \right\| + \left\| \boldsymbol{\beta}^* \cdot \sqrt{|\rho_n|/2} - \boldsymbol{\beta}^* \cdot \sqrt{|\rho|/2} \right\| \\ &= \|\widehat{\boldsymbol{v}} - \boldsymbol{\beta}^*\| \cdot \sqrt{|\rho_n|/2} + \left| \sqrt{|\rho_n|/2} - \sqrt{|\rho|/2} \right| \\ &\leq \|\widehat{\boldsymbol{v}} - \boldsymbol{\beta}^*\| \cdot \sqrt{|\rho_n|/2} + \sqrt{|\rho_n - \rho|/2}, \end{aligned} \tag{A.33}$$

where the last inequality follows from the fact that $|\sqrt{x} - \sqrt{y}| \leq \sqrt{|x - y|}$ for all $x, y \geq 0$. Thus (A.33) implies that

$$\operatorname{dist}(\boldsymbol{\beta}^{(0)}, \overline{\boldsymbol{\beta}}) \leq \sqrt{|\rho_n|/2} \cdot \operatorname{dist}(\widehat{\boldsymbol{v}}, \boldsymbol{\beta}^*) + \sqrt{|\rho_n - \rho|/2}.$$

Moreover, by (A.25), we have

$$\begin{aligned} \|\widehat{\mathbf{v}} - \boldsymbol{\beta}^*\|^2 = 2 - 2\langle \widehat{\boldsymbol{v}}, \boldsymbol{\beta}^* \rangle &= \frac{2(1 - |\langle \widehat{\boldsymbol{v}}, \boldsymbol{\beta}^* \rangle|^2)}{1 + \langle \widehat{\boldsymbol{v}}, \boldsymbol{\beta}^* \rangle} \leq 2(1 - |\langle \widehat{\boldsymbol{v}}, \boldsymbol{\beta}^* \rangle|^2) \\ &\leq 2C_{\text{init}}/|\rho| \cdot s\sqrt{\log(np)/n}. \end{aligned} \tag{A.34}$$

Therefore, combining (A.32), (A.33), and (A.34), we conclude that

$$\operatorname{dist}(\boldsymbol{\beta}^{(0)}, \overline{\boldsymbol{\beta}}) \leq K_2 \left[s^2 \log(np)/n\right]^{1/4} \tag{A.35}$$

for some constant $K_2$.

For any fixed $\alpha \in (0, 1)$, we set $n \geq Cs^2 \log(np)$ for a sufficiently large $C$. By (A.32) and (A.35), if $K$ satisfies $K_1/\sqrt{C} \leq \alpha|\rho|$ and $K_2 \cdot C^{-1/4} \leq \alpha\sqrt{|\rho|/2}$, we have $|\rho_n - \rho| \leq \alpha|\rho|$ and $\operatorname{dist}(\boldsymbol{\beta}^{(0)}, \overline{\boldsymbol{\beta}}) \leq \alpha \cdot \|\overline{\boldsymbol{\beta}}\|$ with probability $1 - \mathcal{O}(1/n)$. Thus we conclude the proof of this lemma. $\square$

## A.2 Proof of Theorem 3.1

The proof of this theorem consists of two major steps. In the first step, let $S = \mathrm{supp}(\boldsymbol{\beta}^*)$, and let $\boldsymbol{z} \in \mathbb{R}^d$ be a point in the neighborhood of $\boldsymbol{\beta}^*$ such that $\mathrm{supp}(\boldsymbol{z}) \subseteq S$. We show that, starting from $\boldsymbol{z}$, the TWF update, restricted to $S$, moves toward the true solution $\overline{\boldsymbol{\beta}}$ (or $-\overline{\boldsymbol{\beta}}$) with high probability. In addition, in the second step, we show that the thresholding map keeps the TWF updates to be supported on $S$ with high probability.

To show the convergence, we first show that if the soft-thresholding $\mathcal{T}$ correctly filters out all the coordinates not in the support of $\boldsymbol{\beta}^*$, we obtain the desired convergence rate. We then show that $\mathcal{T}$ works as desired with high probability.

In what follows, to simplify the notation, we denote $\mathrm{supp}(\boldsymbol{\beta}^*)$ by $S$. Let $\boldsymbol{x}_S^{(i)} = r^{(i)}\boldsymbol{\beta}^* + \widetilde{\boldsymbol{x}}_S^{(i)}$ where $r^{(i)} = \boldsymbol{\beta}^{*\top}\boldsymbol{x}^{(i)}$. Since $\boldsymbol{x}_S$ is Gaussian, we have that $\widetilde{\boldsymbol{x}}_S^{(i)}$ and $r^{(i)}$ are independent. In addition, define $\mathcal{S} = \{\boldsymbol{v} \in \mathbb{R}^p : \mathrm{supp}(\boldsymbol{v}) \subseteq S\}$ as the set of vectors supported on $S$. We first show that restricted on $S$, the following generalized restricted isometry properties (RIP) hold for the data matrix with high probability.

**Lemma A.2** (`RIP`-condition)**.** *Let $\{(\boldsymbol{x}^{(i)}, y^{(i)}\}_{i\in[n]}$ be the $n$ independent observations generated from the misspecified phase retrieval model. Under Assumption 3.1, we define*

$$\boldsymbol{\Xi} = \sum_{i=1}^{n} \boldsymbol{x}_S^{(i)}\boldsymbol{x}_S^{(i)\top}, \quad \boldsymbol{\Phi} = \sum_{i=1}^{n} (y^{(i)} - \mu)\widetilde{\boldsymbol{x}}_S^{(i)}\widetilde{\boldsymbol{x}}_S^{(i)\top}, \tag{A.36}$$

*where $\widetilde{\boldsymbol{x}}_S^{(i)} = \boldsymbol{x}_S^{(i)} - (\boldsymbol{\beta}^{*\top}\boldsymbol{x}_i)\boldsymbol{\beta}^*$. When $n \geq Cs\log^2(ns) + \log^3 n$ with $C$ being a sufficiently large constant, with probability t least $1 - 1/(10n)$, the following inequalities hold*

$$\|\boldsymbol{\Xi} - n\mathbf{I}_S\|_2 \leq 3\sqrt{n}(\sqrt{s} + \sqrt{3\log n}), \qquad \|\boldsymbol{\Xi}\|_{2\to 4} \leq (3n)^{1/4} + \sqrt{s} + \sqrt{3\log n}, \tag{A.37}$$

$$\|\boldsymbol{\Xi}\|_{2\to 6} \leq (15n)^{1/6} + \sqrt{s} + \sqrt{3\log n}, \qquad \|\boldsymbol{\Phi}\|_2 \leq K_1 \cdot \sqrt{ns\log(ns)}, \tag{A.38}$$

$$\forall \boldsymbol{h} \in \mathcal{S} : \frac{1}{n}\sum_{j=1}^{n}(\boldsymbol{x}^{(j)\top}\boldsymbol{\beta}^*)^2(\boldsymbol{x}^{(j)\top}\boldsymbol{h})^2 \geq 2(\boldsymbol{\beta}^{*\top}\boldsymbol{h})^2 + 0.99\|\boldsymbol{\beta}^*\|^2\|\boldsymbol{h}\|^2. \tag{A.39}$$

*Here $K_1$ is an absolute constant that depends solely on $\Psi$, which is an upper bound of $\|Y\|_{\psi_1}$. The generalized `RIP`-condition assumes that (A.37), (A.38), and (A.39) hold.*

The proof of this lemma is given in §C. This lemma can be viewed generalization of the `RIP`-conditions (Candès and Tao, 2005). Let $\boldsymbol{A} = (\boldsymbol{x}^{(1)}, \boldsymbol{x}^{(2)}, \ldots, \boldsymbol{x}^{(n)})^\top$ be the design matrix. Note that $\boldsymbol{\Xi}$ defined in (A.36) can be written as $\boldsymbol{A}_S^\top\boldsymbol{A}_S$. The classical `RIP`-condition states that $\|n^{-1}\boldsymbol{\Xi} - \mathbf{I}_S\|_2 \leq \delta$ for some constant $\delta \in (0, 1)$, which is of the same type as the first inequality in (A.37). The remaining inequalities in our `RIP`-condition consider the higher-order isometries involving the restricted data $\{(\boldsymbol{x}_S^{(i)}, y^{(i)})\}_{i\in[n]}$. This condition is critical in bounding the errors for any vectors that are supported on $\mathrm{supp}(\boldsymbol{\beta}^*)$.

Next, we introduce an additional lemma that establishes upper bounds for a few empirical means.

**Lemma A.3** (`Mean-Concentration` condition). *Let $\{(\boldsymbol{x}^{(i)}, y^{(i)}\}_{i\in[n]}$ be the $n$ independent observations generated from the misspecified phase retrieval model. Under Assumption 3.1, there exists an absolute constant $K_2$ such that the following concentration inequalities hold with probability at least $1-1/(10n)$ when $n$ is sufficiently large.*

$$\left|\frac{1}{n}\sum_{i=1}^n y^{(i)} - \mu\right| \le K_2\sqrt{\frac{\log n}{n}}, \qquad \left|\frac{1}{n}\sum_{i=1}^n [y^{(i)}(\boldsymbol{x}^{(i)\top}\boldsymbol{\beta}^*)^2] - (\mu+\rho)\right| \le K_2\sqrt{\frac{\log n}{n}}, \tag{A.40}$$

$$\left\|\frac{1}{n}\sum_{i=1}^n (\boldsymbol{\beta}^{*\top}\boldsymbol{x}^{(i)})y^{(i)}\widetilde{\boldsymbol{x}}_S^{(i)}\right\| \le K_2\sqrt{\frac{s\log(ns)}{n}}, \quad \left\|\frac{1}{n}\sum_{i=1}^n [(\boldsymbol{x}^{(i)\top}\overline{\boldsymbol{\beta}})^3\boldsymbol{x}_S^{(i)}] - 3\|\overline{\boldsymbol{\beta}}\|^2\overline{\boldsymbol{\beta}}\right\| \le K_2\sqrt{\frac{s\log(ns)}{n}}, \tag{A.41}$$

$$\frac{1}{n}\sum_{i=1}^n\left(\sum_{r=0}^{10}|y^{(i)}|^r\right)\cdot\left(\sum_{w=0}^{10}|\boldsymbol{\beta}^{*\top}\boldsymbol{x}^{(i)}|^w\right) \le K_2, \quad \left\|\frac{1}{n}\sum_{i=1}^n(\boldsymbol{\beta}^{*\top}\boldsymbol{x}^{(i)})\widetilde{\boldsymbol{x}}_S^{(i)}\right\| \le K_2\sqrt{\frac{s\log(ns)}{n}}. \tag{A.42}$$

*Here we have $\mu = \mathbb{E}(Y)$ and $\rho = Cov[Y, (\boldsymbol{X}^\top\boldsymbol{\beta}^*)^2]$, and the sample size $n$ satisfies $n \ge C_1[s^2\log(np) + s\log^5 n]$ for some absolute constant $C_1$ depending on $K_2$ and $\Psi$. The* ***`Mean-Concentration`*** *condition assumes that inequalities (A.40), (A.41), and (A.42) hold.*

The proof of this lemma is presented in §C. This lemma bounds the deviation of a sequence of empirical means from their expected values.

In the following, to simplify the presentation, we condition on the event that the `RIP` and `Mean-Concentration` conditions hold, which is true with probability at least $1-1/(5n)$ by Lemmas A.2 and A.3.

Now, we are ready to prove Theorem 3.1. In particular, using the per iteration error established in Proposition 3.2, which is proved in §A.3, we prove the theorem by establishing a contraction result of our algorithm given a good initial solution $\boldsymbol{\beta}^{(0)}$, which yields the linear convergence of our algorithm.

*Proof of Theorem 3.1.* To facilitate our discussion, we denote by $\boldsymbol{A}$ the data matrix $(\boldsymbol{x}^{(1)}, \ldots, \boldsymbol{x}^{(n)})^\top \in \mathbb{R}^{n\times p}$. Let $\mathcal{E}_0$ be the event that the first $n$ data samples satisfy the `RIP` and `Mean-Concentration` conditions, which are defined in Lemma A.2 and Lemma A.3, respectively. Recall that our TWF algorithm starts with the initialization $\boldsymbol{\beta}^{(0)}$ defined in (2.12). Let $\mathcal{F}_0$ be the intersection of $\mathcal{E}_0$ and the event that $\boldsymbol{\beta}^{(0)} = \widetilde{\boldsymbol{\beta}}^{(0)}$, where $\widetilde{\boldsymbol{\beta}}^{(0)}$ is defined in (A.2). By Lemmas 3.1, A.2, and A.3, we have that $\mathcal{F}_0$ holds with probability at least $1-1/n$. Conditioning on $\mathcal{F}_0$, as shown in the proof of Lemma 3.1, $\boldsymbol{\beta}^{(0)}$ satisfies $\boldsymbol{\beta}^{(0)} \perp\!\!\!\perp \boldsymbol{A}_{S^c}$, $\mathrm{supp}(\boldsymbol{\beta}^{(0)}) \subseteq \mathrm{supp}(\boldsymbol{\beta}^*)$ and $\mathrm{dist}(\boldsymbol{\beta}^{(0)}, \overline{\boldsymbol{\beta}}) \le \alpha\|\overline{\boldsymbol{\beta}}\|$.

Furthermore, similar to the sequence $\{\boldsymbol{\beta}^{(k)}\}_{k\ge0}$, we define another trajectory $\{\widetilde{\boldsymbol{\beta}}^{(k)}\}_{k\ge0}$ as follows. For each $k = 0, 1, \ldots$, we define

$$\widetilde{\boldsymbol{\beta}}^{(k+1)} = \mathcal{T}_{\eta\tau(\widetilde{\boldsymbol{\beta}}^{(k)})}\left\{\widetilde{\boldsymbol{\beta}}^{(k)} - \eta[\nabla\ell_n(\widetilde{\boldsymbol{\beta}}^{(k)})]_S\right\}, \tag{A.43}$$

where $S = \mathrm{supp}(\boldsymbol{\beta}^*)$ and the threshold value $\tau(\widetilde{\boldsymbol{\beta}}^{(k)})$ is defined in (2.7). Together with the definition of $[\nabla \ell_n(\boldsymbol{z})]_S$ in (A.48), we obtain that $\widetilde{\boldsymbol{\beta}}^{(k)}$ is supported on $S$ and is independent of $\boldsymbol{A}_{S^c}$ for all $k \geq 0$. Moreover, for any $k \geq 0$, we define event $\mathcal{G}_{k+1}$ by

$$\mathcal{G}_{k+1} = \Big\{ \max_{j \notin S} |[\nabla \ell_n(\widetilde{\boldsymbol{\beta}}^{(k)})]_j| \leq \tau(\widetilde{\boldsymbol{\beta}}^{(k)}) \Big\}.$$

By definition, conditioning on the event $\mathcal{G}_{k+1}$, if $\widetilde{\boldsymbol{\beta}}^{(k)}$ is supported on $S$, the thresholding operator on $[p]$ is equivalent to that on $S$, i.e.,

$$\mathcal{T}_{\eta\tau(\widetilde{\boldsymbol{\beta}}^{k})}\Big[\widetilde{\boldsymbol{\beta}}^{(k)} - \eta \nabla \ell_n(\widetilde{\boldsymbol{\beta}}^{(k)})\Big] = \mathcal{T}_{\eta\tau(\widetilde{\boldsymbol{\beta}}^{k})}\Big\{\widetilde{\boldsymbol{\beta}}^{(k)} - \eta [\nabla \ell_n(\widetilde{\boldsymbol{\beta}}^{(k)})]_S\Big\}. \tag{A.44}$$

In addition, let $\mathcal{F}_{k+1} = \big(\bigcap_{i=0}^{k} \mathcal{G}_{i+1}\big) \bigcap \mathcal{F}_0$ for all $k \geq 1$. To prove the theorem, we show that, for any $k \geq 1$, conditioning on the event $\mathcal{F}_k$, we have

$$\widetilde{\boldsymbol{\beta}}^{(k)} \perp\!\!\!\perp \boldsymbol{A}_{S^c}, \quad \mathrm{supp}(\widetilde{\boldsymbol{\beta}}^{(k)}) \subseteq \mathrm{supp}(\boldsymbol{\beta}^*), \quad \widetilde{\boldsymbol{\beta}}^{(k)} = \boldsymbol{\beta}^{(k)}, \quad \text{and} \quad \mathrm{dist}(\widetilde{\boldsymbol{\beta}}^{(k)}, \overline{\boldsymbol{\beta}}) \leq \alpha \|\overline{\boldsymbol{\beta}}\|. \tag{A.45}$$

Moreover, the probability of the event $\mathcal{F}_k$ satisfies

$$\mathbb{P}(\mathcal{F}_0 \backslash \mathcal{F}_k) \leq 1 - k/(n^2 p). \tag{A.46}$$

We prove the claim via induction. First, the claim holds for $k = 0$ as we discussed in the beginning of the proof. Next, suppose (A.45) and (A.46) holds for some $\mathcal{F}_k$ with $k \geq 1$. We show that the claim also holds for $\mathcal{F}_{k+1}$. Again, for simplicity we only show the proof under the condition that $\|\boldsymbol{\beta}^{(k)} - \overline{\boldsymbol{\beta}}\| \leq \alpha \|\overline{\boldsymbol{\beta}}\|$, and the other case follows by similar arguments. Since $\mathcal{F}_{k+1} \subseteq \mathcal{F}_k \subseteq \mathcal{F}_{k-1} \subseteq \ldots \subseteq \mathcal{F}_0$, when $\mathcal{F}_k$ holds, we have $\widetilde{\boldsymbol{\beta}}^{(k)} = \boldsymbol{\beta}^{(k)}$. Thus, conditioning on $\mathcal{F}_{k+1}$, by (A.43) and (A.44) we have

$$\widetilde{\boldsymbol{\beta}}^{(k+1)} = \mathcal{T}_{\eta\tau(\widetilde{\boldsymbol{\beta}}^{k})}\Big\{\widetilde{\boldsymbol{\beta}}^{(k)} - \eta [\nabla \ell_n(\widetilde{\boldsymbol{\beta}}^{(k)})]_S\Big\} = \mathcal{T}_{\eta\tau(\widetilde{\boldsymbol{\beta}}^{k})}\Big[\widetilde{\boldsymbol{\beta}}^{(k)} - \eta \nabla \ell_n(\widetilde{\boldsymbol{\beta}}^{(k)})\Big] = \boldsymbol{\beta}^{(k+1)}.$$

Moreover, by Proposition 3.2, we have

$$\|\boldsymbol{\beta}^{(k+1)} - \overline{\boldsymbol{\beta}}\| = \|\widetilde{\boldsymbol{\beta}}^{(k+1)} - \overline{\boldsymbol{\beta}}\| \leq (1 - \eta\rho)\|\boldsymbol{\beta}^{(k)} - \overline{\boldsymbol{\beta}}\| + C\sqrt{s \log(np)/n}, \tag{A.47}$$

where $C > 0$ is a constant. Since $\eta\rho > 0$ is a constant, when $n$ is sufficiently large such that

$$C\sqrt{s \log(np)/n} \leq \eta\alpha\rho \cdot \|\overline{\boldsymbol{\beta}}\|,$$

by (A.47) we have $\mathrm{dist}(\widetilde{\boldsymbol{\beta}}^{(k+1)}, \overline{\boldsymbol{\beta}}) \leq \alpha \|\overline{\boldsymbol{\beta}}\|$. Therefore, conditioning on $\mathcal{F}_{k+1}$, we conclude that (A.45) holds for $k + 1$.

It remains to prove (A.46) holds for $\mathcal{F}_{k+1}$. By the definition of $\mathcal{F}_{k+1}$, it suffices to show that

$$\mathbb{P}(\mathcal{F}_k \setminus \mathcal{F}_{k+1}) = \mathbb{P}(\mathcal{F}_k \setminus \mathcal{G}_{k+1}) \leq 1/(n^2 p).$$

We consider $[\nabla \ell_n(\widetilde{\boldsymbol{\beta}}^{(k)})]_j$ for any each $j \notin S$. By the definition of $\nabla \ell_n(\boldsymbol{\beta})$ in (2.9), we have

$$[\nabla \ell_n(\widetilde{\boldsymbol{\beta}}^{(k)})]_j = -\frac{4}{n}\sum_{i=1}^{n}\big[y^{(i)} - (\boldsymbol{x}^{(i)\top}\widetilde{\boldsymbol{\beta}}^{(k)})^2 - \xi_n(\widetilde{\boldsymbol{\beta}}^{(k)})\big] \cdot (\boldsymbol{x}^{(i)\top}\widetilde{\boldsymbol{\beta}}^{(k)}) \cdot \boldsymbol{x}_j^{(i)},$$

where $\boldsymbol{x}_j^{(i)}$ is the $j$-th entry of $\boldsymbol{x}^{(i)}$. Since $\widetilde{\boldsymbol{\beta}}^{(k)}$ is supported on $S$, whereas $j \notin S$, $\boldsymbol{x}_j^{(i)}$ is independent of

$$[y^{(i)} - (\boldsymbol{x}^{(i)\top}\widetilde{\boldsymbol{\beta}}^{(k)})^2 - \xi_n(\widetilde{\boldsymbol{\beta}}^{(k)})] \cdot (\boldsymbol{x}^{(i)\top}\widetilde{\boldsymbol{\beta}}^{(k)}),$$

which is denoted by $\varphi_i$ hereafter for notational simplicity. Therefore, conditioning on $\{\varphi_i\}_{i\in[n]}$, $[\nabla \ell_n(\widetilde{\boldsymbol{\beta}}^{(k)})]_j$ is a centered Gaussian random variable with variance $16/(n^2) \cdot \sum_{i=1}^n \varphi_i^2$. Thus, with probability at least $1 - 1/(n^2p^2)$, we have

$$\left|[\nabla \ell_n(\widetilde{\boldsymbol{\beta}}^{(k)})]_j\right| \leq \left[\frac{80\log(np)}{n^2}\sum_{i=1}^n \varphi_i^2\right]^{1/2} \leq \tau(\widetilde{\boldsymbol{\beta}}^k),$$

when the constant $\kappa$ in (2.7) is sufficiently large that $\kappa \geq \sqrt{80}$. Here the inequality is obtained by the tail probability of the Gaussian distribution.

Taking a union bound for all $j \notin S$, we obtain that $\mathbb{P}(\mathcal{G}_{k+1}^c \,|\, \mathcal{F}_k) \leq 1/(n^2p)$. Thus we have that

$$\mathbb{P}(\mathcal{F}_k \setminus \mathcal{F}_{k+1}) = \mathbb{P}(\mathcal{F}_k \setminus \mathcal{G}_{k+1}) = \mathbb{P}(\mathcal{G}_{k+1}^c \,|\, \mathcal{F}_k) \cdot \mathbb{P}(\mathcal{F}_k) \leq \mathbb{P}(\mathcal{G}_{k+1}^c \,|\, \mathcal{F}_k) \leq 1/(n^2p),$$

which concludes the proof. □

### A.3 Proof of Proposition 3.2

*Proof.* Recall that we denote $\operatorname{supp}(\boldsymbol{\beta}^*)$ by $S$ for simplicity, and we let $\mathcal{S}$ be the set of vectors in $\mathbb{R}^p$ supported on $S$. Throughout the proof, we assume that the `RIP` and `Mean-Concentration` conditions hold. Note that by Lemmas A.2 and A.3, these conditions hold with probability at least $1 - 1/(5n)$.

For any $\boldsymbol{z} \in \mathcal{S}$, recall that $\operatorname{dist}(\boldsymbol{z}, \overline{\boldsymbol{\beta}}) = \min(\|\boldsymbol{z} - \overline{\boldsymbol{\beta}}\|, \|\boldsymbol{z} + \overline{\boldsymbol{\beta}}\|)$. For ease of presentation, we prove the proposition under the assumption that $\|\boldsymbol{z} - \overline{\boldsymbol{\beta}}\| \leq \alpha\|\overline{\boldsymbol{\beta}}\|$. The proof for the other case where $\|\boldsymbol{z} + \overline{\boldsymbol{\beta}}\| \leq \alpha\|\overline{\boldsymbol{\beta}}\|$ follows similarly. Moreover, we assume $\rho > 0$ without loss of generality. In addition, by the definition of $\ell_n(\cdot)$ in (2.6), since $\operatorname{supp}(\boldsymbol{z}) = S$, we have

$$[\nabla \ell_n(\boldsymbol{z})]_S = \frac{4}{n}\sum_{i=1}^n [y^{(i)} - (\boldsymbol{x}^{(i)\top}\boldsymbol{z})^2 - \xi_n(\boldsymbol{z})]\,(\mathbf{I}_p - \boldsymbol{x}_S^{(i)}\boldsymbol{x}_S^{(i)\top})\boldsymbol{z}, \tag{A.48}$$

where $\xi_n(\boldsymbol{z}) = \mu_n - \|\boldsymbol{z}\|^2$. Here $[\nabla \ell_n(\boldsymbol{z})]_S$ is the restriction of $\nabla \ell_n(\boldsymbol{z})$ on the index set $S$.

Recall that we define the threshold operator $\tau(\boldsymbol{z})$ in (2.7). Then, we write the truncated gradient update $t(\boldsymbol{z})$ as

$$t(\boldsymbol{z}) = \mathcal{T}_{\eta\tau(\boldsymbol{z})}\{\boldsymbol{z} - \eta \cdot [\nabla \ell_n(\boldsymbol{z})]_S\} = \boldsymbol{z} - \eta \cdot [\nabla \ell_n(\boldsymbol{z})]_S + \eta\tau(\boldsymbol{z}) \cdot \boldsymbol{v},$$

for some vector $\boldsymbol{v} \in \mathbb{R}^p$ with $\operatorname{supp}(\boldsymbol{v}) \subset S$ and $\|\boldsymbol{v}\|_\infty \leq 1$. Let $\boldsymbol{h} = \boldsymbol{z} - \overline{\boldsymbol{\beta}} \in \mathbb{R}^p$. By our assumptions, we have $\operatorname{supp}(\boldsymbol{h}) \subseteq S$ and $\|\boldsymbol{h}\| \leq \alpha\|\overline{\boldsymbol{\beta}}\|$. By the triangle inequality, we have

$$\|t(\boldsymbol{z}) - \overline{\boldsymbol{\beta}}\| \leq \|\boldsymbol{h} - \eta \cdot [\nabla \ell_n(\boldsymbol{z})]_S\| + \eta\tau(\boldsymbol{z}) \cdot \|\boldsymbol{v}\| = R_1 + R_2, \tag{A.49}$$

where $R_1 = \|\boldsymbol{h} - \eta \cdot [\nabla \ell_n(\boldsymbol{z})]_S\|$ and $R_2 = \eta\tau(\boldsymbol{z}) \cdot \|\boldsymbol{v}\|$. In the sequel, we derive an upper bound for $\|t(\boldsymbol{z}) - \overline{\boldsymbol{\beta}}\|$. The proof is decomposed into four steps, and we bound the residuals $R_1$ and $R_2$ separately.

**Step 1.** In the first step, we expand $[\nabla \ell_n(\boldsymbol{z})]_S$ in (A.48) into several terms, which will be bounded separately in the next step. In particular, by the definition of $\nabla \ell_n(\boldsymbol{z})$, we have

$$
\begin{aligned}
[\nabla \ell_n(\boldsymbol{z})]_S &= \frac{4}{n}\sum_{i=1}^n \big[y^{(i)} - (\boldsymbol{x}^{(i)\top}\boldsymbol{z})^2 - \mu + \|\boldsymbol{z}\|^2\big]\big[\boldsymbol{z} - (\boldsymbol{x}_S^{(i)\top}\boldsymbol{z})\boldsymbol{x}_S^{(i)}\big] + \frac{4\Delta\mu}{n}\sum_{i=1}^n \big[\boldsymbol{z} - (\boldsymbol{x}_S^{(i)\top}\boldsymbol{z})\boldsymbol{x}_S^{(i)}\big] \\
&= \boldsymbol{t}_1 + \boldsymbol{t}_2 + \boldsymbol{t}_3 + \boldsymbol{t}_4,
\end{aligned}
$$

where $\Delta\mu = \mu_n - \mu$ and

$$
\begin{aligned}
\boldsymbol{t}_1 &= \frac{4}{n}\sum_{i=1}^n \big[y^{(i)} - (\boldsymbol{x}^{(i)\top}\boldsymbol{z})^2 - \mu + \|\boldsymbol{z}\|^2\big] \cdot \boldsymbol{z} \quad & \boldsymbol{t}_2 &= -\frac{4}{n}\sum_{i=1}^n \big[y^{(i)} - (\boldsymbol{x}^{(i)\top}\boldsymbol{z})^2\big](\boldsymbol{x}_S^{(i)\top}\boldsymbol{z}) \cdot \boldsymbol{x}_S^{(i)}, \\
\boldsymbol{t}_3 &= -\frac{4}{n}\sum_{i=1}^n (-\mu + \|\boldsymbol{z}\|^2)(\boldsymbol{x}_S^{(i)\top}\boldsymbol{z}) \cdot \boldsymbol{x}_S^{(i)}, & \boldsymbol{t}_4 &= \frac{4\Delta\mu}{n}\sum_{i=1}^n \big[\boldsymbol{z} - (\boldsymbol{x}_S^{(i)\top}\boldsymbol{z})\boldsymbol{x}_S^{(i)}\big].
\end{aligned} \tag{A.50}
$$

Let $\boldsymbol{A}_S = \sum_{i=1}^n \boldsymbol{x}_S^{(i)}\boldsymbol{x}_S^{(i)\top}$. By the first part of (A.37) in the `RIP`-condition, we have

$$
\forall \boldsymbol{z} \in \mathcal{S} : \left| \frac{1}{n}\sum_{i=1}^n (\boldsymbol{x}^{(i)\top}\boldsymbol{z})^2 - \|\boldsymbol{z}\|^2 \right| \le \|\boldsymbol{A}_S/n - \mathbf{I}_S\|_2 \cdot \|\boldsymbol{z}\|^2 \le \Big(\sqrt{s/n} + \sqrt{3\log n/n}\Big) \cdot \|\boldsymbol{z}\|^2. \tag{A.51}
$$

Note that we can write $\boldsymbol{t}_1$ in (A.50) as

$$
\boldsymbol{t}_1 = \frac{4}{n}\sum_{i=1}^n \big[\|\boldsymbol{z}\|^2 - (\boldsymbol{x}^{(i)\top}\boldsymbol{z})^2\big] \cdot \boldsymbol{z} + 4\Delta\mu \cdot \boldsymbol{z}. \tag{A.52}
$$

Also note that $|\Delta\mu| \le K_2\sqrt{\log n/n}$ for some constant $K_2$ by the `Mean-Concentration` condition (the first part of (A.40)). Then, by (A.51), (A.52), and the triangle inequality, for all $\boldsymbol{v} \in \mathbb{R}^p$, we have

$$
\begin{aligned}
|\boldsymbol{t}_1^\top \boldsymbol{v}| &\le \left| \frac{4}{n}\sum_{i=1}^n \big[\|\boldsymbol{z}\|^2 - (\boldsymbol{x}^{(i)\top}\boldsymbol{z})^2\big] \right| \cdot |\boldsymbol{z}^\top \boldsymbol{v}| + 4|\Delta\mu| \cdot |\boldsymbol{z}^\top \boldsymbol{v}| \\
&\le 4\Big[K_2\sqrt{\log n/n} + \Big(\sqrt{s/n} + \sqrt{3\log n/n}\Big) \cdot \|\boldsymbol{z}\|^2\Big] \cdot |\boldsymbol{z}^\top \boldsymbol{v}|.
\end{aligned} \tag{A.53}
$$

Next we turn to the second term $\boldsymbol{t}_2$ in (A.50). We first let $\boldsymbol{x}^{(i)} = r^{(i)}\boldsymbol{\beta}^* + \widetilde{\boldsymbol{x}}^{(i)}$, where $r^{(i)} = \boldsymbol{x}^{(i)\top}\boldsymbol{\beta}^*$. Then by expanding the square, we obtain

$$
\boldsymbol{x}^{(i)}\boldsymbol{x}^{(i)\top} = (r^{(i)})^2\boldsymbol{\beta}^*\boldsymbol{\beta}^{*\top} + r^{(i)}\big(\boldsymbol{\beta}^*\widetilde{\boldsymbol{x}}^{(i)\top} + \widetilde{\boldsymbol{x}}^{(i)}\boldsymbol{\beta}^{*\top}\big) + \widetilde{\boldsymbol{x}}^{(i)}\widetilde{\boldsymbol{x}}^{(i)\top}. \tag{A.54}
$$

Since $\boldsymbol{z}$ is supported on $S$, by definition, we can write $\boldsymbol{t}_2$ as

$$
\boldsymbol{t}_2 = \frac{4}{n}\sum_{i=1}^n \big[(\boldsymbol{x}^{(i)\top}\boldsymbol{z})^3\boldsymbol{x}_S^{(i)} - y^{(i)}(\boldsymbol{x}^{(i)\top}\boldsymbol{z})\boldsymbol{x}_S^{(i)}\big].
$$

To simplify the notation, we define

$$\boldsymbol{t}_{21} = \frac{4}{n}\sum_{i=1}^{n}(\boldsymbol{x}^{(i)\top}\boldsymbol{z})^3\boldsymbol{x}_S^{(i)}, \quad \boldsymbol{t}_{22} = -\frac{4}{n}\sum_{i=1}^{n} y^{(i)}(\boldsymbol{x}^{(i)\top}\boldsymbol{z})\boldsymbol{x}_S^{(i)}. \tag{A.55}$$

Then, we have $\boldsymbol{t}_2 = \boldsymbol{t}_{21} + \boldsymbol{t}_{22}$. Note that $\boldsymbol{z} = \boldsymbol{h} + \overline{\boldsymbol{\beta}}$. We expand $\boldsymbol{t}_{21}$ in (A.55) by

$$\boldsymbol{t}_{21} = \frac{4}{n}\sum_{i=1}^{n}\Big[\big(\boldsymbol{x}^{(i)\top}\overline{\boldsymbol{\beta}}\big)^3 + 3\big(\boldsymbol{x}^{(i)\top}\overline{\boldsymbol{\beta}}\big)^2\big(\boldsymbol{x}^{(i)\top}\boldsymbol{h}\big) + 3\big(\boldsymbol{x}^{(i)\top}\overline{\boldsymbol{\beta}}\big)\big(\boldsymbol{x}^{(i)\top}\boldsymbol{h}\big)^2 + \big(\boldsymbol{x}^{(i)\top}\boldsymbol{h}\big)^3\Big]\boldsymbol{x}_S^{(i)}. \tag{A.56}$$

In addition, for the second term $\boldsymbol{t}_{22}$ in (A.55), by (A.54), we have

$$\begin{aligned}\boldsymbol{t}_{22} = &-\frac{4}{n}\sum_{i=1}^{n}\big[y^{(i)}(r^{(i)})^2(\boldsymbol{\beta}^{*\top}\boldsymbol{z})\boldsymbol{\beta}^*\big] - \frac{4}{n}\sum_{i=1}^{n} r^{(i)}y^{(i)}\cdot\big[(\boldsymbol{\beta}^{*\top}\boldsymbol{z})\widetilde{\boldsymbol{x}}_S^{(i)} + (\widetilde{\boldsymbol{x}}_S^{(i)\top}\boldsymbol{z})\boldsymbol{\beta}^*\big]\\ &-\frac{4}{n}\sum_{i=1}^{n}\big[y^{(i)}(\widetilde{\boldsymbol{x}}_S^{(i)\top}\boldsymbol{z})\widetilde{\boldsymbol{x}}_S^{(i)}\big].\end{aligned} \tag{A.57}$$

Now we conclude this step. In summary, we decompose $[\nabla\ell_n(\boldsymbol{z})]_S$ into $\boldsymbol{t}_1 + \boldsymbol{t}_2 + \boldsymbol{t}_3 + \boldsymbol{t}_4$ and we further bound $\boldsymbol{t}_1$ in (A.53) and expand $\boldsymbol{t}_2$ in (A.55), (A.56), and (A.57).

**Step 2.** By definition, we have

$$R_1^2 = \|\boldsymbol{h} - \eta\cdot[\nabla\ell_n(\boldsymbol{z})]_S\|^2 = \|\boldsymbol{h}\|^2 - \eta\cdot\boldsymbol{h}^\top[\nabla\ell_n(\boldsymbol{z})]_S + \eta^2\cdot\|[\nabla\ell_n(\boldsymbol{z})]_S\|^2.$$

In this step, we first obtain an lower bound for $\boldsymbol{h}^\top[\nabla\ell_n(\boldsymbol{z})]_S$, and we upper bound $\big\|[\nabla\ell_n(\boldsymbol{z})]_S\big\|^2$ in the next step. Note that, as shown in **Step 1**, we have $\boldsymbol{h}^\top[\nabla\ell_n(\boldsymbol{z})]_S = \boldsymbol{h}^\top(\boldsymbol{t}_1 + \boldsymbol{t}_2 + \boldsymbol{t}_3 + \boldsymbol{t}_4)$.

First, by applying the bound for $\boldsymbol{t}_1$ in (A.53), we have

$$|\boldsymbol{h}^\top\boldsymbol{t}_1| \le 4\Big[K_2\sqrt{\log n/n} + \Big(\sqrt{s/n} + \sqrt{3\log n/n}\Big)\cdot\|\boldsymbol{z}\|^2\Big]\cdot|\boldsymbol{z}^\top\boldsymbol{h}| \le C\sqrt{s\log n/n}, \tag{A.58}$$

for some absolute constant $C$ depending only on $\Psi$, $\rho$, $\mu$ and $\alpha$. Here we assume that $n \ge K[s^2\log(np) + s\log^5 n]$ with some sufficiently large constant $K$. Note that here in the last inequality of (A.58), we use the fact that both $\boldsymbol{z}$ and $\boldsymbol{h}$ are bounded, which follows from $\|\boldsymbol{h}\| \le \alpha\|\overline{\boldsymbol{\beta}}\|$ and $\|\boldsymbol{z}\| \le (1+\alpha)\|\overline{\boldsymbol{\beta}}\|$.

In addition, for $\boldsymbol{h}^\top\boldsymbol{t}_{21}$ and $\boldsymbol{h}^\top\boldsymbol{t}_{22}$, we apply (A.56) and (A.57) and bound each term. For the first term in the expansion of $\boldsymbol{t}_{21}$ in (A.56), by the second part of (A.41) in the `Mean-Concentration` condition, we have

$$\bigg\|\frac{4}{n}\sum_{i=1}^{n}\big(\boldsymbol{x}^{(i)\top}\overline{\boldsymbol{\beta}}\big)^3\boldsymbol{x}_S^{(i)} - 12\|\overline{\boldsymbol{\beta}}\|^2\cdot\overline{\boldsymbol{\beta}}\bigg\| \le K_2\sqrt{s\log(ns)/n}, \tag{A.59}$$

which implies that

$$\bigg|\frac{4}{n}\sum_{i=1}^{n}\big(\boldsymbol{x}^{(i)\top}\overline{\boldsymbol{\beta}}\big)^3\cdot\big(\boldsymbol{h}^\top\boldsymbol{x}_S^{(i)}\big) - 12\|\overline{\boldsymbol{\beta}}\|^2\cdot\big(\overline{\boldsymbol{\beta}}^\top\boldsymbol{h}\big)\bigg| \le C\sqrt{s\log(ns)/n} \tag{A.60}$$

for some absolute constant $C$.

Next, for the second term on the right-hand side of (A.56), by (A.39) in the RIP-condition, we obtain

$$\frac{12}{n}\sum_{i=1}^{n}(\boldsymbol{x}^{(i)\top}\overline{\boldsymbol{\beta}})^2(\boldsymbol{x}^{(i)\top}\boldsymbol{h})^2 \geq 24(\overline{\boldsymbol{\beta}}^\top\boldsymbol{h})^2 + 11.88\|\overline{\boldsymbol{\beta}}\|^2\|\boldsymbol{h}\|^2. \tag{A.61}$$

Furthermore, by Hölder's inequality and the first part of (A.37) in the RIP-condition, when $n$ is sufficiently large, for the third term on the right-hand side of (A.56), we have

$$\begin{aligned}\left|\frac{12}{n}\sum_{j=1}^{n}(\boldsymbol{x}^{(i)\top}\overline{\boldsymbol{\beta}})(\boldsymbol{x}^{(i)\top}\boldsymbol{h})^3\right| &\leq \frac{12}{n}\left[\sum_{j=1}^{n}(\boldsymbol{x}^{(i)\top}\overline{\boldsymbol{\beta}})^4\right]^{1/4}\left[\sum_{j=1}^{n}(\boldsymbol{x}^{(i)\top}\boldsymbol{h})^4\right]^{3/4} \\ &\leq \frac{12}{n}\cdot\|\boldsymbol{A}_S\|_{2\to 4}^4\cdot\|\overline{\boldsymbol{\beta}}\|\|\boldsymbol{h}\|^3 \leq 37\|\overline{\boldsymbol{\beta}}\|\|\boldsymbol{h}\|^3.\end{aligned} \tag{A.62}$$

Finally, for the last term in the expansion of $\boldsymbol{t}_{21}$, we have

$$\boldsymbol{h}^\top\left[\frac{4}{n}\sum_{i=1}^{n}(\boldsymbol{x}^{(i)\top}\boldsymbol{h})^3\cdot\boldsymbol{x}_S^{(i)}\right] = \frac{4}{n}\sum_{j=1}^{n}(\boldsymbol{x}^{(i)\top}\boldsymbol{h})^4 \geq 0. \tag{A.63}$$

Thus, combining (A.60), (A.61), (A.62), and (A.63), we obtain

$$\begin{aligned}\boldsymbol{h}^\top\boldsymbol{t}_{21} \geq &12\|\overline{\boldsymbol{\beta}}\|^2\cdot(\overline{\boldsymbol{\beta}}^\top\boldsymbol{h}) + 24(\overline{\boldsymbol{\beta}}^\top\boldsymbol{h})^2 + 11.88\|\overline{\boldsymbol{\beta}}\|^2\|\boldsymbol{h}\|^2 \\ &- (K_2 + C)\sqrt{s\log(ns)/n} - 37\|\overline{\boldsymbol{\beta}}\|\|\boldsymbol{h}\|^3.\end{aligned} \tag{A.64}$$

Next, we lower bound $\boldsymbol{h}^\top\boldsymbol{t}_{22}$. For the first term on the right-hand side of (A.57), by the second part of (A.40) in the Mean-Concentration condition, we have

$$\left|\frac{4}{n}\sum_{i=1}^{n}y^{(i)}(r^{(i)})^2(\boldsymbol{\beta}^{*\top}z)\cdot(\boldsymbol{\beta}^{*\top}\boldsymbol{h}) - 4(\rho+\mu)\cdot(\boldsymbol{\beta}^{*\top}\boldsymbol{z})\cdot(\boldsymbol{\beta}^{*\top}\boldsymbol{h})\right| \leq C\sqrt{\log n/n}, \tag{A.65}$$

where $C$ is an absolute constant. In addition, for the second term in the expansion of $\boldsymbol{t}_{22}$, by the first part of (A.41), we have

$$\left|\frac{4}{n}\sum_{i=1}^{n}\left[r^{(i)}y^{(i)}(\boldsymbol{\beta}^{*\top}\boldsymbol{z})\cdot(\widetilde{\boldsymbol{x}}_S^{(i)\top}\boldsymbol{h}) + y^{(i)}r^{(i)}(\widetilde{\boldsymbol{x}}_S^{(i)\top}\boldsymbol{z})(\boldsymbol{\beta}^{*\top}\boldsymbol{h})\right]\right| \leq C\sqrt{s\log(ns)/n} \tag{A.66}$$

for some constant $C$.

Meanwhile, we combine the last term of $\boldsymbol{t}_{22}$ with $\boldsymbol{t}_3$ to get

$$\boldsymbol{t}_3' = \boldsymbol{t}_3 - \frac{4}{n}\sum_{i=1}^{n}[y^{(i)}(\widetilde{\boldsymbol{x}}_S^{(i)\top}\boldsymbol{z})\widetilde{\boldsymbol{x}}_S^{(i)}] = -\frac{4}{n}\sum_{i=1}^{n}(-\mu+\|\boldsymbol{z}\|^2)(\boldsymbol{x}_S^{(i)\top}\boldsymbol{z})\boldsymbol{x}_S^{(i)} - \frac{4}{n}\sum_{i=1}^{n}[y^{(i)}(\widetilde{\boldsymbol{x}}_S^{(i)\top}\boldsymbol{z})\widetilde{\boldsymbol{x}}_S^{(i)}].$$

By (A.54), we rewrite $\boldsymbol{t}_3'$ as

$$\begin{aligned}\boldsymbol{t}_3' = &-\frac{4}{n}\sum_{i=1}^{n}[(y^{(i)}-\mu)(\widetilde{\boldsymbol{x}}_S^{(i)\top}\boldsymbol{z})\widetilde{\boldsymbol{x}}_S^{(i)}] + \frac{4}{n}\sum_{i=1}^{n}\mu r^{(i)}\cdot[(\widetilde{\boldsymbol{x}}_S^{(i)\top}\boldsymbol{z})\boldsymbol{\beta}^* + (\boldsymbol{\beta}^{*\top}\boldsymbol{z})\widetilde{\boldsymbol{x}}_S^{(i)}] \\ &+ \frac{4}{n}\sum_{i=1}^{n}\mu(r^{(i)})^2\cdot(\boldsymbol{\beta}^{*\top}\boldsymbol{z})\boldsymbol{\beta}^* - \frac{4}{n}\sum_{i=1}^{n}\|\boldsymbol{z}\|^2\cdot(\boldsymbol{x}_S^{(i)\top}\boldsymbol{z})\boldsymbol{x}_S^{(i)}.\end{aligned} \tag{A.67}$$

Note that $\widetilde{\boldsymbol{x}}^{(i)\top}\boldsymbol{\beta}^* = 0$ by the definition of $\widetilde{\boldsymbol{x}}^{(i)}$. Then for the first term in (A.67), we have

$$-\frac{4}{n}\sum_{i=1}^{n}\big[(y^{(i)}-\mu)(\widetilde{\boldsymbol{x}}_S^{(i)\top}\boldsymbol{z})\widetilde{\boldsymbol{x}}_S^{(i)}\big] = -\frac{4}{n}\sum_{i=1}^{n}\big[(y^{(i)}-\mu)\widetilde{\boldsymbol{x}}_S^{(i)}\widetilde{\boldsymbol{x}}_S^{(i)\top}\big]\boldsymbol{h}.$$

By the second part of (A.38) in the `RIP`-condition, for sufficiently large $n$, there exists a constant $C'$ such that

$$\begin{aligned}&\bigg|-\frac{4}{n}\sum_{i=1}^{n}\boldsymbol{h}^\top\big[(y^{(i)}-\mu)(\widetilde{\boldsymbol{x}}_S^{(i)\top}\boldsymbol{z})\widetilde{\boldsymbol{x}}_S^{(i)}\big]\bigg| \\ &\quad\le \bigg\|\frac{4}{n}\sum_{i=1}^{n}\big[(y^{(i)}-\mu)\widetilde{\boldsymbol{x}}_S^{(i)}\widetilde{\boldsymbol{x}}_S^{(i)\top}\big]\bigg\|_2\cdot\|\boldsymbol{h}\|\cdot\|\boldsymbol{z}\| \le C'\sqrt{s\log(np)/n}. \end{aligned} \tag{A.68}$$

where the last inequality holds when $n$ is sufficiently large. For the second term on the right-hand side of (A.67), recall that we define $r^{(i)} = \boldsymbol{x}^{(i)\top}\boldsymbol{\beta}^*$. By the triangle inequality and the second part of (A.42) in `Mean-Concentration` condition, we have

$$\begin{aligned}&\bigg\|\frac{4}{n}\sum_{i=1}^{n}\mu r^{(i)}\big[(\widetilde{\boldsymbol{x}}_S^{(i)\top}\boldsymbol{z})\boldsymbol{\beta}^* + (\boldsymbol{\beta}^{*\top}\boldsymbol{z})\widetilde{\boldsymbol{x}}_S^{(i)}\big]\bigg\| \\ &\quad\le \bigg\|\frac{4}{n}\sum_{i=1}^{n}\mu r^{(i)}\widetilde{\boldsymbol{x}}_S^{(i)}\bigg\|\cdot\big(\|\boldsymbol{z}\boldsymbol{\beta}^{*\top}\|_2 + |\boldsymbol{\beta}^{*\top}\boldsymbol{z}|\big) \le C'\sqrt{s\log(ns)/n} \end{aligned} \tag{A.69}$$

for some absolute constant $C'$. Finally, for the last two terms in (A.67), by the first part of (A.37) of the `RIP`-condition, we have

$$\begin{aligned}&\bigg\|\frac{4}{n}\sum_{i=1}^{n}\mu(r^{(i)})^2\cdot(\boldsymbol{\beta}^{*\top}\boldsymbol{z})\boldsymbol{\beta}^* - \frac{4}{n}\sum_{i=1}^{n}\|\boldsymbol{z}\|^2\cdot(\boldsymbol{x}_S^{(i)\top}\boldsymbol{z})\boldsymbol{x}_S^{(i)} - \big[4\mu(\boldsymbol{\beta}^{*\top}\boldsymbol{z})\boldsymbol{\beta}^* - 4\|\boldsymbol{z}\|^2\boldsymbol{z}\big]\bigg\| \\ &\quad\le C'\sqrt{s\log(ns)/n}. \end{aligned} \tag{A.70}$$

Combining (A.68), (A.69), and (A.70), we conclude that $\boldsymbol{t}_3'$ in (A.67) satisfies

$$\Big\|\boldsymbol{t}_3' - \big[4\mu(\boldsymbol{\beta}^{*\top}\boldsymbol{z})\boldsymbol{\beta}^* - 4\|\boldsymbol{z}\|^2\boldsymbol{z}\big]\Big\| \le C'\sqrt{s\log(ns)/n} \tag{A.71}$$

for some absolute constant $C'$. Then by (A.57), (A.65), (A.66), and (A.71), we have

$$\boldsymbol{h}^\top(\boldsymbol{t}_{22}+\boldsymbol{t}_3) \ge -4\rho\cdot(\boldsymbol{\beta}^{*\top}\boldsymbol{z})\cdot(\boldsymbol{\beta}^{*\top}\boldsymbol{h}) - 4\|\boldsymbol{z}\|^2(\boldsymbol{z}^\top\boldsymbol{h}) - C'\sqrt{s\log(ns)/n}. \tag{A.72}$$

Furthermore, for $\boldsymbol{t}_4$ in (A.50), by the first part of (A.37) in the `RIP`-condition, we have

$$|\boldsymbol{h}^\top\boldsymbol{t}_4| \le C'\sqrt{s\log n/n}. \tag{A.73}$$

Finally, plugging (A.58), (A.64), (A.72), and (A.73) into $\boldsymbol{h}^\top[\nabla\ell_n(\boldsymbol{z})]_S$, we obtain that

$$\begin{aligned}\boldsymbol{h}^\top[\nabla\ell_n(\boldsymbol{z})]_S \ge\ & 12\|\overline{\boldsymbol{\beta}}\|^2(\overline{\boldsymbol{\beta}}^\top\boldsymbol{h}) + 24(\overline{\boldsymbol{\beta}}^\top\boldsymbol{h})^2 + 11.88\|\overline{\boldsymbol{\beta}}\|^2\cdot\|\boldsymbol{h}\|^2 - 37\|\overline{\boldsymbol{\beta}}\|\cdot\|\boldsymbol{h}\|^3 \\ & - 4\rho\cdot(\boldsymbol{\beta}^{*\top}\boldsymbol{z})(\boldsymbol{\beta}^{*\top}\boldsymbol{h}) - 4\|\boldsymbol{z}\|^2(\boldsymbol{z}^\top\boldsymbol{h}) - C'\sqrt{s\log(ns)/n} \end{aligned} \tag{A.74}$$

for some absolute constant $C'$. Furthermore, since $\overline{\boldsymbol{\beta}} = \sqrt{\rho/2} \cdot \boldsymbol{\beta}^*$ and $\boldsymbol{h} = \boldsymbol{z} - \overline{\boldsymbol{\beta}}$, we have

$$\rho(\boldsymbol{\beta}^{*\top}\boldsymbol{z})(\boldsymbol{\beta}^{*\top}\boldsymbol{h}) = 2(\overline{\boldsymbol{\beta}}^\top\boldsymbol{z})(\overline{\boldsymbol{\beta}}^\top\boldsymbol{h}) = 2\|\overline{\boldsymbol{\beta}}\|^2(\overline{\boldsymbol{\beta}}^\top\boldsymbol{h}) + 2(\overline{\boldsymbol{\beta}}^\top\boldsymbol{h})^2. \tag{A.75}$$

Moreover, by rewriting $\|\boldsymbol{z}\|^2(\boldsymbol{z}^\top\boldsymbol{h})$ using $\overline{\boldsymbol{\beta}}$ and $\boldsymbol{h}$, we have

$$\begin{aligned}\|\boldsymbol{z}\|^2(\boldsymbol{z}^\top\boldsymbol{h}) &= (\|\boldsymbol{h}\|^2 + 2\overline{\boldsymbol{\beta}}^\top\boldsymbol{h} + \|\overline{\boldsymbol{\beta}}\|^2) \cdot (\|\boldsymbol{h}\|^2 + \overline{\boldsymbol{\beta}}^\top\boldsymbol{h}) \\ &= \|\boldsymbol{h}\|^4 + 3(\overline{\boldsymbol{\beta}}^\top\boldsymbol{h}) \cdot \|\boldsymbol{h}\|^2 + \|\overline{\boldsymbol{\beta}}\|^2 \cdot \|\boldsymbol{h}\|^2 + 2(\overline{\boldsymbol{\beta}}^\top\boldsymbol{h})^2 + \|\overline{\boldsymbol{\beta}}\|^2 \cdot (\overline{\boldsymbol{\beta}}^\top\boldsymbol{h}).\end{aligned} \tag{A.76}$$

Plugging (A.75) and (A.76) into (A.74), we obtain

$$\boldsymbol{h}^\top[\nabla\ell_n(\boldsymbol{z})]_S \geq -37\|\overline{\boldsymbol{\beta}}\| \cdot \|\boldsymbol{h}\|^3 - 4\|\boldsymbol{h}\|^4 - 12(\overline{\boldsymbol{\beta}}^\top\boldsymbol{h})\|\boldsymbol{h}\|^2 + 7.87\|\overline{\boldsymbol{\beta}}\|^2\|\boldsymbol{h}\|^2 + 8(\overline{\boldsymbol{\beta}}^\top\boldsymbol{h})^2.$$

Since $\|\boldsymbol{h}\| = \|\boldsymbol{z} - \overline{\boldsymbol{\beta}}\| \leq \alpha\|\overline{\boldsymbol{\beta}}\|$ with $\alpha \leq 1/4$, we have

$$\begin{aligned}\boldsymbol{h}^\top[\nabla\ell_n(\boldsymbol{z})]_S &\geq -37\alpha\|\overline{\boldsymbol{\beta}}\|^2 \cdot \|\boldsymbol{h}\|^2 - 4\alpha^2\|\overline{\boldsymbol{\beta}}\|^2 \cdot \|\boldsymbol{h}\|^2 - 12\alpha\|\overline{\boldsymbol{\beta}}\|^2 \cdot \|\boldsymbol{h}\|^2 + 7.87\|\overline{\boldsymbol{\beta}}\|^2 \cdot \|\boldsymbol{h}\|^2 \\ &\geq (7.87 - 50\alpha)\|\overline{\boldsymbol{\beta}}\|^2 \cdot \|\boldsymbol{h}\|^2,\end{aligned} \tag{A.77}$$

which concludes the second step.

**Step 3.** In this step, we upper bound $\|[\nabla\ell_n(\boldsymbol{z})]_S\|$. Combining this bound with the previous step, we obtain an upper bound for $R_1$.

First, by the triangle inequality, we have.

$$\|[\nabla\ell_n(\boldsymbol{z})]_S\| \leq \|\boldsymbol{t}_1\| + \|\boldsymbol{t}_2 + \boldsymbol{t}_3\| + \|\boldsymbol{t}_4\|.$$

By setting $\boldsymbol{v} = \boldsymbol{t}_1/\|\boldsymbol{t}_1\|$ in (A.53), we have $\|\boldsymbol{t}_1\| \leq C'\sqrt{s \log n/n}$. In addition, by the first part of (A.40) in the `Mean-Concentration` condition, we have $\|\boldsymbol{t}_4\| \leq C'\sqrt{\log n/n}$.

It remains to bound $\|\boldsymbol{t}_2 + \boldsymbol{t}_3\|$. We follow the same method used in bounding $\boldsymbol{h}^\top(\boldsymbol{t}_2 + \boldsymbol{t}_3)$ in the previous step. Recall that we define $\boldsymbol{t}_{21}$ and $\boldsymbol{t}_{22}$ in (A.55). For $\boldsymbol{t}_{21}$, combining (A.56) and (A.59), we have

$$\boldsymbol{t}_{21} = 12\|\overline{\boldsymbol{\beta}}\|^2 \cdot \overline{\boldsymbol{\beta}} + \frac{4}{n}\sum_{i=1}^n \Big[3(\boldsymbol{x}^{(i)\top}\overline{\boldsymbol{\beta}})^2(\boldsymbol{x}^{(i)\top}\boldsymbol{h}) + 3(\boldsymbol{x}^{(i)\top}\overline{\boldsymbol{\beta}})(\boldsymbol{x}^{(i)\top}\boldsymbol{h})^2 + (\boldsymbol{x}^{(i)\top}\boldsymbol{h})^3\Big]\boldsymbol{x}_S^{(i)} + \boldsymbol{\varepsilon}_1, \tag{A.78}$$

where $\boldsymbol{\varepsilon}_1$ is an error vector satisfying $\|\boldsymbol{\epsilon}_1\| \leq K_2\sqrt{s\log(ns)/n}$. Here the constant $K_2$ is specified in in the `Mean-Concentration` condition. In addition, similar to (A.65) and (A.66), for $\boldsymbol{t}_{22}$ we obtain

$$\boldsymbol{t}_{22} + \frac{4}{n}\sum_{i=1}^n [y^{(i)}(\widetilde{\boldsymbol{x}}_S^{(i)\top}\boldsymbol{z})\widetilde{\boldsymbol{x}}_S^{(i)}] = 4(\rho + \mu) \cdot (\boldsymbol{\beta}^{*\top}\boldsymbol{z}) \cdot \boldsymbol{\beta}^{*\top} + \boldsymbol{\varepsilon}_2, \tag{A.79}$$

where $\boldsymbol{\varepsilon}_2$ satisfies $\|\boldsymbol{\epsilon}_2\| \leq C'\sqrt{s\log(ns)/n}$ for some absolute constant $C'$. Moreover, for $\boldsymbol{t}_3$, recall that we define $\boldsymbol{t}_3'$ in (A.67). Similar to (A.68), (A.69), and (A.70), we obtain

$$\boldsymbol{t}_3' = \boldsymbol{t}_3 - \frac{4}{n}\sum_{i=1}^n [y^{(i)}(\widetilde{\boldsymbol{x}}_S^{(i)\top}\boldsymbol{z})\widetilde{\boldsymbol{x}}_S^{(i)}] = [4\mu(\boldsymbol{\beta}^{*\top}\boldsymbol{z})\boldsymbol{\beta}^* - 4\|\boldsymbol{z}\|^2\boldsymbol{z}] + \boldsymbol{\varepsilon}_3 \tag{A.80}$$

with $\|\varepsilon_3\|_2 \le C'\sqrt{s\log(ns)/n}$. Therefore, combining (A.78), (A.79), and (A.80), we have

$$\begin{aligned}\boldsymbol{t}_2+\boldsymbol{t}_3 &= 12\|\overline{\boldsymbol{\beta}}\|^2\cdot\overline{\boldsymbol{\beta}} - 4(\rho+\mu)\cdot(\boldsymbol{\beta}^{*\top}\boldsymbol{z})\cdot\boldsymbol{\beta}^* + 4\mu(\boldsymbol{\beta}^{*\top}\boldsymbol{z})\boldsymbol{\beta}^* - 4\|\boldsymbol{z}\|^2\boldsymbol{z}\\ &\quad + \frac{4}{n}\sum_{i=1}^n\Big[3\big(\boldsymbol{x}^{(i)\top}\overline{\boldsymbol{\beta}}\big)^2\big(\boldsymbol{x}^{(i)\top}\boldsymbol{h}\big) + 3\big(\boldsymbol{x}^{(i)\top}\overline{\boldsymbol{\beta}}\big)\big(\boldsymbol{x}^{(i)\top}\boldsymbol{h}\big)^2 + \big(\boldsymbol{x}^{(i)\top}\boldsymbol{h}\big)^3\Big]\boldsymbol{x}_S^{(i)} + \boldsymbol{\varepsilon},\end{aligned}$$

where the error vector $\boldsymbol{\varepsilon} = \boldsymbol{\varepsilon}_1+\boldsymbol{\varepsilon}_2+\boldsymbol{\varepsilon}_3$ satisfies $\|\boldsymbol{\epsilon}\| \le C's\sqrt{\log(np)/n}$ for some constant $C'$. To simplify this equation, recall that we let $\overline{\boldsymbol{\beta}} = \sqrt{\rho/2}\cdot\boldsymbol{\beta}^*$ and $\boldsymbol{z} = \overline{\boldsymbol{\beta}}+\boldsymbol{h}$. Then we have

$$\begin{aligned}&12\|\overline{\boldsymbol{\beta}}\|^2\cdot\overline{\boldsymbol{\beta}} - 4(\rho+\mu)\cdot(\boldsymbol{\beta}^{*\top}\boldsymbol{z})\cdot\boldsymbol{\beta}^* + 4\mu(\boldsymbol{\beta}^{*\top}\boldsymbol{z})\boldsymbol{\beta}^* - 4\|\boldsymbol{z}\|^2\boldsymbol{z}\\ &\quad = -4\|\boldsymbol{h}\|^2\overline{\boldsymbol{\beta}} - 4\|\boldsymbol{z}\|^2\boldsymbol{h} - 8(\overline{\boldsymbol{\beta}}^\top\boldsymbol{h})\overline{\boldsymbol{\beta}} = -4\|\boldsymbol{h}\|^2\overline{\boldsymbol{\beta}} - 4\|\overline{\boldsymbol{\beta}}\|^2\boldsymbol{h} - 4\|\boldsymbol{h}\|^2\boldsymbol{h} - 16(\overline{\boldsymbol{\beta}}^\top\boldsymbol{h})\overline{\boldsymbol{\beta}}. \qquad \text{(A.81)}\end{aligned}$$

Moreover, by the definition of $\mathcal{S}$, there exists a vector $\boldsymbol{w}\in\mathcal{S}$ such that $\|\boldsymbol{w}\| = 1$ and

$$\bigg\|\frac{4}{n}\sum_{i=1}^n\big[3\big(\boldsymbol{x}^{(i)\top}\overline{\boldsymbol{\beta}}\big)^2\big(\boldsymbol{x}^{(i)\top}\boldsymbol{h}\big)\big]\boldsymbol{x}_S^{(i)}\bigg\|^2 \le \frac{16}{n^2}\bigg|\sum_{i=1}^n\big(\boldsymbol{x}^{(i)\top}\overline{\boldsymbol{\beta}}\big)^2\big(\boldsymbol{x}^{(i)\top}\boldsymbol{h}\big)\big(\boldsymbol{x}^{(i)\top}\boldsymbol{w}\big)\bigg|^2.$$

By Hölder's inequality and the second part in (A.37) in the `RIP`-condition, we have

$$\begin{aligned}&\frac{16}{n^2}\bigg|\sum_{i=1}^n\big(\boldsymbol{x}^{(i)\top}\overline{\boldsymbol{\beta}}\big)^2\big(\boldsymbol{x}^{(i)\top}\boldsymbol{h}\big)\big(\boldsymbol{x}^{(i)\top}\boldsymbol{w}\big)\bigg|^2\\ &\le \frac{16}{n^2}\bigg(\bigg|\sum_{i=1}^n\big|\boldsymbol{x}^{(i)\top}\overline{\boldsymbol{\beta}}\big|^4\bigg|^{1/2}\cdot\bigg|\sum_{i=1}^n\big|\boldsymbol{x}^{(i)\top}\overline{\boldsymbol{\beta}}\big|^4\bigg|\cdot\bigg|\sum_{i=1}^n\big|\boldsymbol{x}^{(i)\top}\boldsymbol{h}\big|^4\bigg|\cdot\bigg|\sum_{i=1}^n\big|\boldsymbol{x}^{(i)\top}\boldsymbol{w}\big|^4\bigg|\bigg)^{1/2}\\ &\le \frac{16}{n^2}\big[(3n)^{1/4}+\sqrt{s}+\sqrt{3\log n}\big]^8\|\overline{\boldsymbol{\beta}}\|^4\|\boldsymbol{h}\|^2 \le 145\|\overline{\boldsymbol{\beta}}\|^4\|\boldsymbol{h}\|^2,\end{aligned}$$

where the last inequality holds when $n$ is sufficiently large. Here the first inequality follows from Hölder's inequality and the second inequality follows from the `RIP`-condition. By the same argument, we obtain

$$\bigg\|\frac{4}{n}\sum_{i=1}^n\big[3\big(\boldsymbol{x}^{(i)\top}\overline{\boldsymbol{\beta}}\big)\big(\boldsymbol{x}^{(i)\top}\boldsymbol{h}\big)^2\big]\boldsymbol{x}_S^{(i)}\bigg\|^2 \le 145\|\overline{\boldsymbol{\beta}}\|^2\|\boldsymbol{h}\|^4 \quad\text{and}\quad \bigg\|\frac{4}{n}\sum_{i=1}^n\big[\big(\boldsymbol{x}^{(i)\top}\boldsymbol{h}\big)\big]^3\boldsymbol{x}_S^{(i)}\bigg\|^2 \le 145\|\boldsymbol{h}\|^6.$$

By the triangle inequality and (A.81), we have

$$\|\boldsymbol{t}_2+\boldsymbol{t}_3\| \le 17\|\overline{\boldsymbol{\beta}}\|\cdot\|\boldsymbol{h}\|^2 + 33\|\overline{\boldsymbol{\beta}}\|^2\|\boldsymbol{h}\| + 17\|\boldsymbol{h}\|^3 + C'\sqrt{s\log(np)/n}.$$

Together with the upper bounds for $\|\boldsymbol{t}_1\|$ and $\|\boldsymbol{t}_4\|$, since $\|\boldsymbol{h}\|\le\alpha\|\overline{\boldsymbol{\beta}}\|$, we have

$$\|\nabla\ell_n(\boldsymbol{z})_S\|^2 \le 2\cdot(33+17\alpha+17\alpha^2)^2\|\overline{\boldsymbol{\beta}}\|^4\|\boldsymbol{h}\|^2 + C's\log(np)/n \qquad \text{(A.82)}$$

for some absolute constant $C'$.

Now, we are ready to bound $R_1$. Combining (A.77) and (A.82), we have

$$\begin{aligned}R_1^2 &\le \|\boldsymbol{h}\|^2 - 2\eta\cdot(7.87-50\alpha)\|\overline{\boldsymbol{\beta}}\|^2\cdot\|\boldsymbol{h}\|^2\\ &\quad + \eta^2\cdot\Big[2\cdot(33+17\alpha+17\alpha^2)^2\|\overline{\boldsymbol{\beta}}\|^4\|\boldsymbol{h}\|^2 + C's\log(np)/n\Big]. \qquad \text{(A.83)}\end{aligned}$$

Recall that $\|\overline{\boldsymbol{\beta}}\|^2 = \rho/2$ and $\alpha$ is an arbitrarily small constant. When $\alpha \leq 1/100$, by (A.83) we have

$$
\begin{aligned}
R_1^2 &\leq (1 - 3.5\eta\rho)\|\boldsymbol{h}\|^2 + (33 + 17\alpha + 17\alpha^2)^2\eta^2\rho^2 \cdot \|\boldsymbol{h}\|^2 + C's\log(np)/n \\
&\leq \Big[(1 - 7\eta\rho/4)\|\boldsymbol{h}\| + C'\sqrt{s\log(np)/n}\Big]^2 
\end{aligned} \tag{A.84}
$$

for some absolute constant $C'$ and a sufficiently small constant $\eta$. Thus we establish an upper bound for $R_1^2$, which concludes the third step.

**Step 4.** In this step, we bound $R_2$ defined in (A.49). Recall that the threshold value $\tau(\cdot)$ is defined in (2.7). To bound $R_2$, it suffices to bound $\tau(\boldsymbol{z})$.

We first rewrite $\tau(\boldsymbol{z})$ as

$$
\begin{aligned}
\tau(\boldsymbol{z}) &= \kappa\bigg[\frac{\log(np)}{n^2}\bigg]^{1/2}\bigg[\sum_{i=1}^n \big[y^{(i)} - (\boldsymbol{x}^{(i)\top}\boldsymbol{z})^2 - \mu_n + \|\boldsymbol{z}\|^2\big]^2\big(\boldsymbol{x}^{(i)\top}\boldsymbol{z}\big)^2\bigg]^{1/2} \\
&= \kappa\bigg[\frac{\log(np)}{n^2}\bigg]^{1/2}\bigg[\sum_{i=1}^n \big[y^{(i)} - (\boldsymbol{x}^{(i)\top}\boldsymbol{z})^2 - \mu + \Delta\mu + \|\boldsymbol{z}\|^2\big]^2\big(\boldsymbol{x}^{(i)\top}\boldsymbol{z}\big)^2\bigg]^{1/2},
\end{aligned}
$$

where $\Delta\mu = \mu - \mu_n$. By the inequality $(a+b)^2 \leq 2a^2 + 2b^2$ and the first part of (A.40) in the `Mean-Concentration` condition, we have

$$
\tau(\boldsymbol{z}) \leq 2\kappa\Big(\sqrt{\log(np)}/n\Big) \cdot \tau_1 + C'\sqrt{\log(np)/n} \tag{A.85}
$$

for some absolute constant $C'$, where $\tau_1$ is defined as

$$
\tau_1 = \bigg[\sum_{i=1}^n \big[y^{(i)} - (\boldsymbol{x}^{(i)\top}\boldsymbol{z})^2 - \mu + \|\boldsymbol{z}\|^2\big]^2\big(\boldsymbol{x}^{(i)\top}\boldsymbol{z}\big)^2\bigg]^{1/2}.
$$

By direct calculation, we write $\tau_1^2$ as

$$
\tau_1^2 = \sum_{i=1}^n \big[y^{(i)} - \mu + \|\boldsymbol{z}\|^2\big]^2\big(\boldsymbol{x}^{(i)\top}\boldsymbol{z}\big)^2 + \big(\boldsymbol{x}^{(i)\top}\boldsymbol{z}\big)^6 - 2\big[y^{(i)} - \mu + \|\boldsymbol{z}\|^2\big]\big(\boldsymbol{x}^{(i)\top}\boldsymbol{z}\big)^4 = \tau_{11} + \tau_{12},
$$

where $\tau_{11}$ and $\tau_{12}$ are given by

$$
\begin{aligned}
\tau_{11} &= \sum_{i=1}^n \big[\big(y^{(i)} - \mu\big)^2\big(\boldsymbol{x}^{(i)\top}\boldsymbol{z}\big)^2\big] \quad \text{and} \\
\tau_{12} &= \sum_{i=1}^n \Big\{\big[2y^{(i)}\|\boldsymbol{z}\|^2 - 2\mu\|\boldsymbol{z}\|^2 + \|\boldsymbol{z}\|^4\big]\big(\boldsymbol{x}^{(i)\top}\boldsymbol{z}\big)^2 + \big(\boldsymbol{x}^{(i)\top}\boldsymbol{z}\big)^6 - 2\big[y^{(i)} - \mu + \|\boldsymbol{z}\|^2\big]\big(\boldsymbol{x}^{(i)\top}\boldsymbol{z}\big)^4\Big\}.
\end{aligned}
$$

In the following, we bound $\tau_{11}$ and $\tau_{12}$ separately. For $\tau_{11}$, by Cauchy-Schwarz inequality, we obtain

$$
\tau_{11} \leq \bigg[\sum_{i=1}^n (y^{(i)} - \mu)^4\bigg]^{1/2}\bigg[\sum_{i=1}^n (\boldsymbol{x}^{(i)\top}\boldsymbol{z})^4\bigg]^{1/2}.
$$

Note that that $(y^{(i)}-\mu)^4$ can be bounded by a degree-4 polynomial of $|y^{(i)}|$. By the `Mean-Concentration` condition, we have $\sum_{i=1}^n (y^{(i)}-\mu)^4 \le C'n$ for some absolute constant $C'$. Moreover, by the second part of (A.37) in the `RIP`-condition, we have

$$\sum_{i=1}^n (\boldsymbol{x}^{(i)\top}\boldsymbol{z})^4 \le \big[(3n)^{1/4} + \sqrt{s} + \sqrt{3\log n}\big]^4 \cdot \|\boldsymbol{z}\|^4$$

for sufficiently large $n$. Note that $\|\boldsymbol{z}\| \le (1+\alpha)\|\overline{\boldsymbol{\beta}}\|$ is bounded by some constant. Thus for $\tau_{11}$, we conclude that

$$\tau_{11} \le C'\sqrt{n} \cdot [(3n)^{1/4} + \sqrt{s} + \sqrt{3\log n}]^2 \le C'n,$$

for some absolute constants $C'$.

In addition, for $\tau_{12}$, we plug the identity $\boldsymbol{z} = \overline{\boldsymbol{\beta}} + \boldsymbol{h}$ into $\tau_{12}$ and write $\tau_{12}$ as $\tau_{\overline{\boldsymbol{\beta}}} + \tau_y + \tau_{12}'$, where

$$\tau_{\overline{\boldsymbol{\beta}}} = \sum_{i=1}^n \Big[(-2\mu\|\overline{\boldsymbol{\beta}}\|^2 + \|\overline{\boldsymbol{\beta}}\|^4)\cdot(\boldsymbol{x}^{(i)\top}\overline{\boldsymbol{\beta}})^2 + \big(\boldsymbol{x}^{(i)\top}\overline{\boldsymbol{\beta}}\big)^6 - 2(-\mu + \|\overline{\boldsymbol{\beta}}\|^2)\big(\boldsymbol{x}^{(i)\top}\overline{\boldsymbol{\beta}}\big)^4\Big],$$
$$\tau_y = \sum_{i=1}^n \Big[(2y^{(i)}\|\boldsymbol{z}\|^2)\big(\boldsymbol{x}^{(i)\top}\boldsymbol{z}\big)^2 - 2y^{(i)}\big(\boldsymbol{x}^{(i)\top}\boldsymbol{z}\big)^4\Big].$$

By direct calculation, it can be seen that $\tau_{12}'$ is a sum of terms of the form

$$\sum_{t=1}^n a \cdot (\boldsymbol{x}^{(i)\top}\overline{\boldsymbol{\beta}})^b \cdot (\boldsymbol{x}^{(i)\top}\boldsymbol{h})^c,$$

where $a$ is a constant, and $b, c \ge 1$ satisfy $b + c \in \{2, 4, 6\}$.

We then bound these three terms of $\tau_{12}$ separately. First, note that $\tau_{\overline{\boldsymbol{\beta}}}$ can be written as the sum of polynomials of $\boldsymbol{x}^{(i)\top}\overline{\boldsymbol{\beta}}$. The first part of (A.42) in the `Mean-Concentration` condition directly implies that $\tau_{\overline{\boldsymbol{\beta}}} \le C'n$ for some absolute constant $C'$.

Next, for the first part of $\tau_y$, by Cauchy-Schwarz inequality and the second part of (A.37) in the `RIP`-condition, we have

$$\Big|\sum_{i=1}^n (2y^{(i)}\|\boldsymbol{z}\|^2)\big(\boldsymbol{x}^{(i)\top}\boldsymbol{z}\big)^2\Big| \le \|\boldsymbol{z}\|^2 \cdot \Big[\sum_{i=1}^n (y^{(i)})^2\Big]^{1/2} \cdot \Big[\sum_{t=1}^n \big(\boldsymbol{x}^{(i)\top}\boldsymbol{z}\big)^4\Big]^{1/2} \le C'n, \tag{A.86}$$

where $C' > 0$ is an absolute constant. For the second part of $\tau_y$, we expand $\big(\boldsymbol{x}^{(i)\top}\boldsymbol{z}\big)^4$ and use triangle inequality to obtain

$$\Big|\sum_{i=1}^n (2y^{(i)}\|\boldsymbol{z}\|^2)\big(\boldsymbol{x}^{(i)\top}\boldsymbol{z}\big)^4\Big| \le 2\|\boldsymbol{z}\|^2 \sum_{a+b=4} \binom{4}{a} \cdot \Big|\sum_{i=1}^n y^{(i)}\big(\boldsymbol{x}^{(i)\top}\overline{\boldsymbol{\beta}}\big)^a\big(\boldsymbol{x}^{(i)\top}\boldsymbol{h}\big)^b\Big|,$$

where $a$ and $b$ are nonnegative integers. For $b = 0$, using the same argument as in (A.86), we have

$$\Big|2\|\boldsymbol{z}\|^2\sum_{i=1}^n y^{(i)}\big(\boldsymbol{x}^{(i)\top}\overline{\boldsymbol{\beta}}\big)^4\Big| \le C'n.$$

Moreover, for $b=1$, by Hölder's inequality we have

$$\left|\sum_{i=1}^{n} y^{(i)}\big(\boldsymbol{x}^{(i)\top}\overline{\boldsymbol{\beta}}\big)^3\big(\boldsymbol{x}^{(i)\top}\boldsymbol{h}\big)\right| \le \left|\sum_{i=1}^{n} |y^{(i)}|^{3/4}\big(\boldsymbol{x}^{(i)\top}\overline{\boldsymbol{\beta}}\big)^4\right|^{4/3}\left|\sum_{i=1}^{n}\big(\boldsymbol{x}^{(i)\top}\boldsymbol{h}\big)^4\right|^{1/4}. \tag{A.87}$$

By the first part of (A.42) in the `Mean-Concentration` condition and the second part of (A.37) in the `RIP`-condition, the right-hand side of (A.87) can be further bounded by $C'n$ when $n$ is sufficiently large. As for $b=2$, by Cauchy-Schwarz inequality we have

$$\left|\sum_{i=1}^{n} y^{(i)}\big(\boldsymbol{x}^{(i)\top}\overline{\boldsymbol{\beta}}\big)^2\big(\boldsymbol{x}^{(i)\top}\boldsymbol{h}\big)^2\right| \le \left|\sum_{i=1}^{n} (y^{(i)})^2\big(\boldsymbol{x}^{(i)\top}\overline{\boldsymbol{\beta}}\big)^4\right|^{1/2}\left|\sum_{i=1}^{n}\big(\boldsymbol{x}^{(i)\top}\boldsymbol{h}\big)^4\right|^{1/2} \le C'n.$$

Here the last inequality follows from the first part of (A.42) in the `Mean-Concentration` condition and the second part of (A.37) in the `RIP`-condition. Similarly, for $b=3$, by Hölder's inequality we have

$$\left|\sum_{i=1}^{n} y^{(i)}\big(\boldsymbol{x}^{(i)\top}\overline{\boldsymbol{\beta}}\big)\big(\boldsymbol{x}^{(i)\top}\boldsymbol{h}\big)^3\right| \le \left|\sum_{i=1}^{n} (y^{(i)})^4\big(\boldsymbol{x}^{(i)\top}\overline{\boldsymbol{\beta}}\big)^4\right|^{1/4}\left|\sum_{i=1}^{n}\big(\boldsymbol{x}^{(i)\top}\boldsymbol{h}\big)^4\right|^{3/4} \le C'n.$$

Finally, for $b=4$, by Hölder's inequality we obtain

$$\begin{aligned}\left|\sum_{i=1}^{n} y^{(i)}\big(\boldsymbol{x}^{(i)\top}\boldsymbol{h}\big)^4\right| &\le \left|\sum_{i=1}^{n} (y^{(i)})^3\right|^{2/6}\left|\sum_{i=1}^{n}\big(\boldsymbol{x}^{(i)\top}\boldsymbol{h}\big)^6\right|^{4/6} \le C'n^{1/3}\cdot\big[(15n)^{1/6}+\sqrt{s}+\sqrt{3\log n}\big]^4\|\overline{\boldsymbol{\beta}}\|\cdot\|\boldsymbol{h}\|^4\\ &\le C'\big[(15n)^{1/6}+\sqrt{s}+\sqrt{3\log n}\big]^6\cdot\|\overline{\boldsymbol{\beta}}\|\cdot\|\boldsymbol{h}\|^4 \le C'(n+s^3)\cdot\|\overline{\boldsymbol{\beta}}\|\cdot\|\boldsymbol{h}\|^4.\end{aligned}$$

Here $C'$ is an absolute constant, the third inequality follows from the fact that

$$n^{1/3} \le \big[(15n)^{1/6}+\sqrt{s}+\sqrt{3\log n}\big]^2,$$

and the last inequality holds when $n$ is sufficiently large such that $n \gg \log^3 n$.

Thus, for $\tau_y$, we conclude that

$$\tau_y \le C'n + C'(n+s^3)\cdot\|\overline{\boldsymbol{\beta}}\|\cdot\|\boldsymbol{h}\|^4.$$

Then we derive an upper bound for $\tau_{12}'$. It suffices to bound terms of the form $(\boldsymbol{x}^{(i)\top}\overline{\boldsymbol{\beta}})^b(\boldsymbol{x}^{(i)\top}\boldsymbol{h})^c$ with $b+c\in\{2,4,6\}$. We first consider the case where $b+c=2$. By Hölder's inequality and the first part of (A.37) in the `RIP`-condition, we have

$$\left|\sum_{i=1}^{n}(\boldsymbol{x}^{(i)\top}\overline{\boldsymbol{\beta}})^b(\boldsymbol{x}^{(i)\top}\boldsymbol{h})^c\right| \le \left[\sum_{i=1}^{n}(\boldsymbol{x}^{(i)\top}\overline{\boldsymbol{\beta}})^2\right]^{b/2}\cdot\left[\sum_{i=1}^{n}(\boldsymbol{x}^{(i)\top}\boldsymbol{h})^2\right]^{c/2} \le C'n.$$

In addition, when $b+c=4$, by Hölder's inequality and the second part of (A.37) in the `RIP`-condition, we obtain

$$\left|\sum_{i=1}^{n}(\boldsymbol{x}^{(i)\top}\overline{\boldsymbol{\beta}})^b(\boldsymbol{x}^{(i)\top}\boldsymbol{h})^c\right| \le \left[\sum_{i=1}^{n}(\boldsymbol{x}^{(i)\top}\overline{\boldsymbol{\beta}})^4\right]^{b/4}\cdot\left[\sum_{i=1}^{n}(\boldsymbol{x}^{(i)\top}\boldsymbol{h})^4\right]^{c/4} \le C''(n+s^2+\log^2 n) \le C'n.$$

Lastly, we consider the case where $b + c = 6$. We treat the cases where $1 \le c \le 3$ and $c \ge 4$ separately. When $c \le 3$, we apply Cauchy-Schwarz inequality to obtain

$$\left|\sum_{i=1}^{n}(\boldsymbol{x}^{(i)\top}\overline{\boldsymbol{\beta}})^b(\boldsymbol{x}^{(i)\top}\boldsymbol{h})^c\right| \le \left[\sum_{i=1}^{n}(\boldsymbol{x}^{(i)\top}\overline{\boldsymbol{\beta}})^{2b}\right]^{1/2} \cdot \left[\sum_{i=1}^{n}(\boldsymbol{x}^{(i)\top}\boldsymbol{h})^{2c}\right]^{1/2}. \tag{A.88}$$

Note that $(\boldsymbol{X}^\top\overline{\boldsymbol{\beta}})^2$ is a sub-exponential random variables with bounded $\psi_1$ norm. Similar to the proof the first part of (A.42) in the `Mean-Concentration` condition, we obtain that $n^{-1}\sum_{i=1}^n(\boldsymbol{x}^{(i)\top}\overline{\boldsymbol{\beta}})^{2b}$ is bounded by some absolute constant. Moreover, for $c = 1$ and 2, by (A.37) in the `RIP` condition, we have that $n^{-1}\sum_{i=1}^n(\boldsymbol{x}^{(i)\top}\boldsymbol{h})^{2c}$ is bounded by some absolute constant. Whereas for $c = 3$, by the first part in (A.38) in the `RIP`-condition, we have

$$\left[\sum_{i=1}^{n}(\boldsymbol{x}^{(i)\top}\boldsymbol{h})^{2c}\right]^{1/2} \le \left[(15n)^{1/6} + \sqrt{s} + \sqrt{3\log n}\right]^3\|\boldsymbol{h}\|^3 \le 4(\sqrt{n} + s^{3/2})\|\boldsymbol{h}\|^3. \tag{A.89}$$

Thus, plugging (A.89) into the right-hand side of (A.88), we have

$$\left|\sum_{i=1}^{n}(\boldsymbol{x}^{(i)\top}\overline{\boldsymbol{\beta}})^b(\boldsymbol{x}^{(i)\top}\boldsymbol{h})^c\right| \le \left[\sum_{i=1}^{n}(\boldsymbol{x}^{(i)\top}\overline{\boldsymbol{\beta}})^{2b}\right]^{1/2} \cdot \left[\sum_{i=1}^{n}(\boldsymbol{x}^{(i)\top}\boldsymbol{h})^{2c}\right]^{1/2} \le C'\left[n + (n+s^3)\|\boldsymbol{h}\|^3\right],$$

where $C'$ is a constant. Next, we consider the case where $c \ge 4$. By Hölder's inequality and the first part in (A.38) in the `RIP`-condition, we obtain

$$\begin{aligned}\left|\sum_{i=1}^{n}(\boldsymbol{x}^{(i)\top}\overline{\boldsymbol{\beta}})^b(\boldsymbol{x}^{(i)\top}\boldsymbol{h})^c\right| &\le \left[\sum_{i=1}^{n}(\boldsymbol{x}^{(i)\top}\overline{\boldsymbol{\beta}})^{6}\right]^{b/6} \cdot \left[\sum_{i=1}^{n}(\boldsymbol{x}^{(i)\top}\boldsymbol{h})^{6}\right]^{c/6}\\ &\le \left[(15n)^{1/6} + \sqrt{s} + \sqrt{3\log n}\right]^6\|\overline{\boldsymbol{\beta}}\|^b\|\boldsymbol{h}\|^c \le 16(n+s^3)\cdot\|\overline{\boldsymbol{\beta}}\|^b\|\boldsymbol{h}\|^c,\end{aligned}$$

where the last inequality holds when $n$ is sufficiently large. Plugging the bounds on $\tau_{11}$ and $\tau_{12}$ into (A.85) yields

$$\begin{aligned}\tau(\boldsymbol{z})^2 &\le 4\kappa^2\log(np)/(n^2)\cdot\tau_1^2 + C'\log(np)/n\\ &\le C'\log(np)/(n^2)\cdot\left[n + (n+s^3)\cdot(\|\overline{\boldsymbol{\beta}}\|^2\|\boldsymbol{h}\|^4 + \|\overline{\boldsymbol{\beta}}\|\|\boldsymbol{h}\|^5 + \|\boldsymbol{h}\|^6)\right] + C'\log(np)/n.\end{aligned} \tag{A.90}$$

Here the second inequality is obtained by plugging in the upper bounds for $\tau_{11}$ and $\tau_{12}$ and $C'$ is an absolute constant. By $\|\boldsymbol{h}\| = \|\boldsymbol{z} - \overline{\boldsymbol{\beta}}\| \le \alpha\|\overline{\boldsymbol{\beta}}\|$, (A.90) can be further bounded by

$$\begin{aligned}\tau(\boldsymbol{z})^2 &\le C'(n+s^3)\cdot\log(np)/(n^2)\cdot(1+\alpha+\alpha^2)\cdot\|\overline{\boldsymbol{\beta}}\|^2\|\boldsymbol{h}\|^4 + C'\log(np)/n\\ &\le C_\tau^2(n+s^3)\cdot\log(np)/(n^2)\cdot\|\overline{\boldsymbol{\beta}}\|^2\|\boldsymbol{h}\|^4 + C'\log(np)/n,\end{aligned} \tag{A.91}$$

where $C_\tau$ is an absolute constant.

We now have $R_2 = \eta\tau(\boldsymbol{z})\cdot\|\boldsymbol{v}\|$ for some $\boldsymbol{v}$ satisfying $\mathrm{supp}(\boldsymbol{v}) \subseteq \mathrm{supp}(\boldsymbol{\beta}^*)$ and $\|\boldsymbol{v}\|_\infty \le 1$. By the norm inequality and the identity $\|\overline{\boldsymbol{\beta}}\|^2 = \rho/2$, we have $\|\boldsymbol{v}\| \le \sqrt{|\mathrm{supp}(\boldsymbol{v})|}\cdot\|\boldsymbol{v}\|_\infty \le \sqrt{s}$. Thus by (A.91), we obtain

$$\begin{aligned}R_2 &\le C_\tau\eta\sqrt{(ns+s^4)\cdot\log(np)/(n^2)}\|\overline{\boldsymbol{\beta}}\|\|\boldsymbol{h}\|^2 + C'\sqrt{s\log(np)/n}\\ &\le 3C_\tau\alpha\eta\rho\|\boldsymbol{h}\| + C'\sqrt{s\log(np)/n}\end{aligned}$$

when $n \geq C[s^2 \log(np) + s \log^5 n]$ for some constant $C$. Here in the last inequality we use the fact that $\|\boldsymbol{h}\| \leq \alpha \|\overline{\boldsymbol{\beta}}\|$. Note that the constant $\alpha$ can be set arbitrarily small. Setting $\alpha$ such that $C_\tau \alpha \leq 1/4$, we obtain

$$R_2 \leq 3/4 \cdot \eta\rho \|\boldsymbol{h}\| + C' \sqrt{s \log(np)/n}. \tag{A.92}$$

Finally, we combine (A.84) and (A.92) to obtain

$$\begin{aligned} \|t(\boldsymbol{z}) - \overline{\boldsymbol{\beta}}\| &\leq R_1 + R_2 \leq (1 - 7/4 \cdot \eta\rho + 3/4 \cdot \eta\rho) \|\boldsymbol{z} - \overline{\boldsymbol{\beta}}\| + C' \sqrt{s \log(np)/n} \\ &= (1 - \eta\rho) \|\boldsymbol{z} - \overline{\boldsymbol{\beta}}\| + C' \sqrt{s \log(np)/n}, \end{aligned}$$

which concludes the proof. □

# B Proofs of Auxiliary Results

In this section, we prove the auxiliary results presented in the paper. In specific, we prove Proposition 2.1, which characterizes the minimizer of $\mathrm{Var}[Y - (\boldsymbol{X}^\top \boldsymbol{\beta})^2]$, and Lemma A.1, which considers spectral perturbation for spiked matrices.

## B.1 Proof of Proposition 2.1

*Proof.* We first decompose $\boldsymbol{\beta}$ into $\boldsymbol{\beta} = \zeta \boldsymbol{\beta}^* + \boldsymbol{\beta}^\perp$, where $\zeta = \boldsymbol{\beta}^\top \boldsymbol{\beta}^*$ and $\boldsymbol{\beta}^\perp$ is perpendicular to $\boldsymbol{\beta}^*$. Note that the Pythagorean theorem implies that $\|\boldsymbol{\beta}^\perp\|_2^2 = \|\boldsymbol{\beta}\|_2^2 - \zeta^2$. By (1.3) we have

$$\boldsymbol{X}^\top \boldsymbol{\beta}^* \perp\!\!\!\perp \boldsymbol{X}^\top \boldsymbol{\beta}^\perp \quad \text{and} \quad Y \perp\!\!\!\perp \boldsymbol{X}^\top \boldsymbol{\beta}^\perp.$$

In the following, we denote $\mu := \mathbb{E} Y$ for brevity. Since $\|\boldsymbol{\beta}^*\| = 1$, it holds that

$$\rho = \mathrm{Cov}[Y, (\boldsymbol{X}^\top \boldsymbol{\beta}^*)^2] = \mathbb{E}[(Y - \mu) \cdot (\boldsymbol{X}^\top \boldsymbol{\beta}^*)^2] = \mathbb{E}[Y (\boldsymbol{X}^\top \boldsymbol{\beta}^*)^2] - \mu. \tag{B.1}$$

By expanding the variance we obtain

$$\begin{aligned} \mathrm{Var}[Y - (\boldsymbol{X}^\top \boldsymbol{\beta})^2] &= \mathbb{E}\{[Y - (\boldsymbol{X}^\top \boldsymbol{\beta})^2]^2\} - \{\mathbb{E}[Y - (\boldsymbol{X}^\top \boldsymbol{\beta})^2]\}^2 \\ &= \mathbb{E}[Y^2 - 2Y (\boldsymbol{X}^\top \boldsymbol{\beta})^2 + (\boldsymbol{X}^\top \boldsymbol{\beta})^4 - \mu^2 + 2\mu \|\boldsymbol{\beta}\|^2 - \|\boldsymbol{\beta}\|^4] \\ &= \mathbb{E}(Y^2) - 2\mathbb{E}[Y (\boldsymbol{X}^\top \boldsymbol{\beta})^2] + 3\|\boldsymbol{\beta}\|^4 - \mu^2 + 2\mu \|\boldsymbol{\beta}\|^2 - \|\boldsymbol{\beta}\|^4, \end{aligned} \tag{B.2}$$

where we use $\mathbb{E}[(\boldsymbol{X}^\top \boldsymbol{\beta})^2] = \|\boldsymbol{\beta}\|_2^2$ and $\mathbb{E}[(\boldsymbol{X}^\top \boldsymbol{\beta})^4] = 3\|\boldsymbol{\beta}\|_2^4$. Decomposing $\boldsymbol{\beta}$ in (B.2), we have

$$\begin{aligned} \mathrm{Var}[Y - (\boldsymbol{X}^\top \boldsymbol{\beta})^2] &= \mathbb{E}(Y^2) - 2\mathbb{E}[Y (\zeta \boldsymbol{X}^\top \boldsymbol{\beta}^* + \boldsymbol{X}^\top \boldsymbol{\beta}^\perp)^2] + 2\|\boldsymbol{\beta}\|^4 - \mu^2 + 2\mu \|\boldsymbol{\beta}\|^2 \\ &= \mathbb{E}(Y^2) - 2\zeta^2 \cdot \mathbb{E}[Y (\boldsymbol{X}^\top \boldsymbol{\beta}^*)^2] - 2\mathbb{E}[Y (\boldsymbol{X}^\top \boldsymbol{\beta}^\perp)^2] + 2\|\boldsymbol{\beta}\|^4 - \mu^2 + 2\mu \|\boldsymbol{\beta}\|^2. \end{aligned} \tag{B.3}$$

By independence, we have $\mathbb{E}[Y (\boldsymbol{X}^\top \boldsymbol{\beta}^\perp)^2] = \mu \|\boldsymbol{\beta}^\perp\|^2 = \mu(\|\boldsymbol{\beta}\|_2^2 - \zeta^2)$. Combining (B.1) and (B.3), we have

$$\begin{aligned} \mathrm{Var}[Y - (\boldsymbol{X}^\top \boldsymbol{\beta})^2] &= \mathbb{E}(Y^2) - 2\zeta^2 \cdot \mathbb{E}[Y (\boldsymbol{X}^\top \boldsymbol{\beta}^*)^2] + 2\zeta^2 \mu - 2\mu \|\boldsymbol{\beta}\|^2 + 2\|\boldsymbol{\beta}\|^4 - \mu^2 + 2\mu \|\boldsymbol{\beta}\|^2 \\ &= \mathbb{E}(Y^2) - 2\zeta^2 \rho + 2\|\boldsymbol{\beta}\|^4 - \mu^2 = \mathrm{Var}(Y) - 2\zeta^2 \rho + 2\|\boldsymbol{\beta}\|^4. \end{aligned} \tag{B.4}$$

Since $\rho > 0$, $\boldsymbol{\beta} = \mathbf{0}$ cannot be a minimizer. To see this, if one sets $\widetilde{\boldsymbol{\beta}} = c\boldsymbol{\beta}^*$ for a sufficiently small but non-zero constant $c < \sqrt{\rho}$, one obtains $\zeta = c$ and $\|\boldsymbol{\beta}\| = c$. Then it holds that $\zeta^2\rho - \|\widetilde{\boldsymbol{\beta}}\|^4 = c^2(\rho - c^2) > 0$. By (B.4) we see that

$$\mathrm{Var}\big[Y - (\boldsymbol{X}^\top\widetilde{\boldsymbol{\beta}})^2\big] < \mathrm{Var}\big[Y - (\boldsymbol{X}^\top\mathbf{0})^2\big].$$

If we fix $\|\boldsymbol{\beta}\| > 0$ as a constant, then $|\zeta| \le \|\boldsymbol{\beta}\|$. By (B.4), $\mathrm{Var}[Y - (\boldsymbol{X}^\top\boldsymbol{\beta})^2]$ can be viewed as a function of $\zeta$. The minimum is achieved only when $\zeta = \pm\|\boldsymbol{\beta}\|$, in which case $\boldsymbol{\beta}^\perp = \mathbf{0}$. Setting $|\zeta| = \|\boldsymbol{\beta}\|$ in (B.4), we have

$$\mathrm{Var}\big[Y - (\boldsymbol{X}^\top\boldsymbol{\beta})^2\big] = \mathrm{Var}(Y) - 2\|\boldsymbol{\beta}\|^2\rho + 2\|\boldsymbol{\beta}\|^4,$$

which, as a function of $\|\boldsymbol{\beta}\|$, is minimized by $\|\boldsymbol{\beta}\| = \sqrt{\rho/2}$.

Putting everything together, we show that $\mathrm{Var}[Y - (\boldsymbol{X}^\top\boldsymbol{\beta})^2]$ is minimized by $\pm\sqrt{\rho/2}\cdot\boldsymbol{\beta}^*$, which concludes the proof. □

### B.2 Proof of Lemma A.1

*Proof.* Without loss of generality, we assume $\lambda > 0$ since otherwise the proof follows similarly by considering $-\mathbf{A}$. Let $\lambda_1$ be an eigenvalue of $\mathbf{A}$ with the largest magnitude. We first show that $\lambda_1 > 0$. Let $\mathbf{x} \in \mathbb{R}^d$ be any unit vector, then we have

$$\mathbf{x}^\top\mathbf{A}\mathbf{x} = \lambda(\mathbf{x}^\top\mathbf{v})^2 + \mathbf{x}^\top\mathbf{N}\mathbf{x} \ge \mathbf{x}^\top\mathbf{N}\mathbf{x} \ge -\|\mathbf{N}\|_2 \ge -\varphi > -\lambda/2.$$

On the other hand, it holds that

$$\mathbf{v}^\top\mathbf{A}\mathbf{v} = \lambda + \mathbf{v}^\top\mathbf{N}\mathbf{v} \ge \lambda - \|\mathbf{N}\|_2 > \lambda/2.$$

Therefore, we obtain that $\lambda_{\min}(\mathbf{A}) \ge -\lambda/2$ and that $\lambda_{\max}(\mathbf{A}) \ge \lambda/2$. This implies that $|\lambda_{\max}(\mathbf{A})| \ge |\lambda_{\min}(\mathbf{A})|$, which further implies that $\lambda_1 = \lambda_{\max}(\mathbf{A}) > 0$ by the definition of $\lambda_1$.

Note that $\widehat{\mathbf{v}}$ is the eigenvector of $\mathbf{A}$ corresponding to eigenvalue $\lambda_1$. Since $\lambda_1 = \lambda_{\max}(\mathbf{A})$ we have

$$\widehat{\mathbf{v}}^\top\mathbf{A}\widehat{\mathbf{v}} \ge \mathbf{v}^\top\mathbf{A}\mathbf{v} \ge \lambda - \|\mathbf{N}\|_2 \ge \lambda - \varphi. \tag{B.5}$$

Moreover, by the definition of $\mathbf{A}$, we have

$$\widehat{\mathbf{v}}^\top\mathbf{A}\widehat{\mathbf{v}} = \lambda(\mathbf{v}^\top\widehat{\mathbf{v}})^2 + \widehat{\mathbf{v}}^\top\boldsymbol{N}\widehat{\mathbf{v}} \le \lambda(\mathbf{v}^\top\widehat{\mathbf{v}})^2 + \varphi. \tag{B.6}$$

Combining (B.5) and (B.6), we we obtain $|\mathbf{v}^\top\widehat{\mathbf{v}}|^2 \ge 1 - (2\varphi)/\lambda$, which further implies that $|\mathbf{v}^\top\widehat{\mathbf{v}}| \ge 1 - \varphi/\lambda$. Thus we conclude the proof. □

## C Proof of Concentration Results

In this section, we prove Lemmas A.2 and A.3, which shows that the `RIP`-condition and the `Mean-Concentration` condition hold with high probability. Before presenting the proof, we first introduce a few supporting concentration results.

## C.1 Auxiliary Concentration Lemmas

We first present three concentration results for Gaussian random matrices with i.i.d. entries. These results are obtained from Cai et al. (2016), where the proofs can be found.

**Lemma C.1** (Cai et al. (2016), Lemma A.5)**.** *Let $\boldsymbol{A} \in \mathbb{R}^{m\times s}$ be a random matrix with i.i.d. entries from the standard Gaussian distribution. For any $t > 0$, with probability at least $1 - 4\exp(-t^2/2)$, we have*

$$\|\boldsymbol{A}\|_{2\to 6} \le (15m)^{1/6} + \sqrt{s} + t, \qquad \|\boldsymbol{A}\|_{2\to 4} \le (3m)^{1/4} + \sqrt{s} + t.$$

**Lemma C.2** (Cai et al. (2016), Lemma A.6)**.** *Let $\boldsymbol{x}_1, \boldsymbol{x}_2, \ldots, \boldsymbol{x}_n$ be independent normal $\mathcal{N}(0, \mathbf{I}_p)$ vectors. Let $S \subseteq [p]$ be a index set with size $s$, and $\delta \in (0,1)$ be any fixed constant. Moreover, let $\boldsymbol{\beta}$ be a fixed vector with $\mathrm{supp}(\boldsymbol{\beta}) \subseteq S$. With probability at least $1 - 1/(100n)$, for all $\boldsymbol{h} \in \mathbb{R}^p$ satisfying $\mathrm{supp}(\boldsymbol{h}) \subset S$, we have*

$$\frac{1}{n}\sum_{j=1}^{n}(\boldsymbol{x}^{(j)\top}\boldsymbol{\beta})^2(\boldsymbol{x}^{(j)\top}\boldsymbol{h})^2 \ge 2(\boldsymbol{\beta}^\top\boldsymbol{h})^2 + (1-\delta)\|\boldsymbol{\beta}\|^2\|\boldsymbol{h}\|^2.$$

*Here we require $n \ge C_\delta \cdot s\log s$, where $C_\delta$ is a constant depends only on $\delta$.*

**Lemma C.3** (Cai et al. (2016), Lemma A.8)**.** *Suppose $\boldsymbol{x}_1, \boldsymbol{x}_2, \ldots, \boldsymbol{x}_n$ are $n$ i.i.d. standard Gaussian random vectors in $\mathbb{R}^m$. Let $a_1, a_2, \ldots, a_n \in \mathbb{R}$ be $n$ fixed numbers. Then for any $t \ge 1$, with probability at least $1 - \exp[-4(m+t)]$, we have*

$$\Big\|\sum_{j=1}^{n} a_j\boldsymbol{x}_j\boldsymbol{x}_j^\top - \sum_{j=1}^{n} a_j\mathbf{I}_m\Big\|_2 \le C_0\bigg\{\bigg[(m+t)\cdot\sum_{j=1}^{n}a_j^2\bigg]^{1/2} + (m+t)\cdot\max_{j\in[n]}|a_j|\bigg\},$$

*where $C_0$ is an absolute constant.*

Besides concentration results for Gaussian random vectors, since the response $Y$ in the misspecified phase retrieval model is sub-exponential, we also need to consider concentration results involving sub-exponential random variables. For convenience of the reader, we first briefly recall the present a result on the concentration of polynomial functions of sub-Gaussian random vectors in Adamczak and Wolff (2015), which is applied in Lemma C.4 below. We first define the $\psi_2$-norm for a random variable $X$ by

$$\|X\|_{\psi_2} = \sup_{p\ge 1}\{p^{-1/2}\cdot(\mathbb{E}|X|^p)^{1/p}\}. \tag{C.1}$$

$X$ is sub-Gaussian if its $\psi_2$-norm is bounded.

Moreover, in the following, we introduce a norm for tensors, which will be used in the concentration results. Let $\ell \in \mathbb{N}_+$ be a positive integer. We denote by $\mathcal{P}_\ell$ the set of its partitions of $[\ell]$ into non-empty and non-intersecting disjoint sets. Moreover, let $\mathbf{A} = (a_{\mathbf{i}})_{\mathbf{i}\in[n]^\ell}$ be a tensor of order-$\ell$, whose entries are of the form

$$a_{\mathbf{i}} = a_{i_1,i_2,\ldots,i_\ell}, \quad \text{where} \quad \mathbf{i} = (i_1, i_2, \ldots, i_\ell).$$

Finally, let $\mathcal{J} = \{J_1, \dots, J_k\} \in \mathcal{P}_\ell$ be a fixed partition of $[\ell]$, where $J_j \subseteq [\ell]$ for each $j \in [k]$. Let $|\mathcal{J}|$ denote the cardinality of the $\mathcal{J}$, which is equal to $k$. We define a norm $\|\cdot\|_{\mathcal{J}}$ by

$$\|\mathbf{A}\|_{\mathcal{J}} = \sup\left\{ \sum_{\mathbf{i}\in[n]^\ell} a_{\mathbf{i}} \prod_{j=1}^{k} \boldsymbol{x}^{(j)}_{\mathbf{i}_{J_j}} : \|\boldsymbol{x}^{(j)}\| \le 1, \boldsymbol{x}^{(j)} \in \mathbb{R}^{n^{|J_j|}}, 1 \le j \le k \right\}, \tag{C.2}$$

where we write $\mathbf{i}_I = (i_k)_{k\in I}$ for any $I \subseteq [\ell]$ and the supremum is taken over all possible $k$ vectors $\{\boldsymbol{x}^{(1)}, \dots, \boldsymbol{x}^{(k)}\}$. Here each $\boldsymbol{x}^{(j)}$ in (C.2) is a vector of dimension $n^{|J_j|}$ with Euclidean norm no more than one. Suppose $J_j = \{t_1, t_2, \dots, t_\alpha\} \subseteq [\ell]$, then the $\mathbf{i}_{J_j}$-th entry of $\boldsymbol{x}^{(j)}$ is

$$\boldsymbol{x}^{(j)}_{\mathbf{i}_{J_j}} = \boldsymbol{x}^{(j)}_{i_{t_1}, i_{t_2}, \dots, i_{t_\alpha}}.$$

The norm defined in (C.2) is a generalization of some commonly seen vector and matrix norms. For example, when $\ell = 1$, (C.2) is reduced to the Euclidean norm of vectors in $\mathbb{R}^n$. In addition, let $\boldsymbol{A} \in \mathbb{R}^{n\times n}$ be a matrix, then $\mathcal{J}$ is a partition of $\{1,2\}$, which implies that $\mathcal{J}$ is either $\{1,2\}$ or $\{\{1\},\{2\}\}$. By the definition in (C.2), we have

$$\|\boldsymbol{A}\|_{\{1,2\}} = \sup\left\{ \sum_{i,j\in[n]} a_{ij} x_{ij} : \sum_{ij\in[n]} x_{ij}^2 \le 1 \right\} = \|\boldsymbol{A}\|_F,$$

which recovers the matrix Frobenius norm. Moreover, when $\mathcal{J} = \{\{1\},\{2\}\}$, we have

$$\|\boldsymbol{A}\|_{\{1\},\{2\}} = \sup\left\{ \sum_{i,j\in[n]} a_{ij} x_i y_j : \sum_{i\in[n]} x_i^2 \le 1, \sum_{j\in[n]} y_j^2 \le 1 \right\} = \|\boldsymbol{A}\|_2,$$

which is the operator norm of $\boldsymbol{A}$. Based on the norm $\|\cdot\|_{\mathcal{J}}$ defined in (C.2), we introduce a concentration result for polynomials of sub-Gaussian random vectors, which is a simplified version of Theorem 1.4 in Adamczak and Wolff (2015).

**Theorem C.1** (Theorem 1.4 Adamczak and Wolff (2015))**.** *Let $\boldsymbol{X} = (X_1, \dots, X_n) \in \mathbb{R}^n$ be a random vector with independent components. Moreover, we assume that such that for all $i \in [n]$, we have $\|X_i\|_{\psi_2} \le \Upsilon$, where the $\psi_2$-norm is defined in (C.1). Then for every polynomial $f : \mathbb{R}^n \to \mathbb{R}$ of degree $L$ and every $p \ge 2$ we have:*

$$\|f(\boldsymbol{X}) - \mathbb{E} f(\boldsymbol{X})\|_p \le K_L \sum_{\ell=1}^{L} \sum_{\mathcal{J}\in\mathcal{P}_\ell} \Upsilon^\ell \cdot p^{|\mathcal{J}|/2} \cdot \left\|\mathbb{E}\mathbf{D}^\ell f(X)\right\|_{\mathcal{J}}.$$

*Here $\|\cdot\|_{\mathcal{J}}$ is defined in (C.2), $\|\cdot\|_p$ is the $\ell_p$-norm of a random variable, and $\mathbf{D}^\ell f(\cdot)$ is the $\ell$-th derivative of $f$, which is takes values in order-$\ell$ tensors.*

Based on this theorem, we are ready to introduce a concentration inequality for the product of two sub-exponential random variables. This inequality might be of independent interest.

**Lemma C.4.** *Let $\{(X_i, Y_i)\}_{i\in[n]}$ be $n$ independent copies of random variables $X$ and $Y$, which are sub-exponential random variables with $\|X\|_{\psi_1} \le \Upsilon$ and $\|Y\|_{\psi_1} \le \Upsilon$ for some constant $\Upsilon$. Here the $\psi_1$-norm is specified in Definition 1.1. Then we have*

$$\left|\frac{1}{n}\sum_{i=1}^{n}\left[X_iY_i - \mathbb{E}(XY)\right]\right| \le K_\Upsilon\sqrt{\frac{\log n}{n}}$$

*with probability at least $1 - 1/(n^2)$, where $K_\Upsilon$ is an absolute constant depending solely on $\Upsilon$.*

*Proof.* For any $i \in [n]$, we define a random variable $A_i^+$ by letting $A_i^+ = X_iY_i$ if $X_iY_i \ge 0$ and 0 otherwise. In addition, let $A_i^- = A_i^+ - X_iY_i$. That is, $A_i^+$ is the non-negative part of $X_iY_i$ and $A_i^-$ is the negative par, which implies that $X_iY_i = A_i^+ - A_i^-$. In the following, we show

$$\left|\frac{1}{n}\sum_{i=1}^{n}A_i^+ - \mathbb{E}(A_i^+)\right| \le K'_\Upsilon\sqrt{\frac{\log n}{n}} \quad\text{and}\quad \left|\frac{1}{n}\sum_{i=1}^{n}A_i^- - \mathbb{E}(A_i^-)\right| \le K''_\Upsilon\sqrt{\frac{\log n}{n}} \tag{C.3}$$

for some absolute constants $K'_\Upsilon$ and $K''_\Upsilon$. Then the Lemma follows by triangle inequality.

To establish (C.3), by symmetry, it suffices to show the first inequality in (C.3). We first construct $n$ random variables $Z_i = \eta_i|A_i^+|^{1/4}$ for $i \in [n]$, where $\{\eta_i\}_{i\in[n]}$ are $n$ independent Rademacher random variables. We show that $Z_i$ is a sub-Gaussian random variable. Notice that by definition, we have $Z_i^4 = A_i^+$, which implies $\mathbb{E}(Z_i^4) = \mathbb{E}(A_i^+)$. By Cauchy-Schwarz inequality, we have

$$\mathbb{E}|Z_i|^p \le \mathbb{E}(|X_i|^{p/4}\cdot|Y_i|^{p/4}) \le \left[\mathbb{E}(|X_i|^{p/2})\cdot\mathbb{E}(|Y_i|^{p/2})\right]^{1/2}. \tag{C.4}$$

In addition, by the definition of the $\psi_1$-norm, we have

$$\left[\mathbb{E}(|X_i|^{p/2})\right]^{2/p} \le p/2\cdot\|X\|_{\psi_1}. \tag{C.5}$$

Combining (C.4) and (C.5), we obtain

$$\mathbb{E}|Z_i|^p \le \left[(p/2)^p\cdot\|X_i\|_{\psi_1}^{p/2}\cdot\|Y_i\|_{\psi_1}^{p/2}\right]^{1/2} \le (p/2)^{p/2}\cdot\Upsilon^{p/2}.$$

Hence, by the definition of $\psi_2$-norm, we have

$$\|Z_i\|_{\psi_2} = \sup_{p\ge1}\{p^{-1/2}(\mathbb{E}|Z_i|^p)^{1/p}\} \le \sup_{p\ge1}\{p^{-1/2}\cdot(p/2)^{1/2}\}\cdot\Upsilon^{1/2} = \sqrt{\Upsilon/2}, \tag{C.6}$$

which implies that $Z_i$ is a sub-Gaussian random variable. In the rest of the proof, we establish a concentration inequality for $\{Z_i^r\}_{i\in[n]}$ with $r \in \{1,2,3,4\}$; the case where $r = 4$ implies (C.3). Specifically, we establish that

$$\left|\frac{1}{n}\sum_{i=1}^{n}Z_i^r - \mathbb{E}(Z_i^r)\right| \le 4eK_\Upsilon\sqrt{\frac{\log n}{n}} \tag{C.7}$$

with probability at least $1 - 1/n^2$. Here $K_\Upsilon$ is an absolute constant that only depends on $\Upsilon$.

We recall the notation preceding Theorem C.1. Let $f(u) = u^r$ and define $F\colon \mathbb{R}^n \to \mathbb{R}$ by $F(\boldsymbol{x}) = \sum_{i=1}^n f(\boldsymbol{x}_i)$. Then by definition, the high-order derivatives of $F$ are diagonal tensors. That is, for any $i_1, i_2, \ldots, i_\ell \in [n]$, we have

$$[\mathbf{D}^\ell F(\boldsymbol{x})]_{i_1,\ldots,i_\ell} = \mathbb{1}\{i_1 = i_2 = \cdots = i_\ell\} \cdot f^{(\ell)}(\boldsymbol{x}_{i_1}).$$

For notational simplicity, for any $\ell \in \{1, \ldots, 4\}$ and $a_1, \ldots, a_n \in \mathbb{R}$, we denote by $\mathrm{diag}_\ell\{a_1, \ldots, a_n\}$ the order-$\ell$ diagonal tensor with diagonal entries $a_1, a_2, \ldots, a_n$. Using this notation, for any $\ell \in [r]$, we can write

$$\mathbf{D}^\ell F(\boldsymbol{x}) = \mathrm{diag}_\ell[f^{(\ell)}(\boldsymbol{x}_1), \ldots, f^{(\ell)}(\boldsymbol{x}_n)]. \tag{C.8}$$

Moreover, by the definition of $\|\cdot\|_{\mathcal{J}}$ in (C.2), for diagonal tensors, we have

$$\|\mathrm{diag}_\ell\{\boldsymbol{x}_1, \ldots, \boldsymbol{x}_n\}\|_{\mathcal{J}} = \mathbb{1}\{|\mathcal{J}| = 1\}\|\boldsymbol{x}\|_2 + \mathbb{1}\{|\mathcal{J}| \geq 2\}\|\boldsymbol{x}\|_{\max}. \tag{C.9}$$

In addition, note that $|f^{(\ell)}(u)| \leq r! \cdot |u|^{r-\ell}$ for any $u \in \mathbb{R}$. Combining (C.8) and (C.9), since $Z_1, \ldots, Z_n$ are i.i.d., we obtain

$$\begin{aligned}\|\mathbb{E}\mathbf{D}^\ell F(\boldsymbol{Z})\|_{\mathcal{J}} &= \mathbb{1}\{|\mathcal{J}| = 1\} \cdot \sqrt{n} \cdot |\mathbb{E}f^{(\ell)}(Z)| + \mathbb{1}\{|\mathcal{J}| \geq 2\}|\mathbb{E}f^{(\ell)}(Z)| \\ &= r! \cdot (\mathbb{1}\{|\mathcal{J}| = 1\} \cdot \sqrt{n} + \mathbb{1}\{|\mathcal{J}| \geq 2\}) \cdot \mathbb{E}(|Z|^{r-\ell}).\end{aligned} \tag{C.10}$$

By the definition of $\psi_2$-norm in (C.1), we have $\mathbb{E}(|Z|^k) \leq (\sqrt{k})^k \cdot \|Z\|_{\psi_2}^k$ for any $k \geq 1$. Therefore, by (C.10) we have

$$\|\mathbb{E}\mathbf{D}^\ell F(\boldsymbol{Z})\|_{\mathcal{J}} \leq r!(\sqrt{r-\ell})^{r-\ell}(\mathbb{1}\{|\mathcal{J}| = 1\} \cdot \sqrt{n} + \mathbb{1}\{|\mathcal{J}| \geq 2\}) \cdot \|Z\|_{\psi_2}^{r-\ell},$$

for $\ell \in [r]$, where with a slight abuse of notation we let $(\sqrt{r-\ell})^{(r-\ell)} = 1$ when $\ell = r$.

Next, we apply Theorem C.1 with $F(\boldsymbol{Z})$ to obtain

$$\begin{aligned}\|F(\boldsymbol{Z}) - \mathbb{E}F(\boldsymbol{Z})\|_p &\leq K_r \sum_{\ell \in [r]} \|Z\|_{\psi_2}^\ell \sum_{\mathcal{J} \in \mathcal{P}_\ell} p^{|\mathcal{J}|/2}(\mathbb{1}\{|\mathcal{J}| = 1\} \cdot \sqrt{n} + \mathbb{1}\{|\mathcal{J}| \geq 2\}) \cdot r!(\sqrt{r-\ell})^{r-\ell}\|Z\|_{\psi_2}^{r-\ell} \\ &\leq K_r \sum_{\ell \in [r]} \sum_{\mathcal{J} \in \mathcal{P}_\ell} (\mathbb{1}\{|\mathcal{J}| = 1\} \cdot \sqrt{np} + \mathbb{1}\{|\mathcal{J}| \geq 2\} \cdot p^2) \cdot r!(\sqrt{r-\ell})^{r-\ell}\|Z\|_{\psi_2}^r,\end{aligned}$$

where $\mathcal{P}_\ell$ is the set of partitions of $[\ell]$, the absolute constant $K_r$ depends solely on $r$. Moreover, since $\ell \leq r \leq 4$, the cardinality of $\mathcal{P}_\ell$ is bounded by a constant. Also note that $\|Z\|_{\psi_2} \leq \sqrt{\Upsilon/2}$ by (C.6). Therefore, there exists a constant $K_\Upsilon$ which depends on $\Upsilon$ such that

$$\|F(\boldsymbol{Z}) - \mathbb{E}F(\boldsymbol{Z})\|_p \leq K_\Upsilon \cdot (p^2 + \sqrt{np}).$$

Then, by Chebyshev's inequality, we have

$$\mathbb{P}(|F(\boldsymbol{Z}) - \mathbb{E}F(\boldsymbol{Z})| \geq nt) \leq \|F(\boldsymbol{Z}) - \mathbb{E}F(\boldsymbol{Z})\|_p^p \cdot (nt)^{-p} \leq t^{-p} \cdot K_\Upsilon^p \cdot \left(\sqrt{p/n} + p^2/n\right)^p. \tag{C.11}$$

Finally, we set $p = 2\lceil \log n \rceil$ and $t = 4eK_\Upsilon \sqrt{\log n/n}$ in (C.11). Note that when $n$ is sufficiently large, we have $\log^2 n/n \le \sqrt{\log n/n}$, which implies that

$$t^{-p} \cdot K_\Upsilon^p \Big(\sqrt{p/n} + p^2/n\Big)^p \le t^{-p} \cdot K_\Upsilon^p \cdot 2^p (p/n)^{p/2} \le \exp(-\lceil 2(\log n)\rceil) \le n^{-2}.$$

Therefore, by (C.11), we conclude that

$$\frac{1}{n}\sum_{i=1}^n [Z_i^r - \mathbb{E}(Z_i^r)] \le 2eK_\Upsilon \sqrt{\frac{\log n}{n}}$$

with probability at least $1 - n^{-2}$. This completes the proof. □

## C.2 Proof of Lemma A.2

*Proof.* In what follows, we show each inequality in this lemma holds with probability at least $1 - 1/(100n)$; the lemma holds by taking a union bound.

For the first part of (A.37), we apply standard results on random matrices. To simplify the notation, we let $\boldsymbol{A} = (\boldsymbol{x}^{(1)}, \boldsymbol{x}^{(2)}, \ldots, \boldsymbol{x}^{(n)})^\top \in \mathbb{R}^{n\times p}$ be the design matrix. In addition, let $\boldsymbol{B} = \boldsymbol{A}_S$, where $S = \mathrm{supp}(\boldsymbol{\beta}^*)$ satisfy $|S| \le s$. By definition, we have $\boldsymbol{\Xi} = \boldsymbol{B}^\top \boldsymbol{B}$. Note that $\mathrm{rank}(\boldsymbol{\Xi}) = \mathrm{rank}(\boldsymbol{B}) \le s$. Let $\sigma_1(\boldsymbol{B}) \ge \sigma_2(\boldsymbol{B}) \ge \ldots \sigma_s(\boldsymbol{B})$ be the leading $s$ singular values of $\boldsymbol{B}$. Then the eigenvalues of $\boldsymbol{\Xi}$ satisfy $\lambda_i(\boldsymbol{\Xi}) = \sigma_i^2(\boldsymbol{B})$ for each $i \in [s]$. By Corollary 5.35 of Vershynin (2010), with probability at least $1 - 2\exp(-t^2/2)$, we obtain

$$\sqrt{n} - \sqrt{s} - t \le \sigma_s(\boldsymbol{B}) \le \sigma_1(\boldsymbol{B}) \le \sqrt{n} + \sqrt{s} + t.$$

Squaring each term, we obtain

$$n - 2\sqrt{n}(\sqrt{s} + t) + (\sqrt{s} + t)^2 \le \lambda_s(\boldsymbol{\Xi}) \le \lambda_1(\boldsymbol{\Xi}) \le n + 2\sqrt{n}(\sqrt{s} + t) + (\sqrt{s} + t)^2. \tag{C.12}$$

When $n \ge Cs\log(ns)$ with a sufficiently large constant $C$, we have $n \ge (\sqrt{s} + \sqrt{3\log n})^2$. Setting $t = \sqrt{3\log n}$ in (C.12), we obtain that, with probability at least $1 - 2\exp(3\log n/2) \ge 1/(100n)$,

$$n - 2\sqrt{n}(\sqrt{s} + \sqrt{3\log n}) \le \lambda_s(\boldsymbol{\Xi}) \le \lambda_1(\boldsymbol{\Xi}) \le n + 3\sqrt{n}(\sqrt{s} + \sqrt{3\log n}). \tag{C.13}$$

Note that (C.13) implies $\|\boldsymbol{\Xi} - n\boldsymbol{I}_S\|_2 \le 3\sqrt{n}(\sqrt{s} + \sqrt{3\log n})$ as desired.

In addition, both the second part of (A.37) and the first part of (A.38) follows from Lemma C.1. We set $t = \sqrt{3\log n}$ in this lemma and obtain these two inequalities directly. By Lemma C.1, two inequalities hold with probability at least $1 - 4\exp(-3\log n/2) \ge 1 - 1/(100n)$ for sufficiently large $n$.

Moreover, to show (A.39), we apply Lemma C.2 with $\delta = 0.01$. Thus the we show that (A.39) holds with probability at least $1 - 1/(100n)$, provided $n \ge Cs\log s$ with constant $C$ sufficiently large.

Finally, for the second part of (A.38), recall that we define

$$\widetilde{\boldsymbol{x}}_S^{(i)} = (\mathbf{I}_S - \boldsymbol{\beta}^* \boldsymbol{\beta}^{*\top}) \cdot \widetilde{\boldsymbol{x}}^{(i)},$$

which is independent of $\boldsymbol{\beta}^{*\top}\boldsymbol{x}^{(i)}$. Thus $\widetilde{\boldsymbol{x}}_S^{(i)}$ and $y^{(i)}$ are independent. Furthermore, let $\boldsymbol{\xi}_1, \boldsymbol{\xi}_2, \ldots, \boldsymbol{\xi}_{s-1}$ and $\boldsymbol{\beta}^*$ be a orthonormal base for the $s$-dimensional subspace of $\mathbb{R}^p$ spanned by vectors restricted on support $S$. Thus each $\boldsymbol{\xi}_i \in \mathbb{R}^p$ is perpendicular to $\boldsymbol{\beta}^*$ and is supported $S$. We denote $\boldsymbol{U} = (\boldsymbol{\xi}_1, \boldsymbol{\xi}_2, \ldots, \boldsymbol{\xi}_{s-1}) \in \mathbb{R}^{p\times(s-1)}$, which is a matrix with orthonormal columns. Thus we can rewrite each $\widetilde{\boldsymbol{x}}_S^{(i)}$ as $\widetilde{\boldsymbol{x}}_S^{(i)} = \boldsymbol{U}\boldsymbol{u}^{(i)}$, where each $\{\boldsymbol{u}^{(i)}\}_{i\in[n[} \subseteq \mathbb{R}^{s-1}$ are $n$ i.i.d. $s-1$-dimensional standard Gaussian random vectors.

Note that each $\boldsymbol{u}^{(i)}$ is independent with $y^{(i)}$. Then we can write $\boldsymbol{\Phi}$ in (A.36) as

$$\boldsymbol{\Phi} = n \cdot \boldsymbol{U}\boldsymbol{\Phi}_1\boldsymbol{U}^\top, \quad \text{where} \quad \boldsymbol{\Phi}_1 = \frac{1}{n}\sum_{i=1}^n (y^{(i)} - \mu)\boldsymbol{u}^{(i)}\boldsymbol{u}^{(i)\top}.$$

Hence we have $\|\boldsymbol{\Phi}\|_2 = n \cdot \|\boldsymbol{\Phi}_1\|_2$. In the following, we establish an upper bound for $\|\boldsymbol{\Phi}_1\|$. Since $y^{(i)}$ and $\boldsymbol{u}^{(i)}$ are independent, conditioning on $\{y^{(i)}\}_{i\in[n]}$, by setting $t = \log n$ and $m = s-1$ in Lemma C.3, with probability at least $1 - \exp(-4(s + \log n)) \geq 1 - 1/(200n)$, we have

$$\begin{aligned} &\big\|\boldsymbol{\Phi}_1 - (\mu_n - \mu)\cdot \mathbf{I}_{s-1}\big\|_2 \\ &\quad \leq C_0\bigg\{\bigg[\frac{(s-1+\log n)}{n^2}\cdot\sum_{j=1}^n (y^{(i)} - \mu)^2\bigg]^{1/2} + \frac{(s-1+\log n)}{n}\cdot\max_{j\in[n]}|y^{(j)} - \mu|\bigg\}, \end{aligned} \tag{C.14}$$

where we define $\mu_n = n^{-1}\sum_{i=1}^n y^{(i)}$.

Note that we have $\mathbb{E}(Y) = \mu$ and $\|Y\|_\psi \leq \Psi$, which implies that $\|Y - \mu\|_{\psi_1} \leq 2\|Y\|_{\psi_1}$. Using the concentration of polynomials of sub-exponential random variables in (C.7), we obtain with probability at least $1 - 2/n^2$ that

$$\bigg|\sum_{j=1}^n\Big[|y_i - \mu|^2 - \mathbb{E}\big(|y_i - \mu|^2\big)\Big]\bigg| \leq C_1\sqrt{n\log n}, \quad \bigg|\sum_{j=1}^n (y_i - \mu)\bigg| \leq C_1'\sqrt{n\log n}, \tag{C.15}$$

where $C_1$ and $C_1'$ are absolute constants that only depends on $\Psi$. Furthermore, by the tail probability of the maxima of sub-exponential random variables, with probability at least $1 - 1/n^2$, we obtain

$$\max_{j\in[n]}|y_j - \mu| \leq C_2\log n, \tag{C.16}$$

for some constant $C_2$ depending only on $\Psi$. Hence, (C.15) and (C.16) hold simultaneously with probability at least $1 - 3/n^2 \geq 1 - 1/(200n)$ for sufficiently large $n$. Moreover, since $\mathbb{E}(|y_i - \mu|^2) = \mathrm{Var}(Y)$ is a constant, by (C.15) we have

$$\sum_{j=1}^n |y_i - \mu|^2 \leq C_3 n, \tag{C.17}$$

for some absolute constant $C_3$. Combining (C.14), (C.16), and (C.17), when $n \geq s\log^2 n + \log^3 n$, we obtain

$$\big\|\boldsymbol{\Phi}_1 - (\mu_n - \mu)\cdot\mathbf{I}_{s-1}\big\|_2 \leq C\sqrt{(s-1)\log n/n}$$

for some absolute constant $C$.

Finally, note that the right part of (C.15) is equivalent of $|\mu_n - \mu| \le C_1' \sqrt{\log / n}$. This implies that

$$\|\boldsymbol{\Phi}\|_2 = n \cdot \|\boldsymbol{\Phi}_1\|_2 \le (C + C_1') \cdot \sqrt{ns \log n},$$

which establish the second part of (A.38). Therefore, we conclude the proof of this lemma. $\square$

## C.3 Proof of Lemma A.3

*Proof.* We prove that each inequality in the lemma holds with probability at least $1 - 1/(100n)$ for some constant $K_2$ sufficiently. Then the lemma holds by applying a union bound.

To begin with, note that we have $\mathbb{E}(Y) = \mu$ and $\mathbb{E}[Y(\boldsymbol{X}^\top \boldsymbol{\beta}^*)^2] = \mu + \rho$. Also note that both $Y$ and $(\boldsymbol{X}^\top \boldsymbol{\beta}^*)^2$ are sub-exponential random variables with finite $\psi_1$-norms. Following directly from Lemma C.4, the two inequalities in (A.40) hold with probability at least $1 - 1/(100n)$, provided $n \ge 100$. Here the constant $K_2$ only depends on $\Psi$.

Next, we establish the two inequalities in (A.41) and the second part of (A.42) using similar arguments. To simplify the notation, we define

$$\boldsymbol{g} = \frac{1}{n} \sum_{i=1}^{n} (\boldsymbol{\beta}^{*\top} \boldsymbol{x}^{(i)}) \cdot y^{(i)} \widetilde{\boldsymbol{x}}^{(i)}, \quad \widetilde{\boldsymbol{x}}^{(i)} = \boldsymbol{x}^{(i)} - (\boldsymbol{\beta}^{*\top} \boldsymbol{x}^{(i)}) \cdot \boldsymbol{\beta}^*.$$

Note that $\widetilde{\boldsymbol{x}}^{(i)}$ and $(\boldsymbol{\beta}^{*\top} \boldsymbol{x}^i) y^{(i)}$ are independent, which implies that $\mathbb{E}[\boldsymbol{g}] = \boldsymbol{0}$. Furthermore, we denote $S = \operatorname{supp}(\boldsymbol{\beta}^*)$. For each $j \in S$, both $(\boldsymbol{\beta}^{*\top} \boldsymbol{x}^{(i)}) \cdot \widetilde{\boldsymbol{x}}_j^{(i)}$ and $y^i$ are sub-exponential random variables with finite $\psi_1$-norms. Therefore, by Lemma C.4, with probability at least $1 - 1/(n^2)$, we have

$$|\boldsymbol{g}_j| \le K_2 \sqrt{\log(ns)/n}$$

for some large constant $K_2$. By taking a union bound over all $j \in S$, when $n \ge 100s$, we conclude that with probability at least $1 - 1/(100n)$, $\|\boldsymbol{g}_S\|_\infty \le K_2 \sqrt{\log(ns)/n}$, which implies that

$$\|\boldsymbol{g}\|_2 \le \sqrt{s} \|\boldsymbol{g}\|_\infty \le K_2 \sqrt{s \log(ns)/n}.$$

Thus we establish the first part of (A.41). Similarly, since both $(\boldsymbol{\beta}^{*\top} \boldsymbol{x}^{(i)}) \cdot \widetilde{\boldsymbol{x}}_j^{(i)}$ and $(\boldsymbol{\beta}^{*\top} \boldsymbol{x}^{(i)})^2$ are sub-exponential random variables with finite $\psi_1$-norms, by Lemma C.4 we obtain the second part of (A.41). In addition, to establish the second part of (A.42), we first apply Lemma C.4 with $(\boldsymbol{\beta}^{*\top} \boldsymbol{x}^{(i)})$ and $\widetilde{\boldsymbol{x}}_j^{(i)}$ and then take a union bound over $j \in S$.

Finally, we prove that the first part of (A.42) holds with probability at least $1 - 1/100n$. Let $f(x, y)$ be a polynomial on $\mathbb{R}^2$ with bounded degree. Since $|Y|$ is sub-exponential and $|\boldsymbol{X}^\top \boldsymbol{\beta}^*|$ is sub-Gaussian, we obtain that

$$\mathbb{E}\big[f\big(|Y|, |\boldsymbol{X}^\top \boldsymbol{\beta}^*|\big)\big] = C_1 \quad \text{and} \quad \operatorname{Var}\big[f\big(|Y|, |\boldsymbol{X}^\top \boldsymbol{\beta}^*|\big)\big] = C_2$$

for some constants $C_1$ and $C_2$. Hence, by Chebyshev's inequality, for any $t > 0$, we have

$$\mathbb{P}\biggl\{\biggl|\sum_{i=1}^{n} f\bigl(|y^{(i)}|, |\boldsymbol{x}^{(i)\top}\boldsymbol{\beta}^*|\bigr) - \mathbb{E}\bigl[f\bigl(|Y|, |\boldsymbol{X}^\top\boldsymbol{\beta}^*|\bigr)\bigr]\biggr| \geq tn\biggr\} \leq \frac{\mathrm{Var}\bigl[f\bigl(|Y|, |\boldsymbol{X}^\top\boldsymbol{\beta}^*|\bigr)\bigr]}{nt^2} \leq \frac{C_2}{nt^2}. \tag{C.18}$$

Now we set the polynomial $f(x, y)$ to be

$$f(x, y) = \biggl(\sum_{r=0}^{10} x^r\biggr) \cdot \biggl(\sum_{r=0}^{10} y^r\biggr).$$

Setting $t$ in (C.18) to be a sufficiently large constant, we show that the first part of (A.42) holds with probability at least $1 - 1/100n$. Therefore, we conclude the proof of this lemma. □